\documentclass[journal]{IEEEtran}
\usepackage{amsmath,graphicx,bm}
\usepackage[labelformat=simple]{subcaption}
\usepackage{algorithmic,amsfonts}
\usepackage{algorithm}
\usepackage{cite,float}

\usepackage{xcolor}
\usepackage{float}

\newcommand{\tabincell}[2]{\begin{tabular}{@{}#1@{}}#2\end{tabular}}

\title{An Efficient Hypergraph Approach to
Robust Point Cloud Resampling}
\author{Qinwen~Deng, Songyang~Zhang,~\IEEEmembership{Student Member,~IEEE},
	and~Zhi~Ding,~\IEEEmembership{Fellow,~IEEE}
	\IEEEcompsocitemizethanks{
		\IEEEcompsocthanksitem 
		Q. Deng, S. Zhang, and Z. Ding are with Department of Electrical and Computer Engineering, University of California, Davis, CA, 95616. (E-mail: mrdeng@ucdavis.edu, sydzhang@ucdavis.edu, and zding@ucdavis.edu).
	}    
}
\begin{document}
%
\maketitle
\begin{abstract}
Efficient processing and feature extraction of large-scale point clouds 
are important in related computer vision and cyber-physical systems.
This work investigates point cloud resampling based on
hypergraph signal processing (HGSP) to 
better explore the underlying relationship among 
different cloud points and 
to extract contour-enhanced features. 
Specifically, we design hypergraph spectral filters to capture 
multi-lateral interactions among the signal nodes of point clouds and 
to better preserve their surface outlines. 
Without the need and the computation
to first construct the underlying hypergraph,
our low complexity approach directly
estimates hypergraph spectrum of point clouds by leveraging hypergraph 
stationary processes from the observed 3D coordinates.
Evaluating the proposed resampling methods with several metrics,
our test results validate the high efficacy of hypergraph characterization of
point clouds and demonstrate 
the robustness of hypergraph-based resampling
under noisy observations. 
\end{abstract}
\section{Introduction}

3D perception plays an important role in the 
high growth fields of robotics and cyber-physical systems
and continues to drive many progresses made in
advanced point cloud processing.
3D point clouds provide efficient exterior
representation for complex objects and their surroundings. 
Point clouds have seen broad applications in many areas, such as computer vision, autonomous driving and robotics. Notable examples of point cloud processing include 
surface reconstruction \cite{d1}, 
rendering \cite{d2}, feature extraction \cite{d3}, 
shape classification \cite{g1}, and object detection/tracking \cite{g2}. 
When constructing
point cloud of a target object, however, modern laser scan systems 
can generate millions of data points \cite{g3}. 
To achieve better storage efficiency and lower point cloud
processing complexity, 
point cloud resampling aims to reduce the number of points 
in a cloud to achieve data compression while preserving 
the vital 3D structural and surface features. 
Point cloud resampling represents an important tool in various applications
such as point cloud
segmentation, object classification, and efficient transmission/storage.
An example of point cloud resampling proposed in \cite{c3} suggested a 
graph-based filter to downsample point clouds and to capture the 
original object surface contour. 

The literature already contains a variety of works on 
different aspects of point cloud resampling. For instance,
a centroidal Voronoi tessellation method in \cite{c1} can 
progressively generate
high-quality resampling results with isotropic or anisotropic distributions 
from a given point cloud to form compact representations of the 
underlying cloud surface. Another 
3D filtering and downsampling technique \cite{c2} relies on a 
growing neural gas network, 
to deal with noise and outliers within
data provided by Kinect sensors. 
Of particular interest is graph-based resampling approach
which has exhibited desirable capability to capture the underlying 
structures of point clouds \cite{f1}. The
graph-based method of \cite{f1b} applies embedded binary trees 
to compress the dynamic point cloud data. Another interesting work
\cite{c3} proposes several graph-based filters 
to capture the distribution of point data to achieve
computationally efficient resampling. {In addition, a contour-enhanced resampling method
introduced in \cite{d4}  
utilizes graph-based high-pass filters.} 

In addition to the aforementioned class of
graph-based methods, a competing class of feature-based approaches via 
edge detection 
and feature extraction has also been popular.
In \cite{c4}, the authors presented a sharp feature detector via
Gaussian map clustering on local neighborhoods. 
Bazazian \textit{et al.} \cite{c5} extended this principle by 
leveraging principal component analysis (PCA)
to develop a new agglomerative clustering method. 
The efficiency and accuracy of this work can further benefit
from spectral analysis of the covariance matrix defined by $k-$nearest neighbors. Another typical approach represented by \cite{f2}
processes a noisy and possibly outlier-ridden point set in an 
edge-aware manner.

Both graph-based and feature-based  methods have clearly
achieved successes in point cloud resampling. However, some
limitations remain. Graph-based methods tend to focus on 
pairwise relationship between different points, since each graph
edge only connects two signal nodes. However, it is clear that
multilateral interactions of data points
could model the more informative characteristics of 3D point clouds. 
Bilateral graph node connections cannot even describe
multilateral relationship among 
points on the same surface (e.g. 3 points of a triangle) directly 
\cite{c11}. Furthermore, in graph-based methods, 
efficient construction of a suitable graph to 
represent an arbitrary point cloud poses another challenge.
Among feature-based methods, performance would vary with respect to
the feature selection and filter designs. 
The open issues are the adequate and robust selection of features and
filter parameters for practical point cloud processing. 

More recently, hypergraphs have been successfully applied in representing 
and characterizing the underlying multilateral interactions among
multimedia data points \cite{d5}. A hypergraph extends basic graph
concept into 
higher dimensions, 
in which each hyperedge can connect more than two nodes \cite{f2}.
Therefore for point clouds, hypergraph provides a more general
representation to characterize multilateral relationship
for points on object surfaces such that one hyperedge can cover 
multiple nodes on the same surface. Furthermore, by
generalizing graph signal processing \cite{d9},
hypergraph signal processing (HGSP) \cite{d6}\cite{d8} 
provides a theoretical foundation for spectral analysis 
in hypergraph-based point cloud processing. Specifically, 
stationarity-based hypergraph estimation, in conjunction 
with hypergraph-based filters, has
demonstrated notable successes in 
processing point clouds for 
various tasks including 
segmentation, sampling, and denoising \cite{d7,c11}.

\begin{figure}[t]
	\begin{subfigure}{.24\textwidth}
		\centering
		\includegraphics[width=1\linewidth]{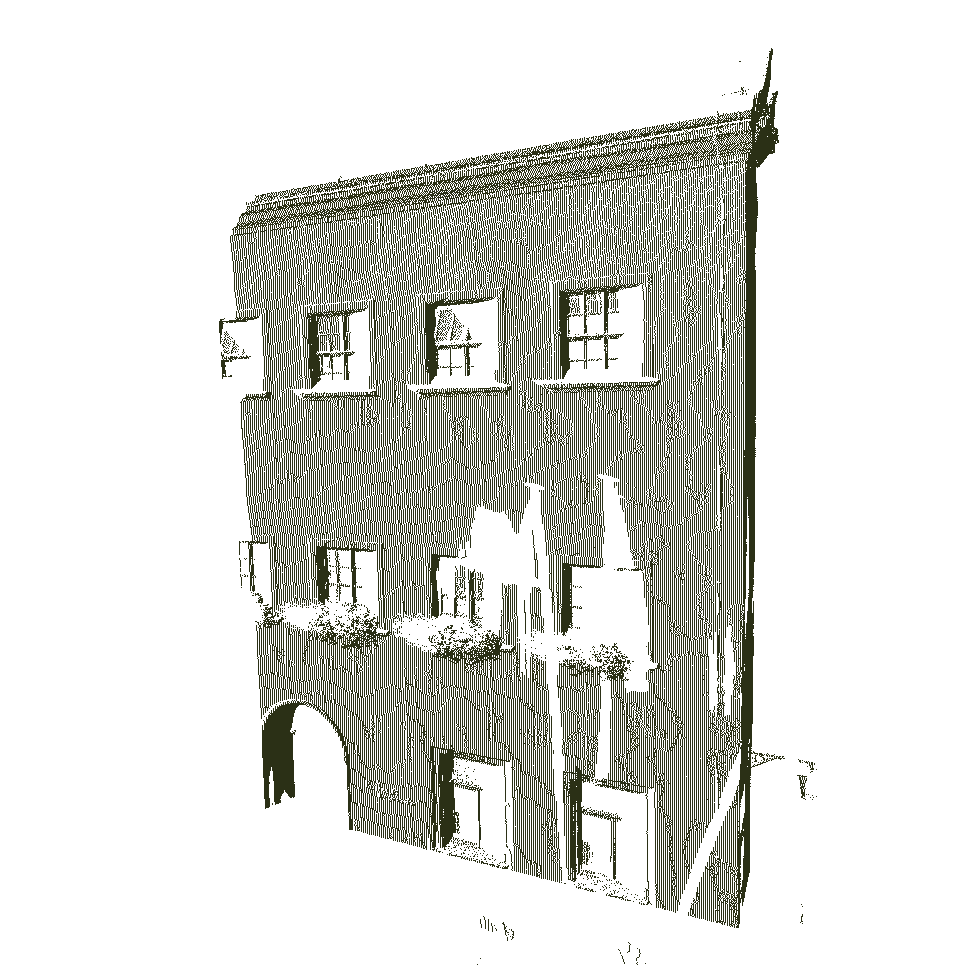}
		\caption{Original.}
		\label{bid1}
	\end{subfigure}
	\begin{subfigure}{.24\textwidth}
		\centering
		\includegraphics[width=1\linewidth]{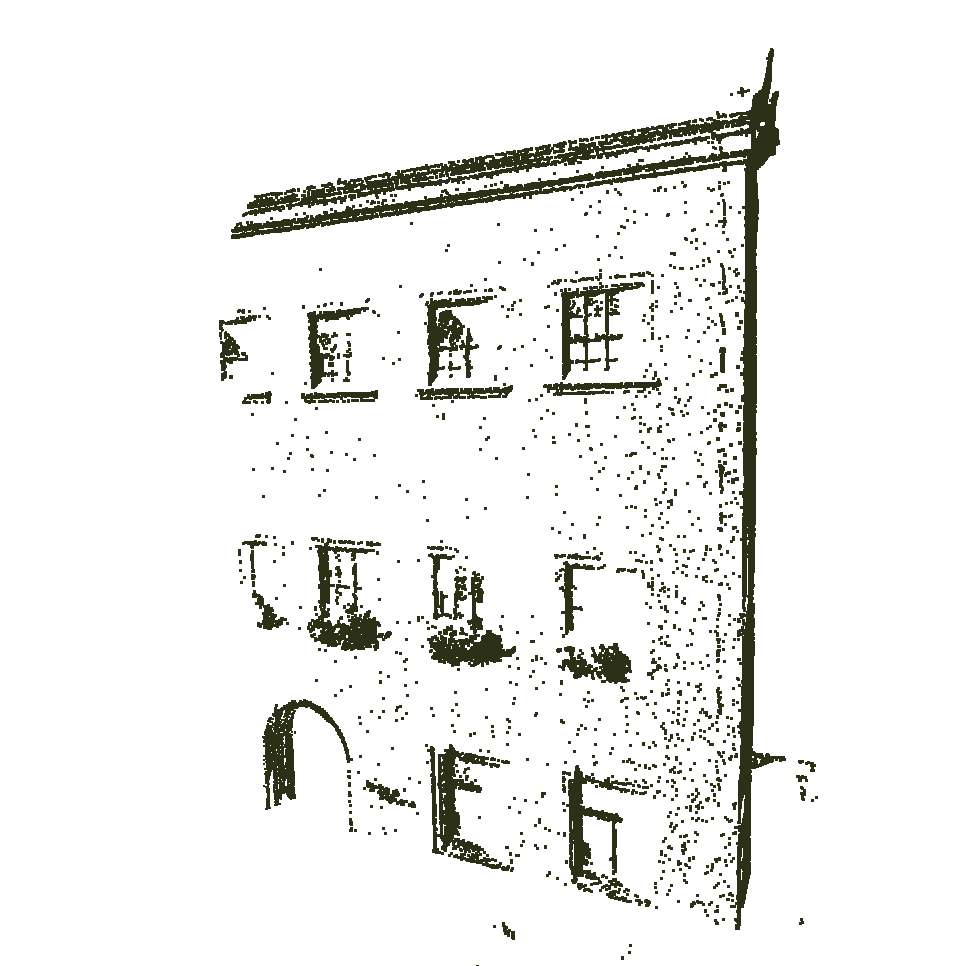}
		\caption{Resampled.}
		\label{bid2}
	\end{subfigure}
	\caption{Example of Contour-Enhanced Resampling: (a) original point cloud with 272705 points, and (b) resampled building based on the proposed local hypergraph filtering with 20\% samples.}
	\label{bid}
\end{figure}

In this paper, we investigate point cloud resampling based on 
hypergraph spectral analysis. Instead of the traditional uniform 
resampling, we investigate contour-enhanced resampling to select subset 
of cloud points and to extract distinct surface features. A heuristic example
is illustrated in Fig. \ref{bid} showing a point cloud successfully
resampled with only 20\% samples for the building.
To briefly highlight the novelty of our proposed approaches, 
we estimate the hypergraph spectrum basis for point clouds
{under study} by leveraging the hypergraph stationary process. 
We propose three novel 3D point cloud resampling methods:
\begin{itemize}
\item[1)] Hypergraph kernel convolution method (HKC);
\item[2)] Hypergraph kernel filtering method (HKF);
\item[3)] Local hypergraph filtering method (LHF). 
\end{itemize}

The kernel convolution method defines a local smoothness among
signals based on an operator and hypergraph convolution. 
The kernel filtering method defines the local smoothness with respect 
to high-pass filtering in spectrum domain. 
The local hypergraph filtering method utilizes a local sharpness
definition with respect to high-pass filtering in spectrum domain. 
In order to test the model preserving property 
on complex point cloud models, 
we apply a simple method for point cloud recovery based on alpha complex 
and Poisson sampling. We then test the proposed methods under several metrics 
to demonstrate the compression efficiency and robustness of our proposed 
resampling methods with respect to the general feature preservation of
point clouds under study. 

We summarize the major contributions of this manuscript: 
\begin{itemize}
	\item We propose three novel hypergraph-based resampling methods to 
	preserve distinct and sharp point cloud features; 
	\item We provide novel definitions of hypergraph-based indicators 
to evaluate the smoothness or sharpness over point clouds;
	\item We apply different metrics to demonstrate the effectiveness of 
	our propose methods.
\end{itemize}

We organize the rest of the manuscript as follows. 
Section II briefly describe basic point cloud model and 
introduces the fundamentals of hypergraph signal processing.
We develop the foundation of HGSP based point cloud resampling
and derive three new resampling methods in Section III. 
We provide the test results of the proposed resampling methods
in Section IV, before formalizing our conclusions in Section V.

\section{Fundamentals and Background}

\subsection{Point Cloud}

A point cloud is a collection of points on
the surface of a 3D target object. 
Each point consists of its 3D coordinates 
and may contain further features, such as colors and normals \cite{c6}. 
In this work, we focus on the coordinates of data points and point cloud 
resampling. A point cloud can be represented by the
coordinates of $N$ data points written as
an $ N\times 3$ real-valued location matrix 
\begin{equation}
		\mathbf{P}=[\mathbf{X_1\quad X_2\quad X_3}]=
	\begin{bmatrix}
	\mathbf{p}_1^T\\
	\mathbf{p}_2^T\\
	\ddots\\
	\mathbf{p}_N^T
	\end{bmatrix}\in\mathbb{R}^{N\times 3},
\end{equation}
where $\mathbf{X}_i$ denotes a vector of the $i$-th coordinates 
of all $N$ data points
whereas $3\times 1$ vector $\mathbf{p}_i$ indicates the $i$-th point's coordinates.

\subsection{Hypergraph Signal Processing} 

Hypergraph signal processing (HGSP) is an analytic framework 
that uses hypergraph and
tensor representation to model high-order signal interactions \cite{d6}. 
Within this framework, an $M$th-order $N$-dimensional representation tensor $\textbf{A}=(a_{i_1 i_2 \cdots i_M})\in \mathbb{R}^{N^M}$ models 
a hypergraph with $N$ vertices in each hyperedge, which is
capable of connecting 
maximum of $M$ nodes. We may call the number of nodes connected by
a hyperedge as its
{length}. Weights of hyperedges with {length} 
less than $M$ 
are normalized according to combinations and permutations \cite{d11}.

Orthogonal CANDECOMP/PARAFAC (CP) decomposition 
\cite{f5}-\cite{f7}
enables the (approximate) decomposition of a representing tensor into
\begin{align}\label{CP_de}
\mathbf{A}\approx\sum_{r=1}^N \lambda_r \cdot \underbrace{\mathbf{f}_r \circ \ldots \circ \mathbf{f}_r}_{M \text{ times}},
\end{align}
where $\circ$ denotes tensor outer product, 
$\{\mathbf{f}_1,\cdots,\mathbf{f}_N\}$ are orthonormal basis to
represent spectrum components, and $\lambda_r$ is the 
$r$-th spectrum coefficient corresponding to the $r$-th
basis. Spectrum components \{$\mathbf{f}_1,\cdots,\mathbf{f}_N$\} 
form the full hypergraph spectral space. 

Similar to GSP, hypergraph signals are attributes of nodes. 
Intuitively, a signal is defined as $\mathbf{s}=[s_1\quad s_2\quad...\quad s_N]^{\mathrm{T}}\in\mathbb{R}^{N}$. Since the adjacency 
tensor $\mathbf{A}$ describes high-dimensional interactions of signals, 
we define a special form of the hypergraph signal 
to work with the representing tensor, i.e.,
\begin{equation}
\mathbf{s}^{[M-1]}=\underbrace{\mathbf{s\circ...\circ s}}_{\text{M-1 times}}.
\end{equation}
Given the definitions of hypergraph spectrum and hypergraph 
signals, hypergraph Fourier transform (HGFT) is given by
\begin{align}\label{hgft}
\mathbf{\hat s}
=\mathcal{F}_C(\mathbf{s})
=[(\mathbf{f}_1^{\mathrm{T}}\mathbf{s})^{M-1}\cdots(\mathbf{f}_N^{\mathrm{T}}\mathbf{s})^{M-1}]^\mathrm{T}.
\end{align}

From the graph specific HGFT,  hypergraph spectral convolution 
can be generalized \cite{d5} as 
\begin{equation}\label{conv}
\mathbf{x}\diamond \mathbf{y}=\mathcal{F}_C^{-1}(\mathcal{F}_C(\mathbf{x})
\odot\mathcal{F}_C(\mathbf{y})),	
\end{equation}
where $\mathcal{F}_C$ is the HGFT, $\mathcal{F}_C^{-1}$ denotes 
inverse HGFT (iHGFT), 
and $\odot$ denotes 
Hadamard product \cite{d6}. This definition applies the basic relationship
between convolution and spectrum product, and generalizes
convolution in the vertex domain into product in the hypergraph spectrum domain.

To be concise, we refrain from a full review
of HGSP here.  Instead, we refer readers to \cite{d6} and related works for
a more extensive introduction of
HGSP concepts, such as filtering, hypergraph Fourier transform, and sampling theory,
among others. Equally important are concept and properties of
hypergraph stationary processes which can be found in, e.g., \cite{c11}.

\section{HGSP Point Cloud Resampling}

Based on the framework of HGSP, we now develop three new 
edge-preserving resampling methods for point clouds: 
1) kernel convolution based method, 2) kernel filtering based method, and 3) local hypergraph filtering based method.

\subsection{Hypergraph Kernel Convolution (HKC) based Method}

\begin{figure}[htb]
	\centering
	\begin{subfigure}{.2\textwidth}
		\centering
		\includegraphics[width=1\linewidth]{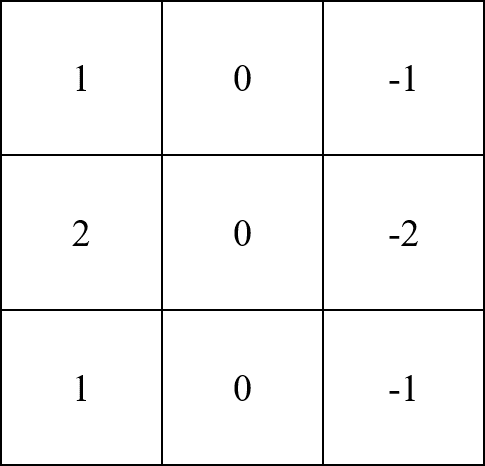}
		\caption{2D Sobel Kernel.}
	\end{subfigure}
	\hspace{0.2cm}
	\begin{subfigure}{.2\textwidth}
		\centering
		\includegraphics[width=1\linewidth]{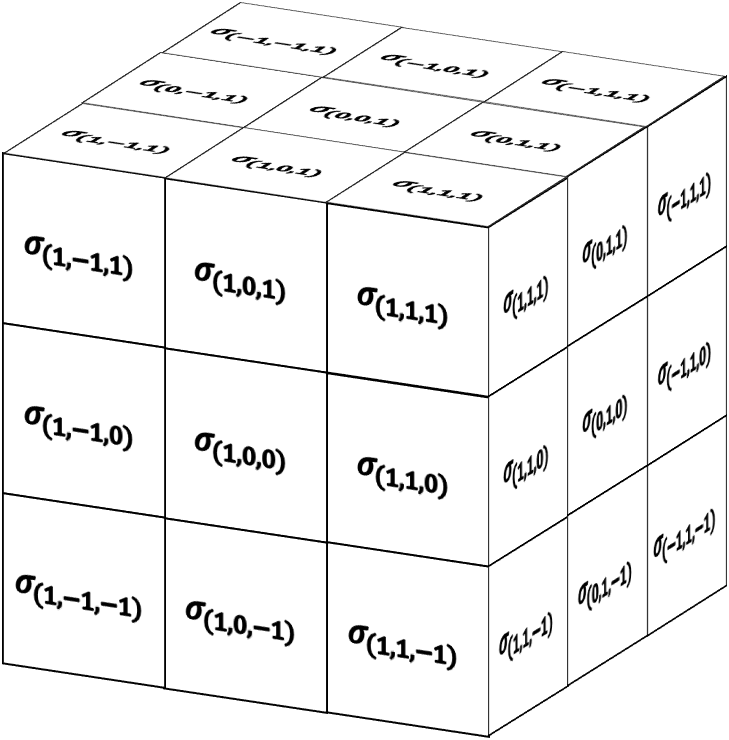}
		\caption{3D Hypergraph Kernel.}
	\end{subfigure}
	\caption{Example of Convolution Kernels.}
	\label{conv1}
\end{figure}

In traditional image processing, kernel convolution methods such as Sobel and Prewitt 
\cite{f8} have achieved notable successes in edge detection.
Inspired by these 2D kernel convolution methods in 2D image processing, e.g.
Fig. \ref{conv1}(a), we 
define a square $k\times k\times k$ 3D cube as the slicing block to 
define a local signal {$\mathbf{s}_i\in\mathbb{R}^{N_k}$} 
and a convolution kernel {$\mathbf{G}\in\mathbb{R}^{N_k}$}
with $N_k=k^3$ aimed at extracting sharp outliers of the
point cloud under study. 
{An example of $3^3$ cubic 3D convolution kernel is shown} in Fig. \ref{conv1}(b). Note that the hypergraph convolution kernel can assume
different shapes and sizes depending on the datasets. 

\begin{figure}[htb]
	\centering
	\includegraphics[width=.4\textwidth]{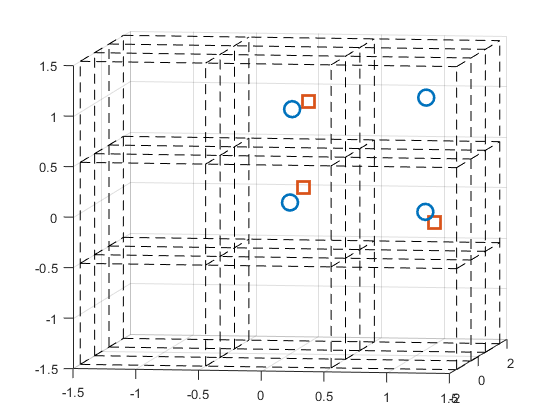}
	\caption{An example of local signal $\mathbf{s}_i$: Points in the middle layer 
are marked by blue circles and points in the back layer are marked by 
red squares. The voxels of slicing block are delineated by the dash black 
lines. 
There is at most one point in each voxel in this example, 
such that $\mathbf{s}_i(n)=1$ or 0, respectively,  depending on whether 
or not the $n$-th voxel contains a point.}
	\label{fig:si}
	\vspace{-2mm}
\end{figure}

For the $i$-th point in an original point cloud, its corresponding local signal $\mathbf{s}_i\in \mathbb{R}^{N_k}$ is defined according to the number of points 
in the voxel of kernel centered at $i$-th point. An example of the local signal 
is shown 
in Fig. \ref{fig:si}. 
Although the idea behind the use of, e.g., 3D sobel operator 
is straightforward, technical obstacles 
arise mainly due to two reasons: (1) nodes in 3D point cloud are not always on grid; (2) two signals in graph/hypergraph based convolution must have the same length. Thus, we let $N_k$ be 
equal to the number of voxels in the 
kernel. Let $d$ be the distance between the centers of two nearby voxels in the kernel. A proper selection of $d$ should allow $\mathbf{s}_i$ to 
capture the local 
geometric information. If $d$ is too small, only a few neighbors of $i$-th 
point are included in the $\mathbf{s}_i$ {and it is sensitive to 
measurement noise}; If $d$ is too large, 
large number of neighbors are included 
in each voxel, which may lead to the blurring of detailed local 
geometric information. We choose $d$ to be the intrinsic resolution 
of a point cloud. 

Given the definition of signals, we now propose a new
hypergraph convolution based method. Our kernel convolution based 
method consists of two main steps: i) hypergraph spectrum estimation, 
and ii) kernel convolution using hypergraph spectrum. 

We first estimate the hypergraph spectrum based on 
hypergraph stationary process. Introduced in \cite{c11}, 
a stochastic signal $\mathbf{x}$ is weak-sense hypergraph 
stationary {if and only if} it has zero-mean and its covariance matrix 
has the same eigenvectors as the hypergraph spectrum basis, i.e., 
\begin{equation}
	\mathbb{E}[\mathbf{x}] =\mathbf{0},
\end{equation} and
\begin{equation}\label{s2}
	\mathbb{E}[\mathbf{x}\mathbf{x}^H]=\mathbf{V}\Sigma_\mathbf{x}\mathbf{V}^{H},
\end{equation}
where $\mathbf{V}$ is the hypergraph spectrum basis. Since the 3D coordinates 
can be viewed as three observations of the point 
cloud from different projection 
directions, 
we can estimate the hypergraph spectrum from the covariance matrix of three coordinates based on assumption of signal stationarity. 
Here, we use the same spectrum estimation strategy as \cite{c11}, and consider 
the adjacency tensor as a third-order tensor, 
which is the minimum number of nodes 
to form a surface. 

\subsubsection{HKC Algorithm}
Let {$\mathbf{P_c}=[\mathbf{X_1}\ \mathbf{X_2}\ \mathbf{X_3}]\in \mathbb{R}^{N_k\times3}$} be the coordinates of the total $N_k$ 
voxel centers in the kernel. For our $3\times 3\times 3$ kernel example, $N_k=27$. {We then normalize the coordinates $\mathbf{P_c}$ to 
obtain a zero mean signal $\mathbf{P'_c}$.} By calculating the eigenvectors \{$\mathbf{f}_1,\cdots,\mathbf{f}_{N_k}$\} for {$R_{\mathbf{P'_c}}=\mathbf{P'_c}\mathbf{P'_c}^T$}, we can estimate the 
hypergraph spectrum basis $\mathbf{V}=[\mathbf{f}_1,\cdots,\mathbf{f}_{N_k}]$.

Next, we implement convolution between the local signal $\mathbf{s}_i$ and the 
kernel $\mathbf{G}$. Since we consider the third-order tensor 
and are only interested in signal energies in the spectrum domain, 
we can utilize the hypergraph spectrum to calculate a 
simplified-form of HGFT $\hat{\mathbf{s}}_i$ and a corresponding
inverse HGFT (iHGFT) of original signals $\mathbf{s}_i$, instead of the HGFT and iHGFT of the hypergraph signal $\mathbf{s}_i^{[M-1]}$ in Eq. (\ref{hgft}). 

Recall from \cite{d6} that simplified HGFT and iHGFT of original signal 
can are, respectively, 
\begin{align}
\mathcal{F}(\mathbf{s}_i)&=\hat{\mathbf{s}}_i=\mathbf{V}^H\mathbf{s}_i,\label{FT}\\
\mathcal{F}^{-1}(\hat{\mathbf{s}}_i)&=\mathbf{s}_i=\mathbf{V}\hat{\mathbf{s}}_i.\label{iFT}
\end{align}
Note that the hypergraph signal is an $(M-1)$-fold tensor 
outer product of the original signal in vertex domain,
corresponding to $(M-1)$-fold Hadamard product in spectrum domain, 
where they share the same bandwidth.

\floatname{algorithm}{Algorithm}
\begin{algorithm}[t]
	\caption{Hypergraph Kernel Convolution (HKC)}
	\label{alg:A}
	\begin{algorithmic}
		\STATE {\textbf{Input}: A point cloud of $N$ nodes with coordinates $\mathbf{P}=[\mathbf{p}_1^T \cdots \mathbf{p}_N^T]^T$, resampling ratio $\alpha$}.
		\STATE {\textbf{1.}} {Calculate the intrinsic resolution of point cloud};
		\STATE {\textbf{2.}} {{Use intrinsic resolution $d$ to find coordinates $\mathbf{P_c}\in\mathbb{R}^{N_k\times3}$ of voxel centers in the kernel}};
		\STATE {\textbf{3.}} {{Use the coordinates $\mathbf{P_c}$ of voxel centers in the kernel to estimate the hypergraph spectrum basis $\mathbf{V}=[\mathbf{f}_1,\cdots,\mathbf{f}_{N_k}]$ and corresponding eigenvalues $\bm{\lambda}$}};
		\FOR{$i=1,2,\cdots, N$}
		\STATE {\textbf{4.}} {Use hypergraph spectrum basis $\mathbf{V}$ to calculate the Fourier transform $\hat{\mathbf{s}}_i$ in Eq. (\ref{FT})};
		\STATE {\textbf{5.}} {Calculate the Hadamard product $\hat{\mathbf{s}}_{oi}=\hat{\mathbf{s}}_i \odot \hat{\mathbf{G}}$};
		\STATE {\textbf{6.}} {Calculate inverse Fourier transform $\mathbf{s}_{oi}$ using Eq. (\ref{iFT})};
		\STATE {\textbf{7.}} {Calculate local smoothness $\beta_i$ in Eq. (\ref{smooth1}) };
		\ENDFOR
		\STATE {\textbf{8.}} {Sort the local smoothness $\beta_i$ in descending order
		and select the bottom $N_r=\alpha N$ nodes as the resampled point cloud}.
	\end{algorithmic}
\end{algorithm}

Directly designing convolution
kernel $\mathbf{G}$ in vertex domain is challenging 
for two main reasons. First, since hypergraph spectrum basis $\mathbf{V}$ would vary for different kernels, finding a general $\mathbf{G}$ 
in vertex domain that performs equally well for various different bases
would be difficult. Second, since the convolution result $\mathbf{s}_{oi}$
between the signal $\mathbf{s}_i$ and the kernel $\mathbf{G}$ can be expressed as
\begin{align}
	\mathbf{s}_{oi} &= \mathbf{V}(\hat{\mathbf{G}} \odot \hat{\mathbf{s}}_i)\notag\\
	&= \mathbf{V}\bigl(diag(\hat{\mathbf{G}}) \hat{\mathbf{s}}_i\bigr)\notag\\
	&=\mathbf{V}diag(\hat{\mathbf{G}}) \mathbf{V}^H\mathbf{s}_i,
\end{align}
it is convenient to design spectrum domain $\hat{\mathbf{G}}$ directly.

In order to preserve edges in the resampled point cloud, we use a 
highpass filter defined in spectrum domain.
 Here, we use a Haar-like highpass design i.e., $\hat{\mathbf{G}}=\mathbf{1}-\bm{\lambda}$, where $\bm{\lambda}=[\lambda_1\ \lambda_2 
 \cdots \lambda_{N_k}]^T$ are eigenvalues 
 corresponding 
 to eigenvectors \{$\mathbf{f}_1,\cdots,\mathbf{f}_{N_k}$\}, respectively. 
 We use the ratio between the norm of the convolution output 
 $\mathbf{s}_{oi}$ and the norm of 
 $\mathbf{s}_i - \mathbf{s}_{oi}$ to measure
 smoothness $\beta_i$ for the $i$-th point, i.e.,
\begin{align}\label{smooth1}
\beta_i=\frac{\Vert \mathbf{s}_{oi}\Vert}{\Vert \mathbf{s}_i - \mathbf{s}_{oi}\Vert}.
\end{align}
In resampling, we would like to extract distinct and
sharp features by selecting 
points that exhibit lower $\beta_i$ value from the
resampling output. Our algorithm is summarized as 
Algorithm \ref{alg:A}, also known as the HKC resampling
algorithm. 

\subsubsection{Complexity Analysis}
In an unorganized point cloud, the computational complexity order
for generating $\mathbf{s}_i$ of all points is ${O}(N^2)$.
This is because we would have to search the point cloud to 
find actual neighbors for each data point. 
To compute the estimated hypergraph spectrum basis $\mathbf{V}$ 
and $\mathbf{V}^H$ requires finding eigenvectors of {$R_{\mathbf{P'_c}}$} which has computational complexity 
of $O(N_k^3)$. The cost of computing all HGFT and iHGFT 
pairs equals $O(N N_k^2)$. The total complexity of computing $\beta_i$ in Eq.~(\ref{smooth1}) would be $O(N N_k)$ and it
further requires ${O}(N\log{N})$ steps to sort the $\beta_i$s. 
The computational complexity of the HKC algorithm 
amounts to $O(N^2+N\log{N}+ N_k(N_k+1)N+N_k^3)$. In an organized point cloud, the computational complexity of generating 
$\mathbf{s}_i$ for all points can be reduced to ${O}(N)$, 
such that the computational complexity of the HKC resampling
is only $O(N\log{N}+ (N_k^2+N_k+1)N+N_k^3)$.

\subsection{Hypergraph Kernel Filtering (HKF) Resampling}

To present method, we still use the $3\times 3\times 3$ cube as the {example} kernel, and the same local signal $\mathbf{s}_i$ in the Kernel convolution based method.

\subsubsection{HKF Resampling}
Consider convolution via Hadamard product and inverse Fourier
transform in the HKC resampling algorithm. 
We need to transform the local signal $\mathbf{s}_i$ from vertex 
domain to spectrum domain. 
A simpler alternative is to compute 
local smoothness directly for signals in spectrum domain. 
The computational complexity of the algorithm is 
reduced by eliminating the inverse transform.

For edge-preserving, we wish
to separate the high frequency coefficients from the 
low frequency coefficients in 
spectrum domain. Recall that the spectrum bases corresponding 
to smaller $\lambda$ represent higher frequency 
components
\cite{d6}. Thus, we sort the eigenvalues of $R_{\mathbf{s'}}$ 
as $0\leq \lambda_1\leq \lambda_2 \leq \ldots \leq \lambda_{N_k}$ with corresponding eigenvectors \{$\mathbf{f}_1,\cdots,\mathbf{f}_{N_k}$\}. 
We can devise a threshold $\theta$ to separate the high frequency components 
from the low frequency components according to
a sharp rise of successive eigenvalues. 

\begin{algorithm}[t]
	\caption{Hypergraph Kernel Filtering (HKF)}
	\label{alg:B}
	\begin{algorithmic}
		\STATE {\textbf{Input}: A point cloud with $N$ nodes characterized by $\mathbf{P}=[\mathbf{p}_1^T \cdots \mathbf{p}_N^T]^T$, resampling ratio $\alpha$}.
		\STATE {\textbf{1.}} {Calculate the intrinsic resolution of point cloud};
		\STATE {\textbf{2.}} {{Use the intrinsic resolution as $d$ to get the coordinates $\mathbf{P_c}\in\mathbb{R}^{N_k\times3}$ of voxel centers in the kernel}};
		\STATE {\textbf{3.}} {{Use the coordinates $\mathbf{P_c}$ of the voxel centers in the kernel to estimate the hypergraph spectrum basis $\mathbf{V}=[\mathbf{f}_1,\cdots,\mathbf{f}_{N_k}]$}};
		\FOR{$i=1,2,\cdots, N$}
		\STATE {\textbf{4.}} {Use hypergraph spectrum basis $\mathbf{V}$ to calculate the Fourier transform $\hat{\mathbf{s}}_i$ in Eq. (\ref{FT})};
		\STATE {\textbf{5.}} {Calculate the local smoothness $\sigma_i$ in Eq. (\ref{local smoothness})};
		\ENDFOR
		\STATE {\textbf{6.}} {Sort the local smoothness $\sigma_i$ and select the bottom $N_r=\alpha N$ points as the resampled point cloud}.
	\end{algorithmic}
\end{algorithm}

Given a threshold selection of $\theta$, we could further define a local smoothness $\sigma_i$ to select the resampled points:
\begin{align}\label{local smoothness}
{\sigma_i=\frac{\sum_{k \in \{1,2,\cdots, \theta\}}\vert \hat{\mathbf{s}}_i(k) \vert}{\sum_{k \in \{1,2, \cdots, {N_k}\}}\vert\hat{ \mathbf{s}}_i(k)\vert}},
\end{align}
which is the fraction of high frequency energy 
within total signal energy. 
\begin{figure*}[t]
	\centering
	\begin{subfigure}{.45\textwidth}
		\centering
		\includegraphics[width=1\linewidth]{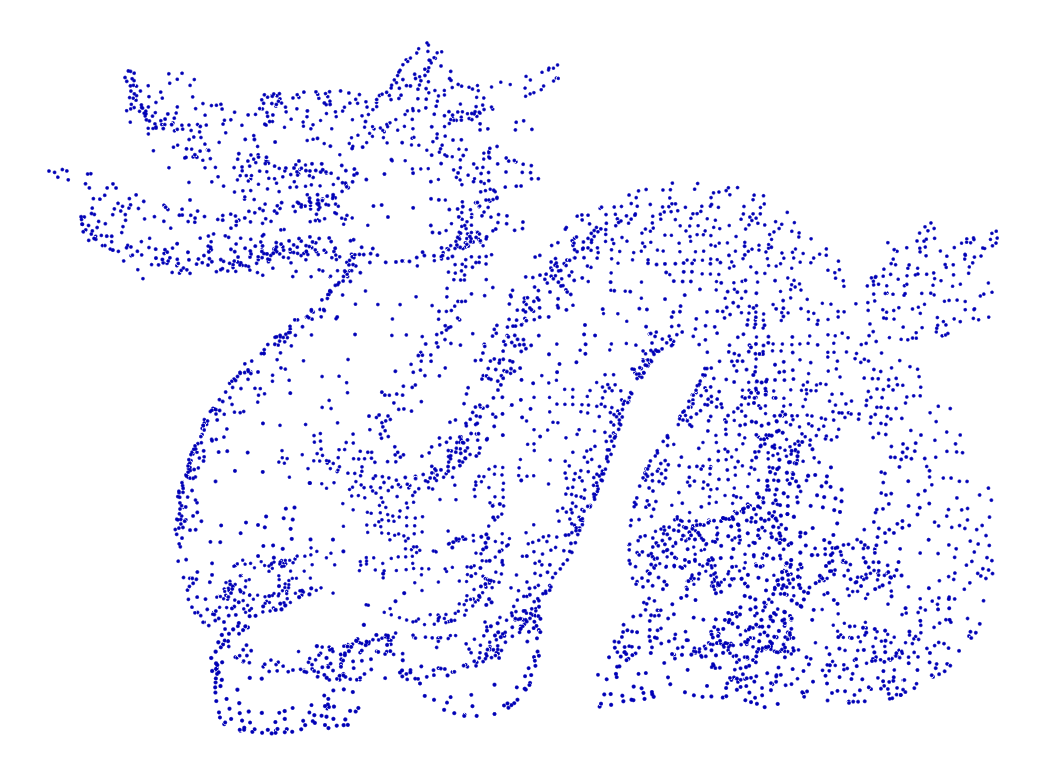}
		\caption{Resampled results of ``dragon'' using local hypergraph filtering based method with signal length $N_{i}=3$. Details in the body part are kept.}
		\label{Fig:dragon_s}
	\end{subfigure}
	\hspace{1cm}
	\begin{subfigure}{.45\textwidth}
		\centering
		\includegraphics[width=1\linewidth]{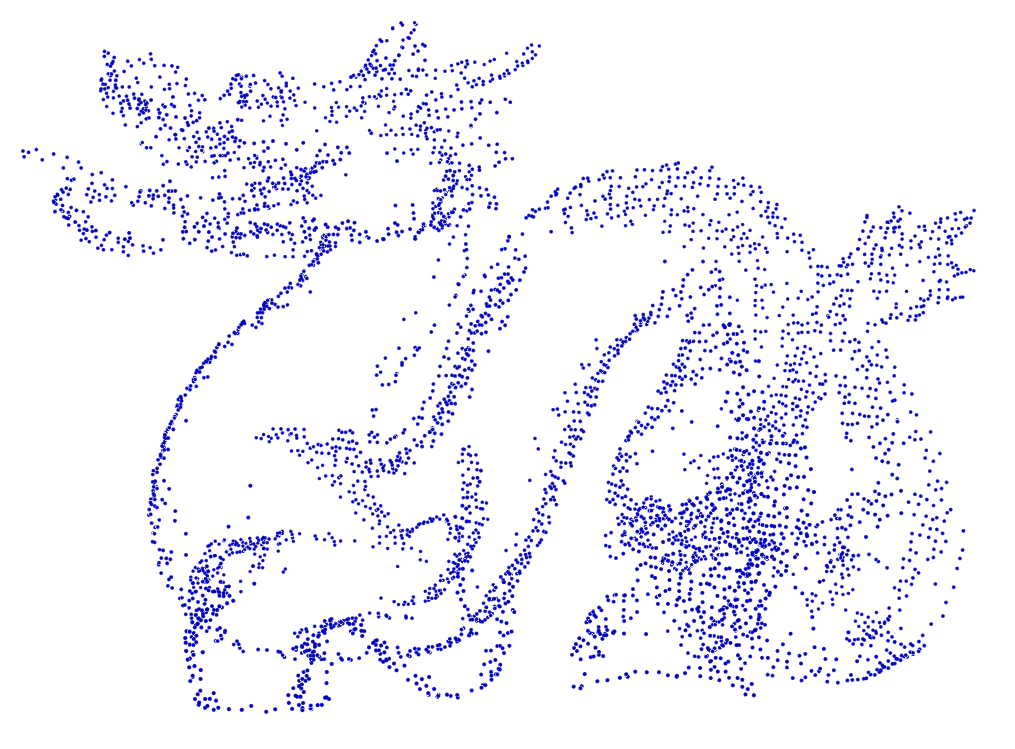}
		\caption{Resampled results of ``dragon'' using local hypergraph filtering based method with signal length $N_{i}=6$. Details in the body part are ignored.}
		\label{Fig:dragon_l}
	\end{subfigure}
	\caption{Resampled Results of Dragon Using Local Hypergraph Filtering based Method.}
	\label{Fig:Dragon}
\end{figure*}
Finally, we resample the point cloud by selecting 
the points with smaller $\sigma_i$, which 
correspond to points containing larger amount of 
higher-frequency components in the hypergraph. We summarize our algorithm as Algorithm \ref{alg:B}, also known as HKF resampling algorithm. 
Since the resampled point clouds favor high-frequency points, they
tend to contain more sharp features and are less smooth.

\subsubsection{HKF Algorithm Complexity}
Similar to HKC resampling, for an unorganized point cloud, 
the computational complexity for generating $\mathbf{s}_i$ for 
all points is ${O}(N^2)$. To estimate the hypergraph 
spectrum basis $\mathbf{V}$ and $\mathbf{V}^H$, the complexity is $O(N_k^3)$. 
The cost of computing requisite Fourier transforms amounts to
$O(N N_k^2)$. The total complexity of computing $\beta_i$ in Eq. (\ref{local smoothness}) is $O(N N_k)$. It further requires
${O}(N\log{N})$ to sort the $\{\sigma_i\}$. 
In summary, the computational complexity of Algorithm \ref{alg:B}
(HKF algorithm) is $O(N^2+N\log{N}+ N_k(N_k+1)N+N_k^3)$. 
For an organized point cloud, however,
the computational complexity for generating $\mathbf{s}_i$ 
for all points is reduced to ${O}(N)$, such
that the computational complexity of 
HKF is reduced to $O(N\log{N}+ (N_k^2+N_k+1)N+N_k^3)$.

\subsection{Local Hypergraph Filtering (LHF) Algorithm}

We further propose another novel HGSP
approach to model the local signal 
by incorporating the vector between 
the $i$-th point and each of its ($N_i-1$) closest neighbors.
In particular, we define a local signal for the $i$-th node
of order $N_i$ as
\begin{equation}\label{sig}
\mathbf{s}_{p,i}=[\mathbf{0}^T\ (\mathbf{p_{n_1}}-\mathbf{p_i})^T \cdots (\mathbf{p_{n_{N_i-1}}}- \mathbf{p_i})^T]^T\in \mathbb{R}^{N_i\times3},
\end{equation}
where $n_1,\cdots,n_{N_i-1}$ are the indices of its ($N_i-1$) 
nearest neighbors. 

We then build hypergraphs over these cloud points and use 
hyperedge to connect point $i$ and its ($N_i-1$) nearest neighbors. 
Similar to the HKF algorithm, 
we also devise a filter defined in spectrum domain to process 
the local signal. 
Although, strictly speaking, we could define a unique $N_i$ for 
each of the $i$-th point, we find it more convenient to
consider some fixed selections for all points to avoid comparing 
hyperedges of different {length}. 

Fig. \ref{Fig:Dragon} provides an example to show the effect of $N_i$ selection. When $N_i$ is small, $\mathbf{s}_{p,i}$ describes the local geometric information in a smaller region. Consequently, 
the filtered results tend to vary more and captures 
sharp features of the point cloud in Fig. \ref{Fig:dragon_s}.
When $N_i$ is large, $\mathbf{s}_{p,i}$ characterize the local geometry of a larger region around the $i$-th point. As a result, 
its local information is blurred with
other local information from its neighbors such that
the filtered results tend to be smoother and tend to highlight
the contour of the point cloud as seen from
Fig. \ref{Fig:dragon_l}.

\subsubsection{Local Hypergraph Filtering (LHF) Resampling}
Because we would like to preserve both the sharp features 
and the surface contour of the point cloud to achieve
consistently good performance across different point clouds,
we propose to apply several values of $N_i$ for
all points. 
In particular, we consider two different {lengths} $N_{a},N_{b}$ to construct
two different sets of local signals.
We would then integrate the filtered results.

Our local hypergraph filtering (LHF) based resampling consists 
of two main steps: i) hypergraph spectrum construction, ii) spectrum domain filtering.
We first estimate hypergraph spectrum by applying the same process used in
HKC and HKF algorithms. In
this new LHF method, each cloud point has its own (small-scale)
hypergraph. We should estimate the hypergraph spectrum for each point
using
two different local signals $\mathbf{s}_{p,i,a}\in \mathbb{R}^{N_{a}\times3}$ and $\mathbf{s}_{p,i,b}\in \mathbb{R}^{N_{b}\times3}$. 
Once the estimation of the corresponding hypergraph spectrum bases 
$\mathbf{V}_{i,a}$ and $\mathbf{V}_{i,b}$ is completed, 
we apply Eq. (\ref{FT}) to derive the Fourier
transform $\hat{\mathbf{s}}_{p,i,a}$ and $\hat{\mathbf{s}}_{p,i,b}$,
respectively. 

Similar to spectrum filter in the HKF method, we 
define two thresholds $\theta_{i,a}$ and $\theta_{i,b}$,
respectively, for $\hat{\mathbf{s}}_{p,i,a}$ and 
$\hat{\mathbf{s}}_{p,i,b}$. 
Two local sharpness metrics are
further defined as
\begin{subequations}
\begin{align}\label{Local Sharpness}
\gamma(\hat{\mathbf{s}}_{p,i,a})&=\frac{\sum_{j \in \{1,2,\cdots, 
\theta_{i,a}\}} \sum_{k=1}^{3}\vert \hat{\mathbf{s}}_{p,i,a}(j,k)\vert}{\sum_{j \in \{1,2,\cdots, N_i\}} \sum_{k=1}^{3}\vert \hat{\mathbf{s}}_{p,i,a}(j,k)\vert},\\
\gamma(\hat{\mathbf{s}}_{p,i,b})&=\frac{\sum_{j \in \{1,2,\cdots, 
\theta_{i,b}\}} \sum_{k=1}^{3}\vert \hat{\mathbf{s}}_{p,i,b}(j,k)\vert}{\sum_{j \in \{1,2,\cdots, N_i\}} \sum_{k=1}^{3}\vert \hat{\mathbf{s}}_{p,i,b}(j,k)\vert},\label{Local Sharpness_b}
\end{align}
\end{subequations}
where $\theta_{i,a}$ and $\theta_{i,b}$ correspond to
the thresholds for $\hat{\mathbf{s}}_{p,i,a}$ 
and $\hat{\mathbf{s}}_{p,i,b}$, respectively, with
$N_a$ and $N_b$ as the respective corresponding {lengths}.

Upon completion of sharpness evaluation, for each signal point,
we apply a weighted average of $\gamma(\hat{\mathbf{s}}_{p,i,a})$
and $\gamma(\hat{\mathbf{s}}_{p,i,b})$
to form a combined sharpness result 
{
\begin{align}\label{Weight_average}
\gamma_i = \epsilon \gamma(\hat{\mathbf{s}}_{p,i,a}) + (1-\epsilon) \gamma(\hat{\mathbf{s}}_{p,i,b}),
\end{align}
where $\epsilon$ denotes the weight.}

{To balance the effect of two local sharpness metrics, we sort both $\gamma(\hat{\mathbf{s}}_{p,i,a})$ and $\gamma(\hat{\mathbf{s}}_{p,i,b})$ and design the weight $\epsilon$ according to the top $\alpha$ fraction of $\gamma(\hat{\mathbf{s}}_{p,i,a})$ and $\gamma(\hat{\mathbf{s}}_{p,i,b})$, denoted by $\Gamma_{a}$ and $\Gamma_{b}$, respectively. We can define node-specific weights}
\begin{align}\label{Weight}
\epsilon = \frac{\Gamma_{b}}{\Gamma_{a}+\Gamma_{b}}.
\end{align}
Finally, we sort the $\gamma_i$ of Eq. (\ref{Weight_average}) and select the top $N_\alpha=\alpha N$ points as the resampled point cloud. 
The whole algorithm is summarized as Algorithm $\ref{alg:C}$, also known as the
LHF algorithm. 

\subsubsection{LHF Algorithm Complexity}
Given an unorganized point cloud, the computational complexity for generating $\mathbf{s}_{p,i}$ for all points is ${O}(N^2)$. 
{To estimate the hypergraph spectrum bases $\mathbf{V}_{i,a}$s and $\mathbf{V}_{i,b}$s}, the computational complexity will be $O(N N_k^3)$. The cost of computing all the Fourier transforms and inverse Fourier transforms 
amounts to $O(N N_k^2)$. The total complexity of computing $\gamma_i$ is $O(N N_k)$. It further requires ${O}(N\log{N})$ 
computations to sort all $\gamma_i$. The computational complexity of the LHF Algorithm is $O(N^2+N\log{N}+ N_k(N_k^2+N_k+1)N)$. Given an organized point cloud, the computational complexity of generating $\mathbf{s}_{p,i}$ for all points is reduced to ${O}(N)$. Thus, the overall computational complexity of LHF algorithm reduces to $O(N\log{N}+ (N_k^3+N_k^2+N_k+1)N)$.

\begin{algorithm}[t]
	\caption{Local Hypergraph Filtering (LHF) Resampling}
	\label{alg:C}
	\begin{algorithmic}
		\STATE {\textbf{Input}: A point cloud with $N$ nodes characterized by  $\mathbf{P}=[\mathbf{p}_1^T \cdots \mathbf{p}_N^T]^T$, resampling ratio $\alpha$}, local lengths {$N_{a},N_{b}$.}
		\FOR{$i=1,2,\cdots, N$}
		\STATE {\textbf{1.}} {Find the nearest {($N_{a}-1$) and ($N_{b}-1$)} neighbors of point $i$};
		\STATE {\textbf{2.}} {Use coordinates of point $i$ and its {($N_{a}-1$) and ($N_{b}-1$)} neighbors to estimate the hypergraph spectrum bases {$\mathbf{V}_{i,a}$, $\mathbf{V}_{i,b}$}, respectively};
		\STATE {\textbf{3.}} {Use hypergraph spectrum basis {$\mathbf{V}_{i,a}$ and $\mathbf{V}_{i,b}$} to calculate the Fourier transform {$\hat{\mathbf{s}}_{p,i,a}$ and $\hat{\mathbf{s}}_{p,i,b}$}, respectively};
		\STATE {\textbf{4.}} {Calculate the local sharpness {$\gamma(\hat{\mathbf{s}}_{p,i,a})$ and $\gamma(\hat{\mathbf{s}}_{p,i,b})$} in Eq. (\ref{Local Sharpness}) and Eq. (\ref{Local Sharpness_b})};
		\ENDFOR
		\STATE {\textbf{5.}} {Calculate the weighted average of {$\gamma_i(\hat{\mathbf{s}}_{p,i,a})$ and $\gamma_i(\hat{\mathbf{s}}_{p,i,b})$} using Eq. (\ref{Weight_average}) and Eq. (\ref{Weight})}.
		\STATE {\textbf{6.}} {Sort the local sharpness $\gamma_i$ and select the top $N_r=\alpha N$ points as the resampled point cloud}.
	\end{algorithmic}
\end{algorithm}	

\section{Experimental results}

\begin{table*}[t]
	\centering
	\begin{tabular}{|c||c|c|c|c||c|c|c|c|}
		\hline
		& \multicolumn{4}{|c||}{HKC} & \multicolumn{4}{|c|}{HKF} \\
		\hline
		Noise Level & Precision & Recall & F1-Score & Mean Distance & Precision & Recall & F1-Score & Mean Distance \\
		\hline
		No noise & 0.3810 & 0.9957 & 0.5497 & 0.6285 & 0.3652 & 0.9585 & 0.5276 & 0.6442\\
		\hline
		10\% & 0.3801 & 0.9934 & 0.5485 & 0.7322 & 0.3501 & 0.9199 & 0.5060 & 1.1475\\
		\hline
		20\% & 0.2560 & 0.6709 & 0.3697 & 2.5501 & 0.1499 & 0.3913 & 0.2163 & 3.7606\\
		\hline
	\end{tabular}
	
	\begin{tabular}{|c||c|c|c|c||c|c|c|c|}
		\hline
		& \multicolumn{4}{|c||}{LHF} & \multicolumn{4}{|c|}{GFR}\\
		\hline
		Noise Level & Precision & Recall & F1-Score & Mean Distance & Precision & Recall & F1-Score & Mean Distance\\
		\hline
		No noise & 0.3578 & 0.9371 & 0.5166 & 1.0376 & 0.3827 & 1 & 0.5522 & 0.4516\\
		\hline
		10\% & 0.1437 & 0.3763 & 0.2075 & 2.4587 & 0.3818 & 0.9978 & 0.5509 & 2.9787\\
		\hline
		20\% & 0.0899 & 0.2358 & 0.1299 & 3.4143 & 0.2312 & 0.6050 & 0.3337 & 3.9326 \\
		\hline
	\end{tabular}
	
	\begin{tabular}{|c||c|c|c|c||c|c|c|c|}
		\hline
		& \multicolumn{4}{|c||}{EA} & \multicolumn{4}{|c|}{PCA-AC}\\
		\hline
		Noise Level & Precision & Recall & F1-Score & Mean Distance & Precision & Recall & F1-Score & Mean Distance\\
		\hline
		No noise & 0.3827 & 1 & 0.5522 & 1.0393 & 0.3752 & 0.9825 & 0.5418 & 0.8292\\
		\hline
		10\% & 0.3785 & 0.9902 & 0.5463 & 0.9792 & 0.3539 & 0.9318 & 0.5117 & 0.7424\\
		\hline
		20\% & 0.3597 & 0.9421 & 0.5193 & 1.2916 & 0.3418 & 0.8988 & 0.4941 & 0.9367\\
		\hline
	\end{tabular}
	\caption{Numerical results of methods for edge preserving resampling without and with Gaussian noise.}
	\label{table:1}
\end{table*}

We now describe our experiment setup and
present test results of the three proposed
new resampling algorithms. 

\subsection{Edge Preservation of Simple Synthetic Point Clouds}

\begin{figure}
	\centering
	\includegraphics[width=4.5in,trim=+11cm 0 0 0]{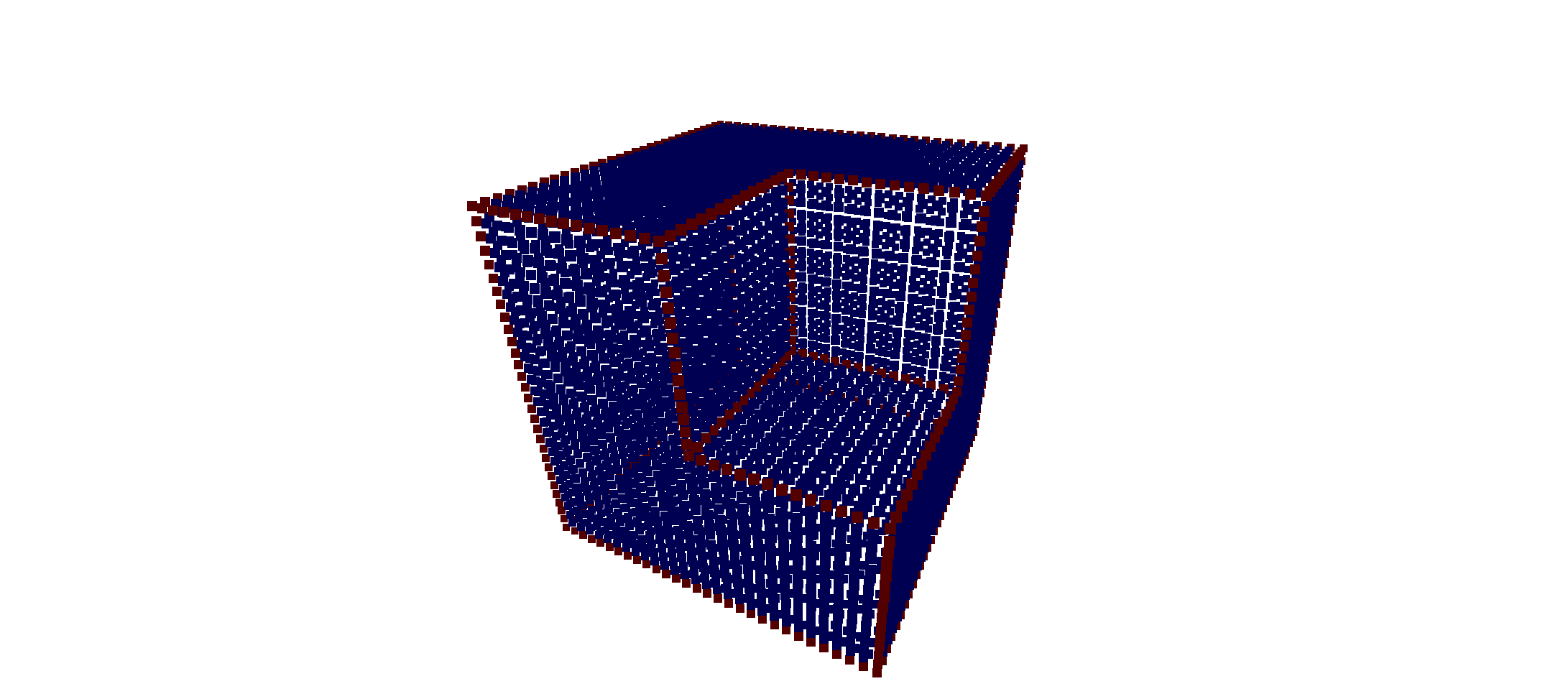}
	\caption{A Synthetic Point Cloud with Labeled Edge.}
	\label{Fig:1}
\end{figure}

As we described in Section III, one important resampling objective
is to preserve sharp features in a point cloud such as edges and corners.
In this part, we study the edge preserving capability of our proposed 
algorithms by testing over several simple synthetic point clouds.
The reason for selecting synthetic point clouds in
this test is 
to take advantage of the known ground truth regarding edges
and our ability to label them.  
We generate
these synthetic point 
clouds by uniformly sampling the external surface of models 
constructed from combinations of cubes of various sizes. 
One example of synthetic point clouds is shown in {Fig. \ref{Fig:1}}, 
where the points on edges are marked in red while the
remaining points are in blue.

To measure the accuracy of the preserved edges, 
we evaluate the $F_1$ score, defined by
\begin{align}\label{F1}
F_1 = 2\cdot \frac{\rm{Precision} \cdot \rm{Recall}}{\rm{Precision} + \rm{Recall}},
\end{align}
where precision denotes the fraction of edge points correctly 
preserved among all (false or correct)
edge points for a resampling algorithm 
while the recall is the ratio of correctly preserved
edge points versus all ground truth edge points. 
We also calculate the mean distance to their closest ground
truth edge point respectively 
to show the ability of the algorithms in
capturing the model surface.

For baseline comparison, we also test the 
graph-based fast resampling (GFR) method 
of \cite{c3} on the same datasets.
We also compare with an edge detection method based on
eigenvalues analysis (EA) and another edge detection algorithm
based on Principal Component Analysis (PCA) and agglomerative clustering 
(PCA-AC) \cite{c5}. 
The parameters of GFR method are set to 
the typical values suggested by \cite{c3}. For 
EA and PCA-AC, we retain the points with higher cluster numbers or 
larger surface variation in the resampled point cloud to yield the same resampling ratio. 
{Here}, we set the resampling ratio $\alpha=0.2$ for all point clouds 
as an example. 
additional results with different resampling ratios will be further 
presented in Section IV-C. 
In order to study the robustness of algorithms, 
we also add 10\% and 20\% of Normal measurement
noises to the coordinates of point cloud, respectively. 
{Table \ref{table:1}} summarizes  our test results.

Compared with the GSP based GFR method, 
the newly proposed hypergraph kernel convolution 
(HKC) algorithm performs robustly for point clouds under
larger measurement noises. 
Since the local signals in HKC are defined by 
the number of points in the voxel of kernel, weaker
noises on a single point with perturbation 
below $d$/2 (the intrinsic resolution of 
original point clouds) would not affect local signals.

On the other hand, LHF algorithm does not perform well 
against noisy point clouds because sizable noises directly
distort the local hypergraph and neighbors.
Overall, our proposed HGSP-based methods 
demonstrate stronger
robustness than the traditional graph-based GFR algorithm 
for noisy data. Using a generic signal processing approach
they also deliver competitive performance
against non-graph based EA and PCA-AC methods
that were designed specifically for edge detection. 

\begin{figure*}[t]
	\begin{subfigure}{.23\textwidth}
		\begin{minipage}{\linewidth}
			\includegraphics[width=1\linewidth]{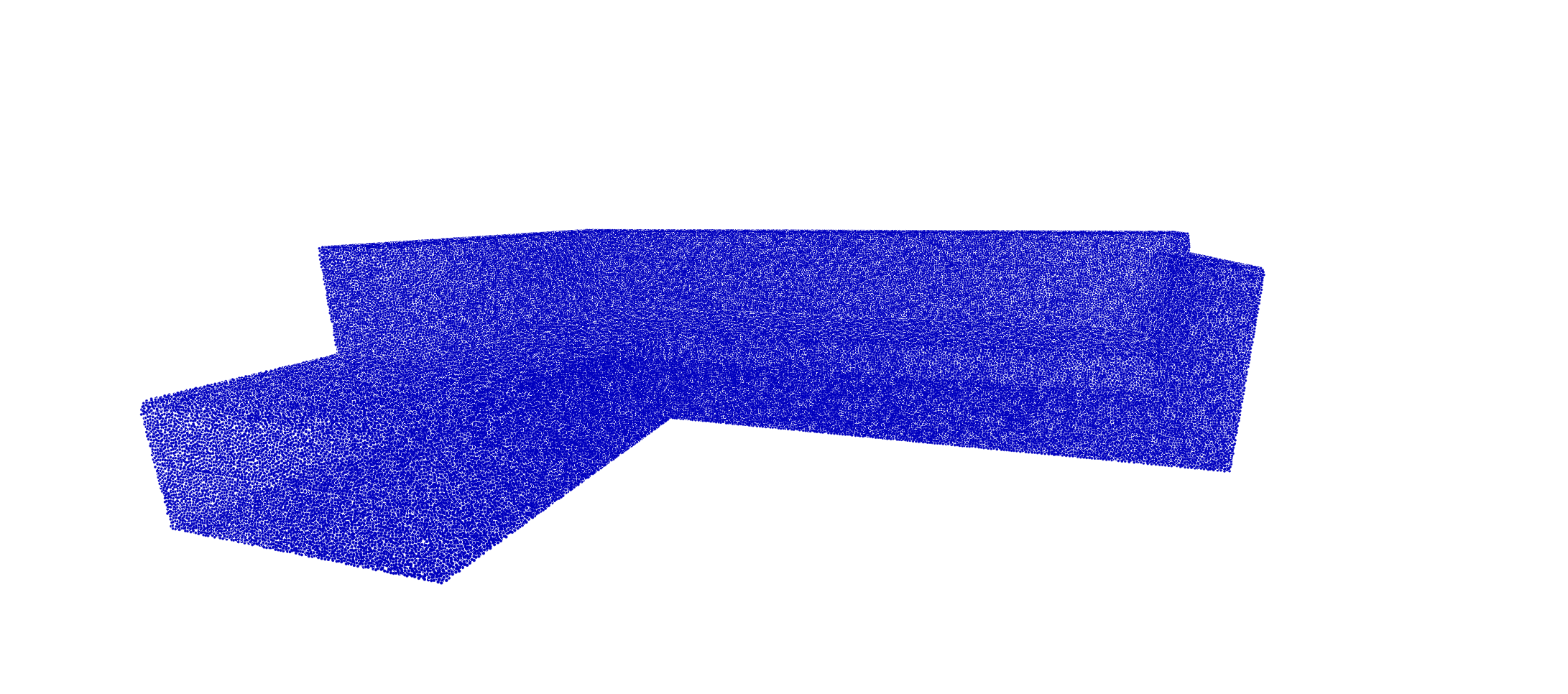}
			\includegraphics[width=1\linewidth]{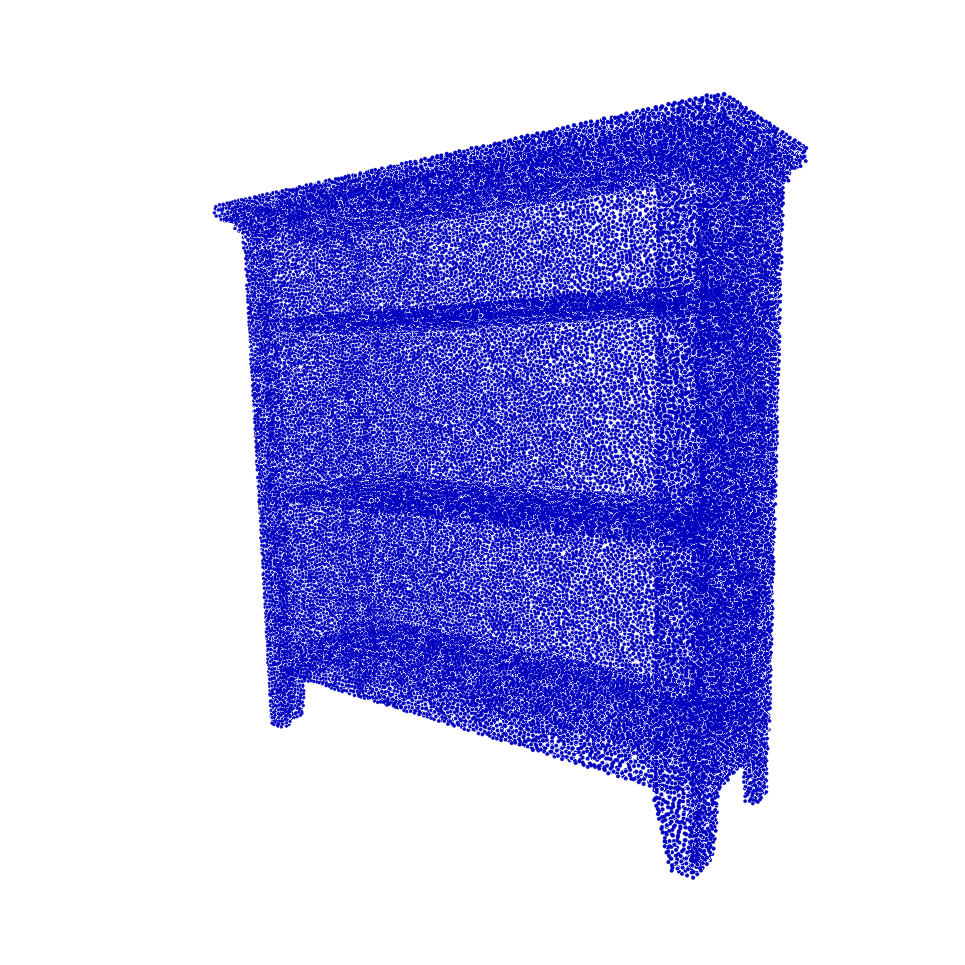}
			\includegraphics[width=1\linewidth]{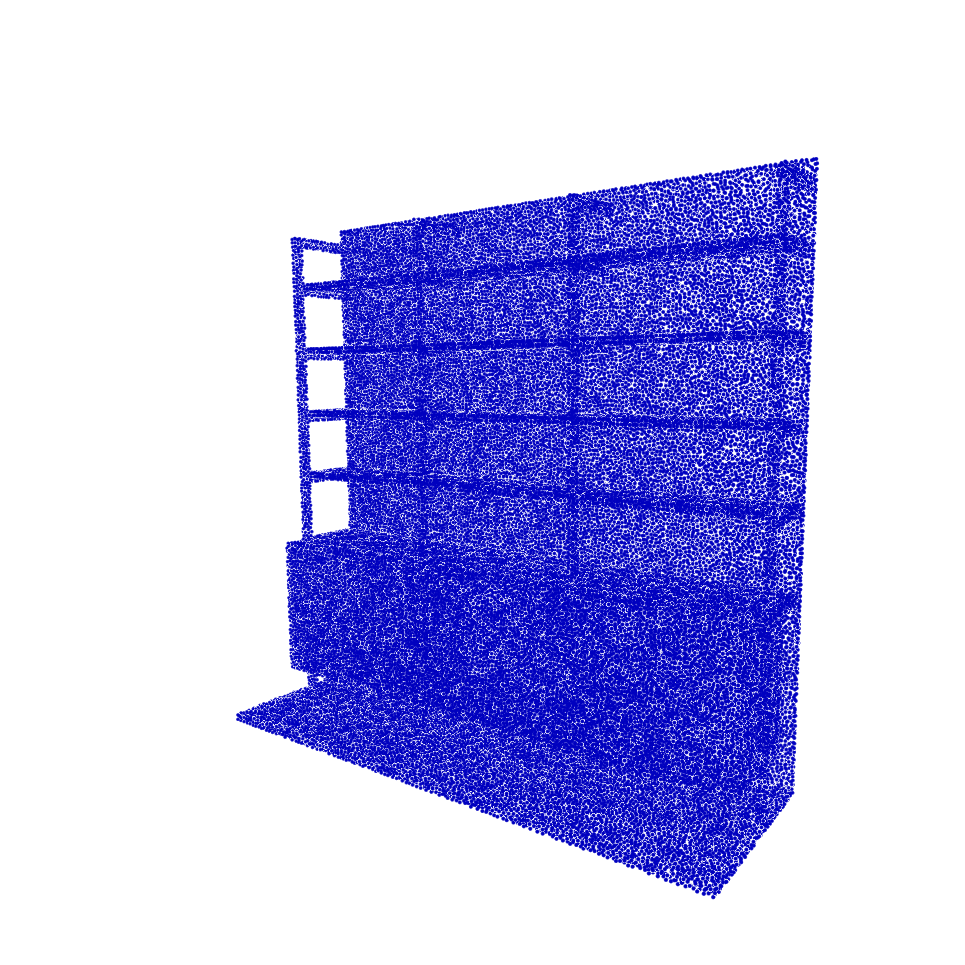}
		\end{minipage}
		\caption{Original Point Clouds}
	\end{subfigure}
	\begin{subfigure}{.23\textwidth}
		\begin{minipage}{\linewidth}
			\includegraphics[width=1\linewidth]{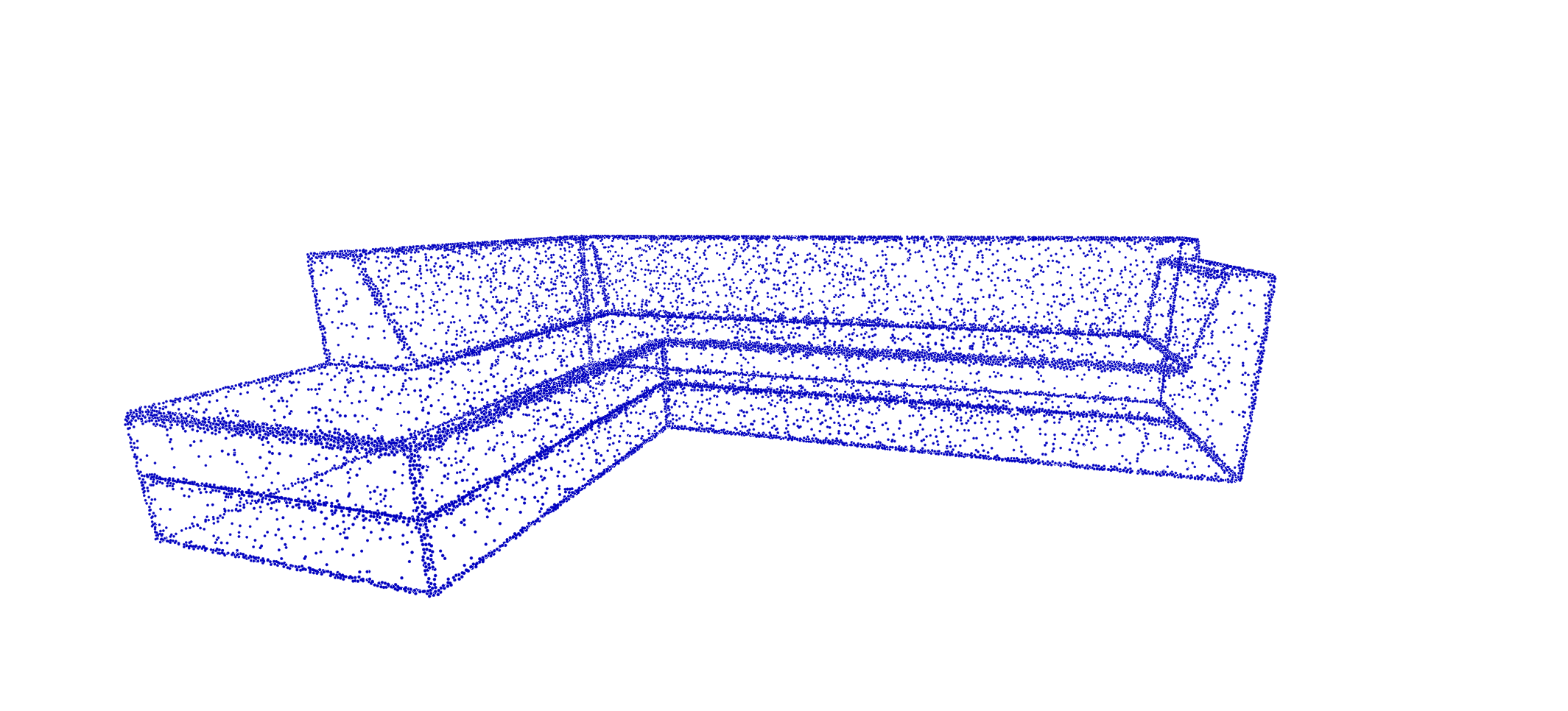}
			\includegraphics[width=1\linewidth]{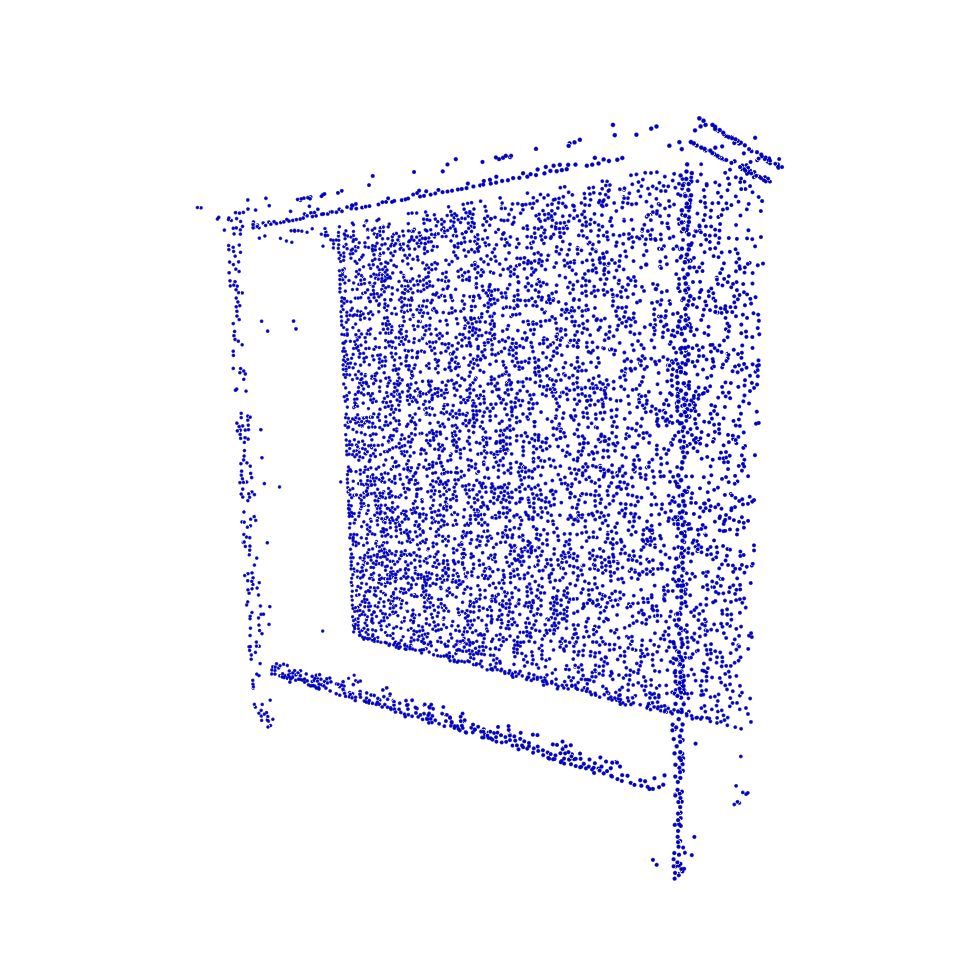}
			\includegraphics[width=1\linewidth]{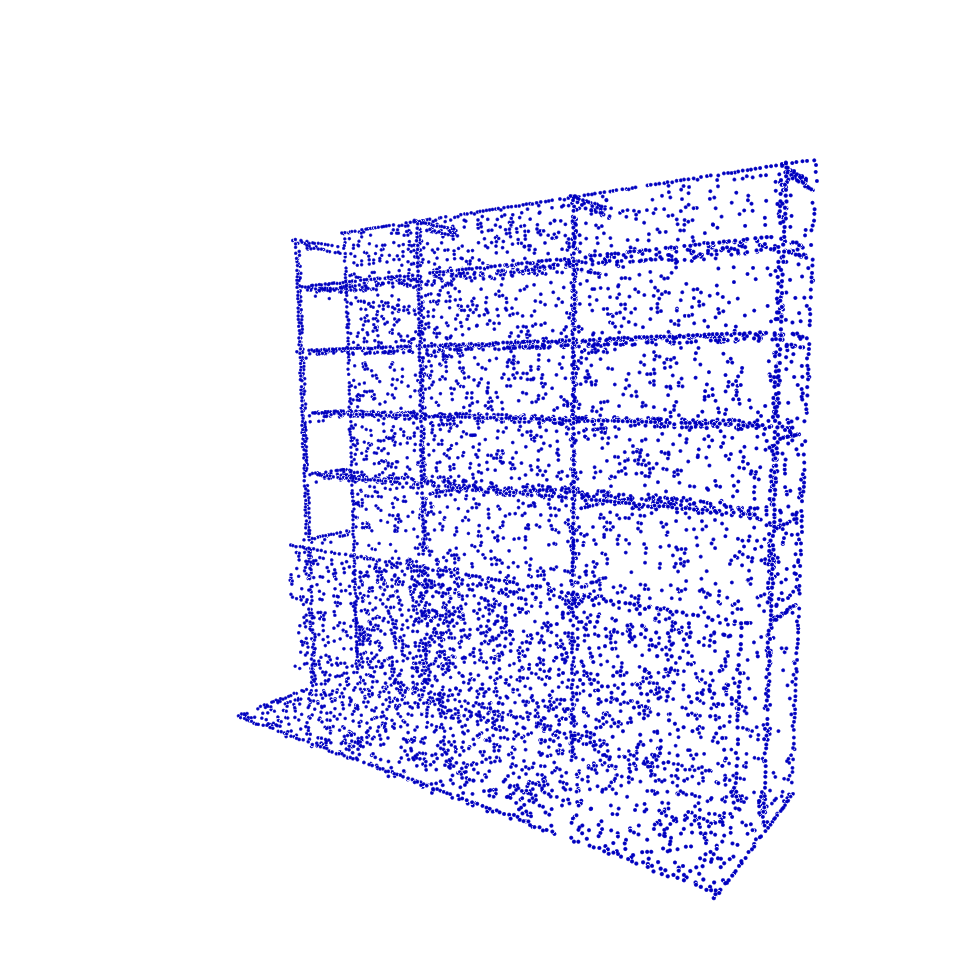}
		\end{minipage}
		\caption{Hypergraph Convolution}
	\end{subfigure}
	\begin{subfigure}{.23\textwidth}
		\begin{minipage}{\linewidth}
			\includegraphics[width=1\linewidth]{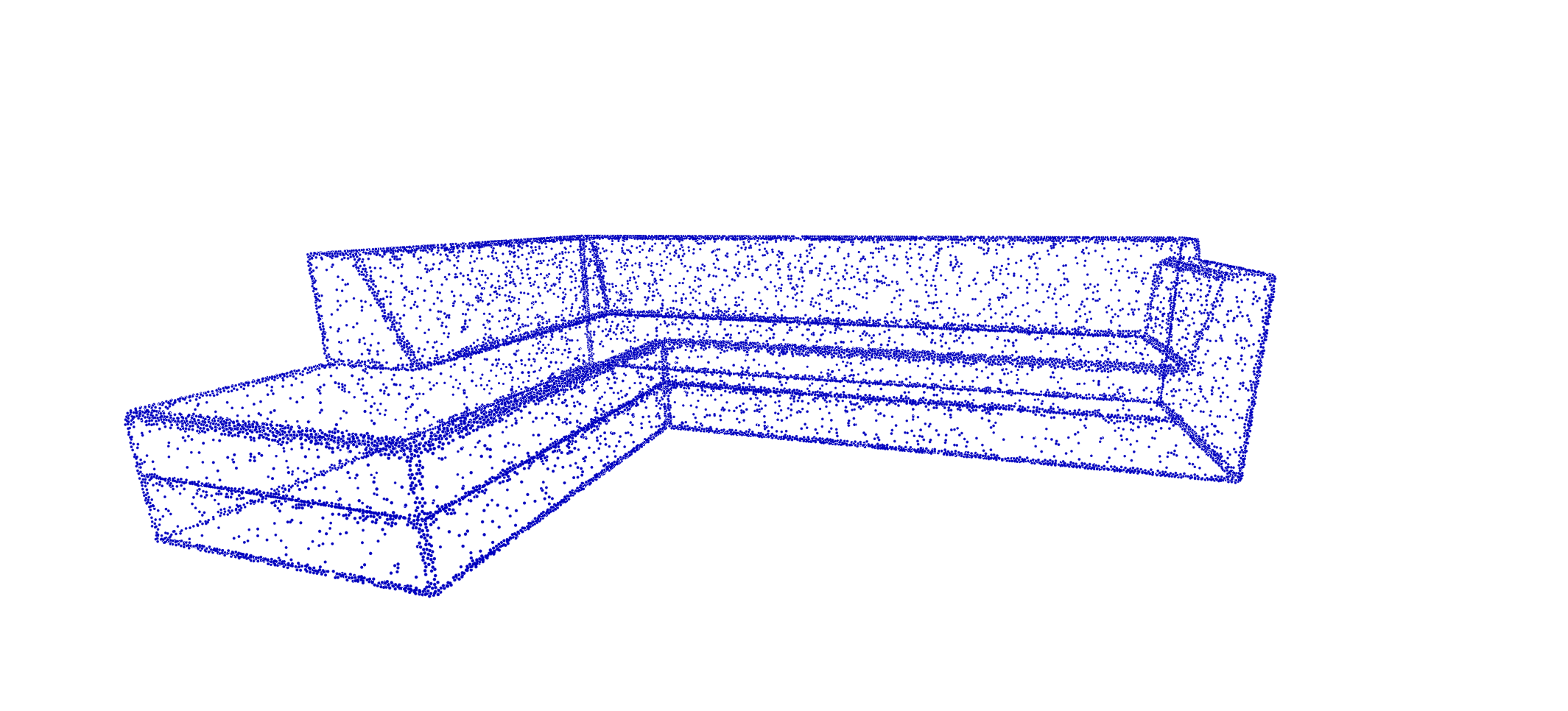}
			\includegraphics[width=1\linewidth]{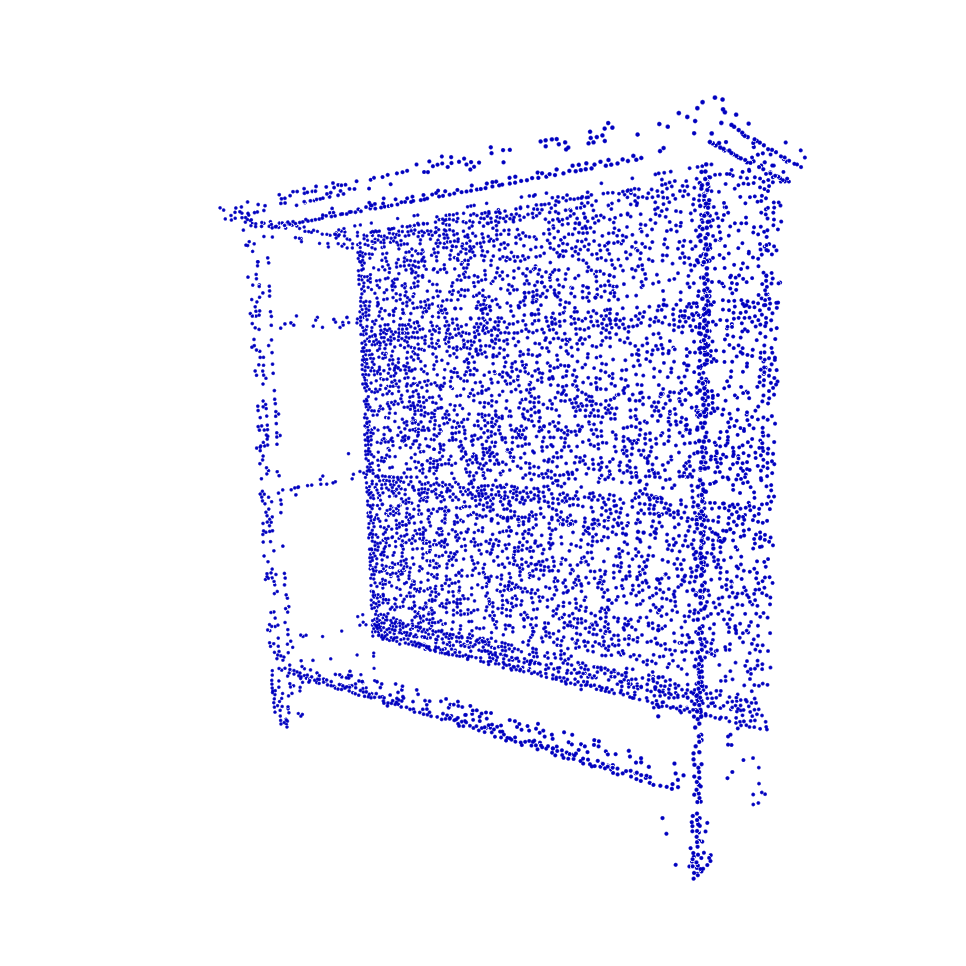}
			\includegraphics[width=1\linewidth]{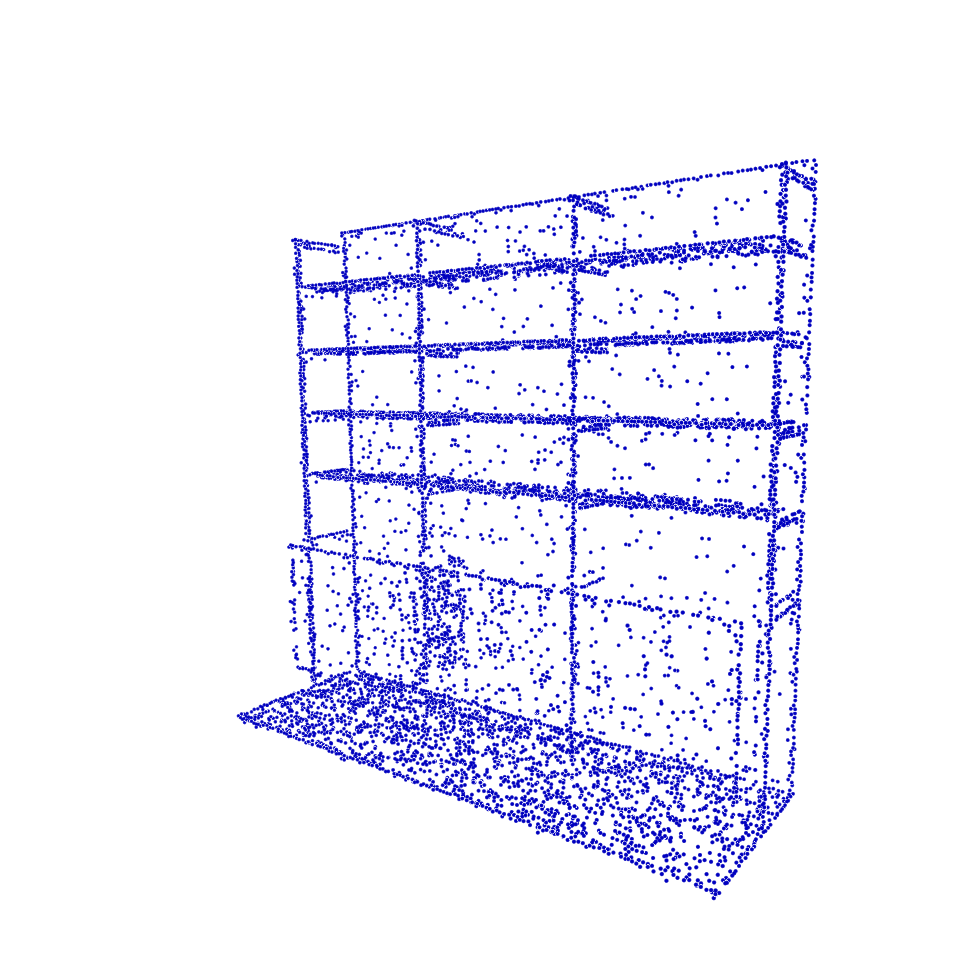}
		\end{minipage}
		\caption{Hypergraph Filtering}
	\end{subfigure}
	\begin{subfigure}{.23\textwidth}
		\begin{minipage}{\linewidth}
			\includegraphics[width=1\linewidth]{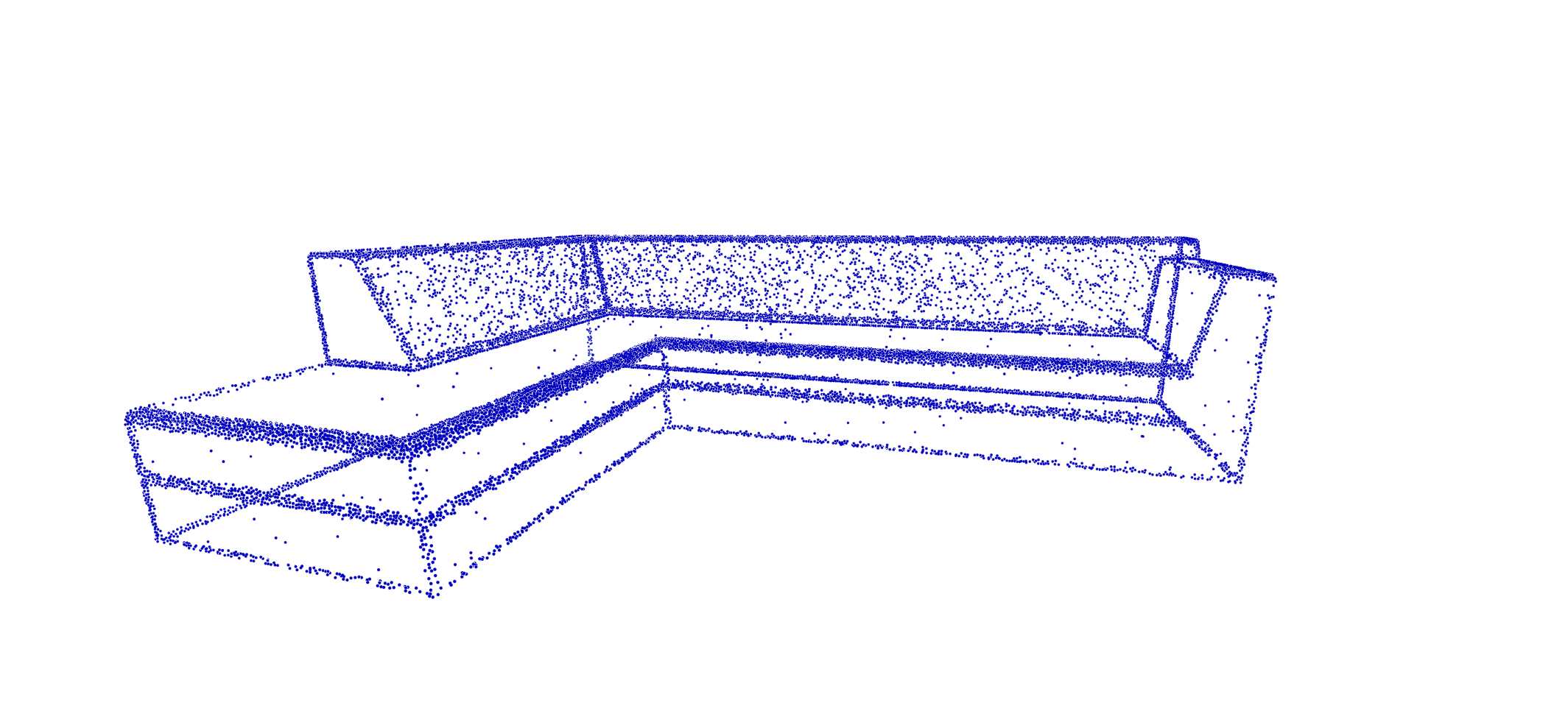}
			\includegraphics[width=1\linewidth]{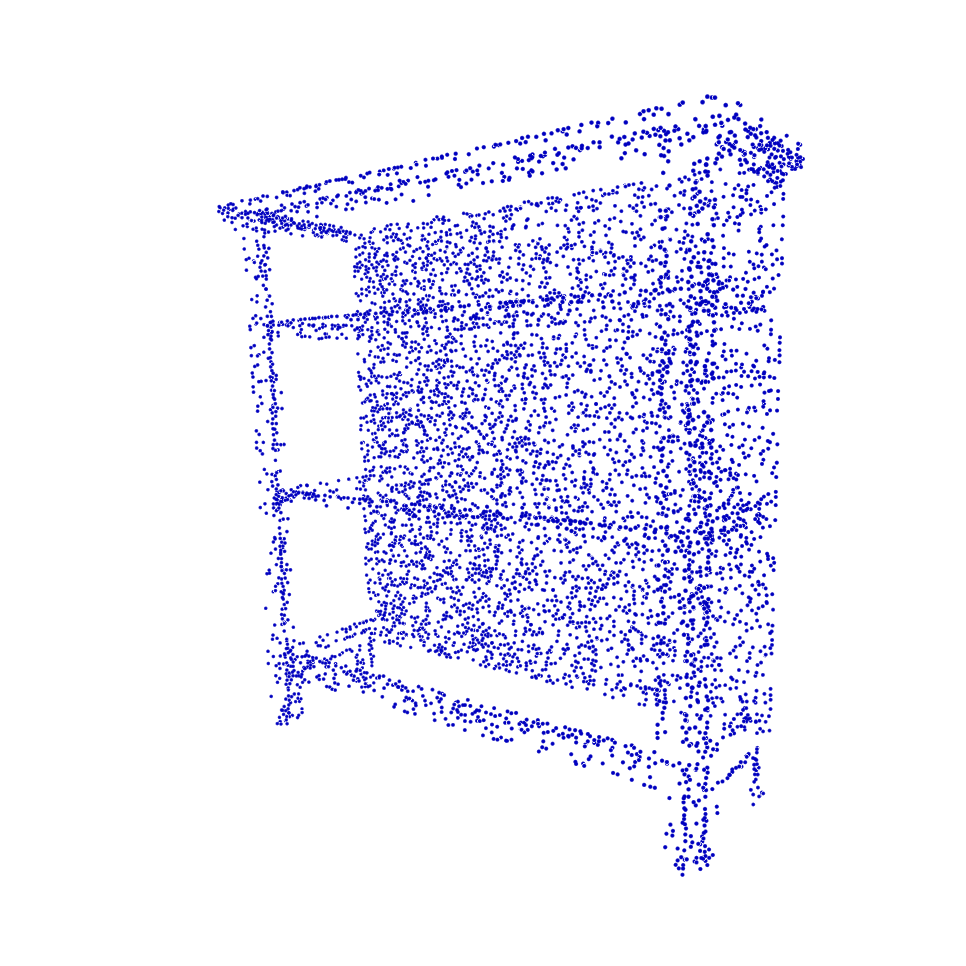}
			\includegraphics[width=1\linewidth]{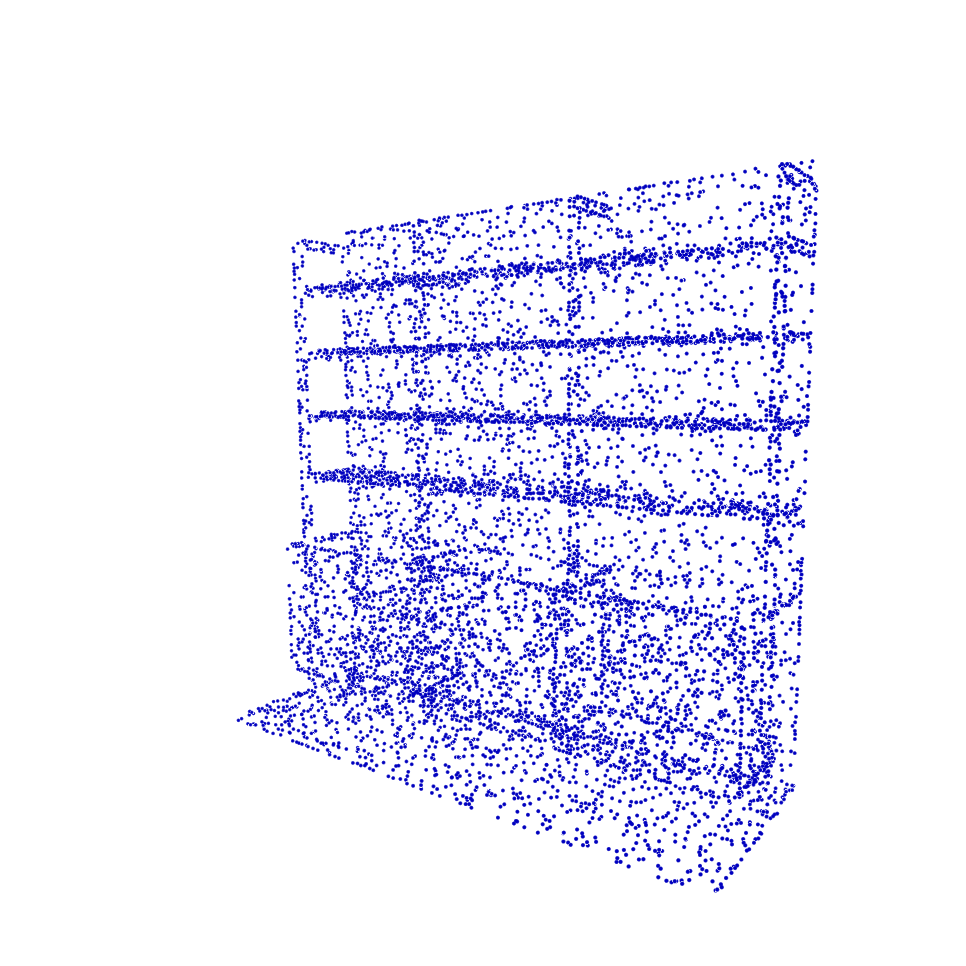}
		\end{minipage}
		\caption{Local Hypergraph Filtering}
	\end{subfigure}
	\caption{Examples of Edge Detection for Real Point Cloud.}
	\label{fig:Edge}
\end{figure*}

\subsection{Edge Preservation Results on Real-Life Point Clouds}
To test our proposed algorithm in a more general setting, we 
also implement edge-detection based on our resampled data 
in more complex practical point clouds. 
For these datasets, there is no explicit ground truth edges 
to provide quantitative results. 
Therefore, we present these results as
visible point cloud pictures to illustrate
the test performance in Fig. \ref{fig:Edge},
where the left column shows original point clouds and the 
right columns are the three
resampled point clouds for our proposed methods,
respectively.
These results show that our resampling method effectively
recover 3D object contours (outlines). 

\begin{figure}[htb]
	\centering
	\begin{subfigure}{.28\textwidth}
		\centering
		\includegraphics[width=1\linewidth]{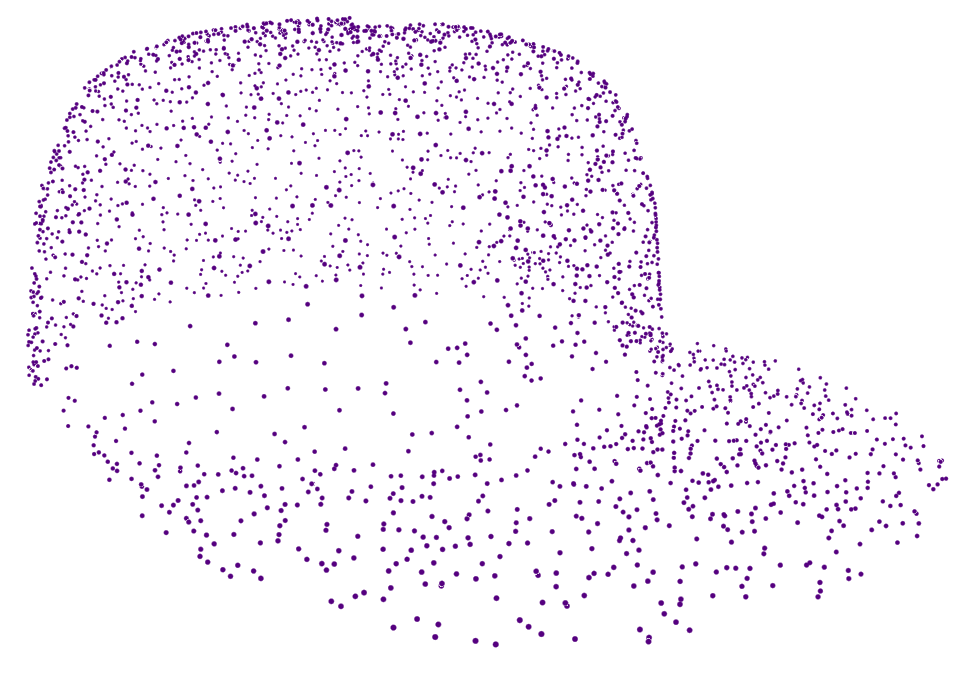}
		\caption{Original Point Cloud of a Cap.}
		\label{fig:originalpt}
	\end{subfigure}
\hfill
	\begin{subfigure}{.28\textwidth}
		\centering
		\includegraphics[width=1\linewidth]{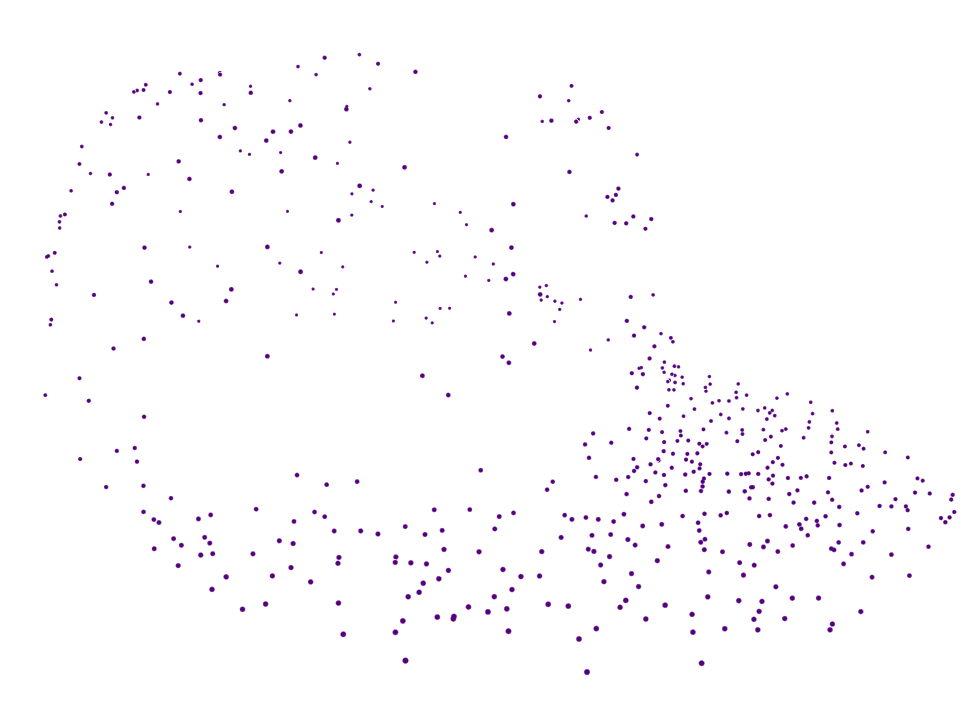}
		\caption{Resampled Result with $\alpha=0.2$.}
		\label{fig:downsamplept}
	\end{subfigure}
\hfill
	\begin{subfigure}{.28\textwidth}
		\centering
		\includegraphics[width=1\linewidth]{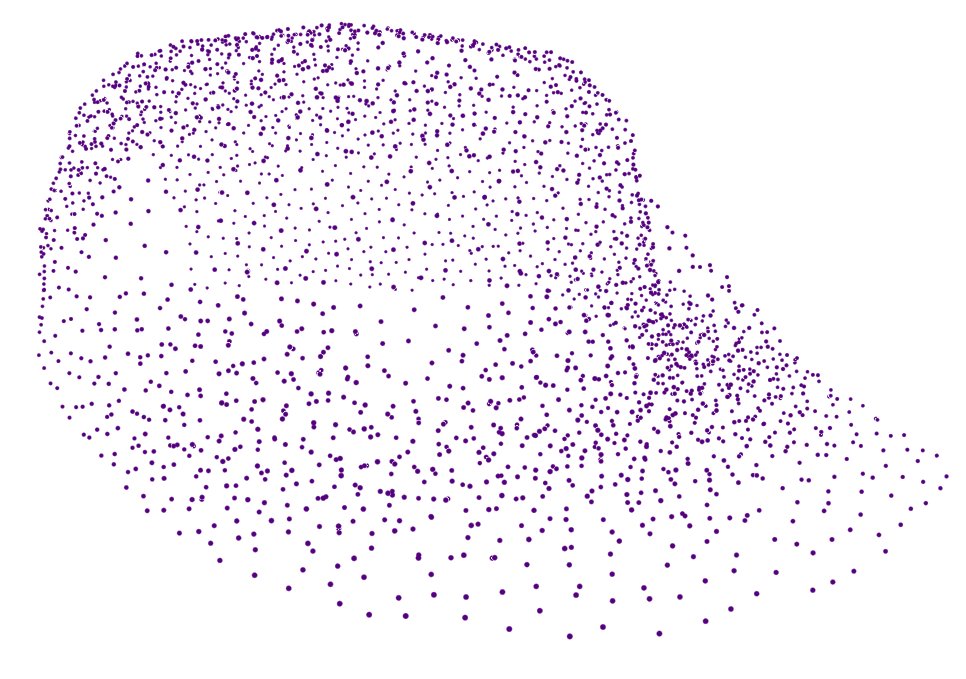}
		\caption{Recovered Point Cloud.}
		\label{fig:recoverpt}
	\end{subfigure}
	\caption{Example of original point cloud, resampled point cloud and the recovered point cloud.		\label{fig:examplehat}}
	\vspace*{-3mm}
\end{figure}

\subsection{Point Cloud Recovery from Resampling}

In the next test, we investigate the new algorithms' 
ability to preserve
high degree of point cloud information after resampling. 
In particular, we shall attempt to 
recover the dense point cloud after resampling
and assess the similarity between the
original point cloud and the recovered point cloud from
resampling. 

\begin{figure*}[htp]
	\centering
	\begin{subtable}{0.9\textwidth}
		\begin{tabular}{|c|c|c|c|c|c|c|}
			\hline
			Original & \tabincell{c}{HKC} & \tabincell{c}{HKF} & \tabincell{c}{LHF} & \tabincell{c}{GFR} & \tabincell{c}{EA} & \tabincell{c}{PCA-AC}\\
			\hline		
			\begin{minipage}[b]{.11\textwidth}	
				\vspace{0.1mm}
				\centering
				\includegraphics[height=\linewidth]{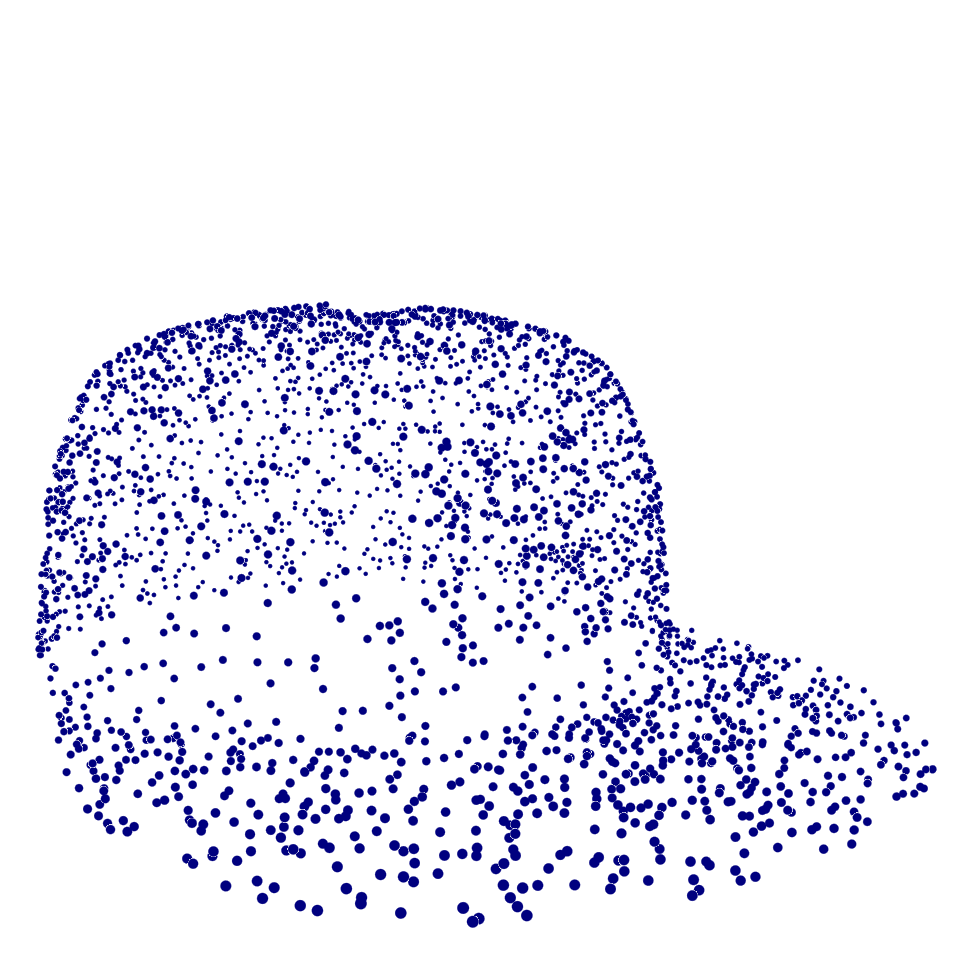}
			\end{minipage}&
			\begin{minipage}[b]{.11\textwidth}
				\centering
				\includegraphics[height=\linewidth]{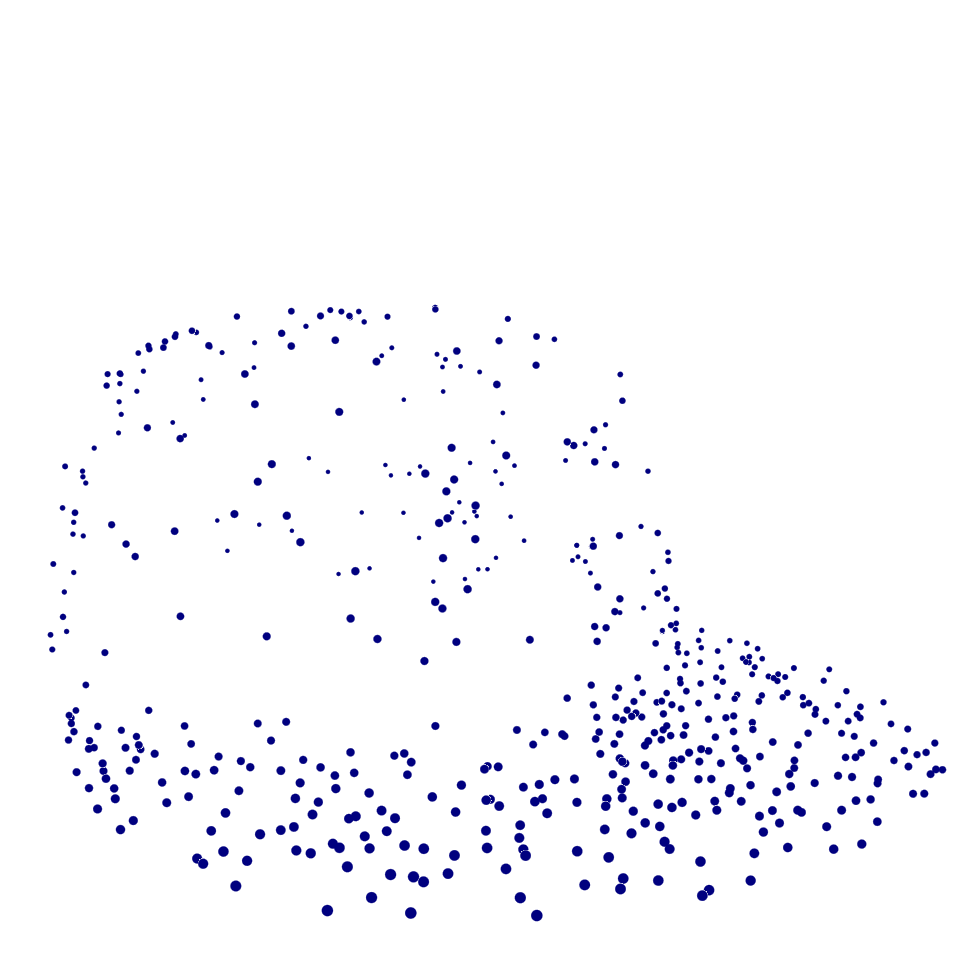}
			\end{minipage} &
			\begin{minipage}[b]{.11\textwidth}
				\centering
				\includegraphics[height=\linewidth]{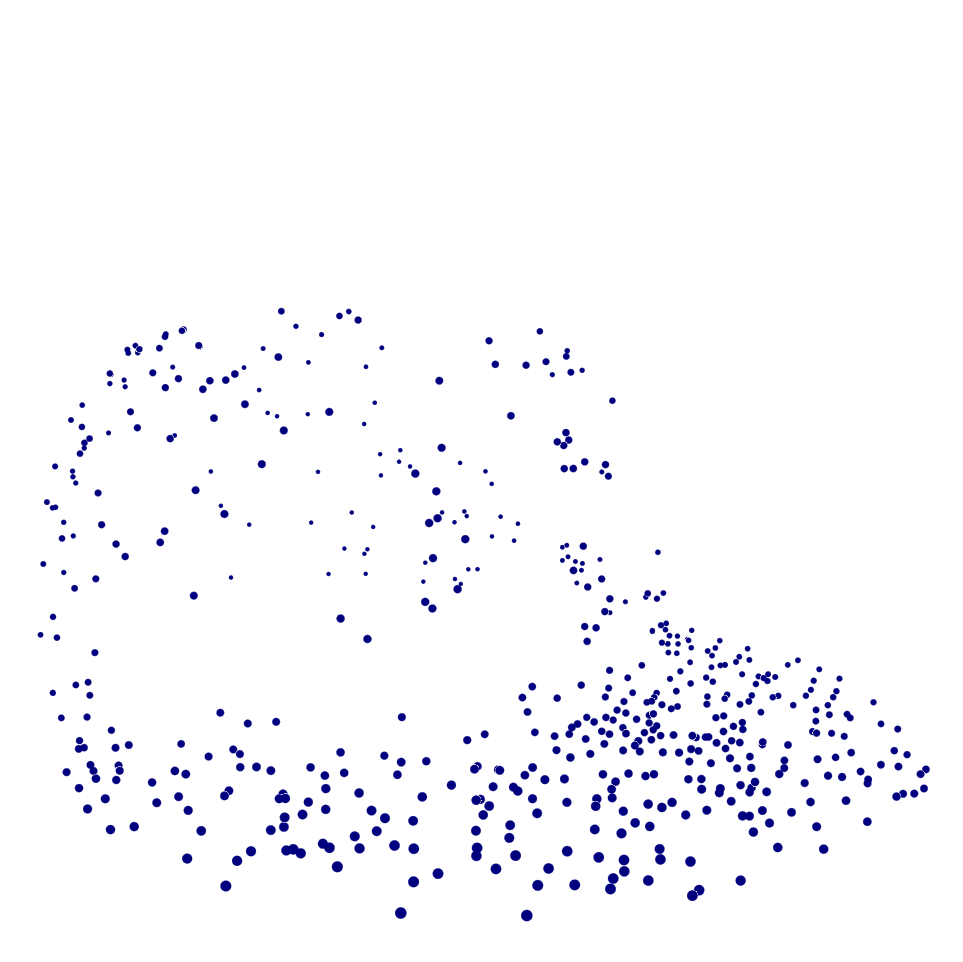}
			\end{minipage} &
			\begin{minipage}[b]{.11\textwidth}
				\centering
				\includegraphics[height=\linewidth]{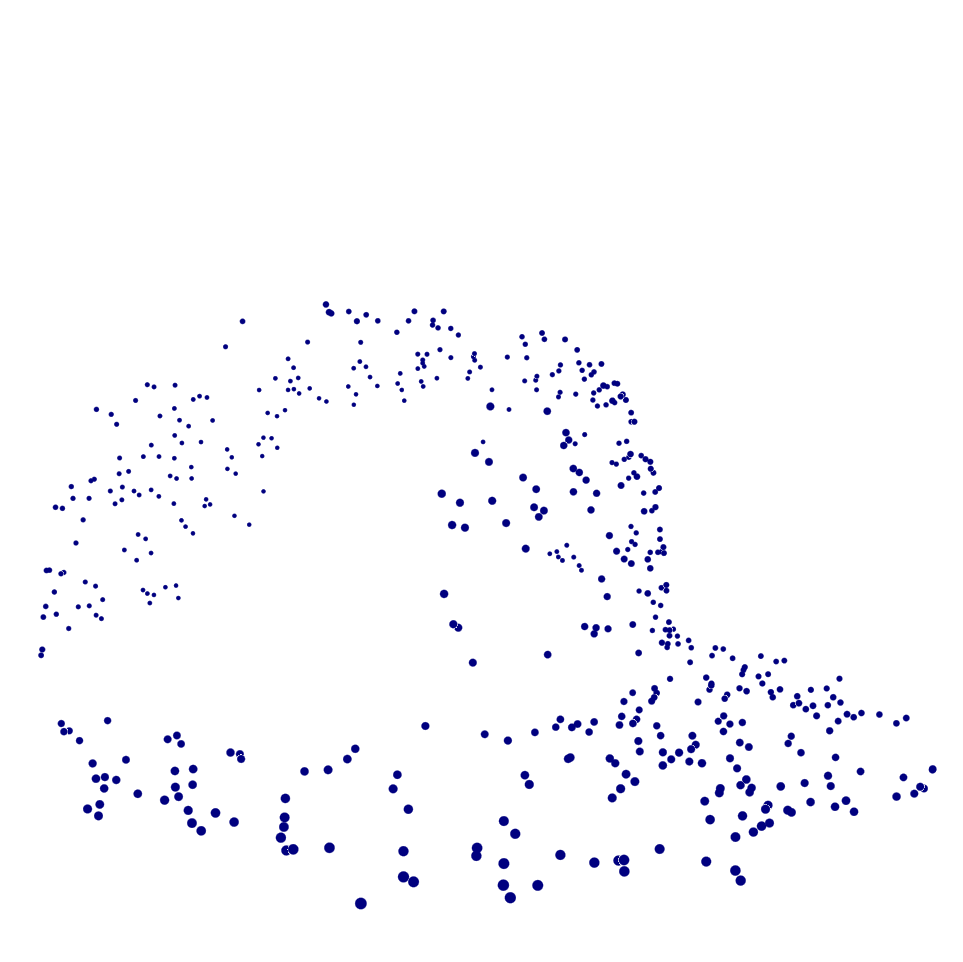}
			\end{minipage} &
			\begin{minipage}[b]{.11\textwidth}
				\centering
				\includegraphics[height=\linewidth]{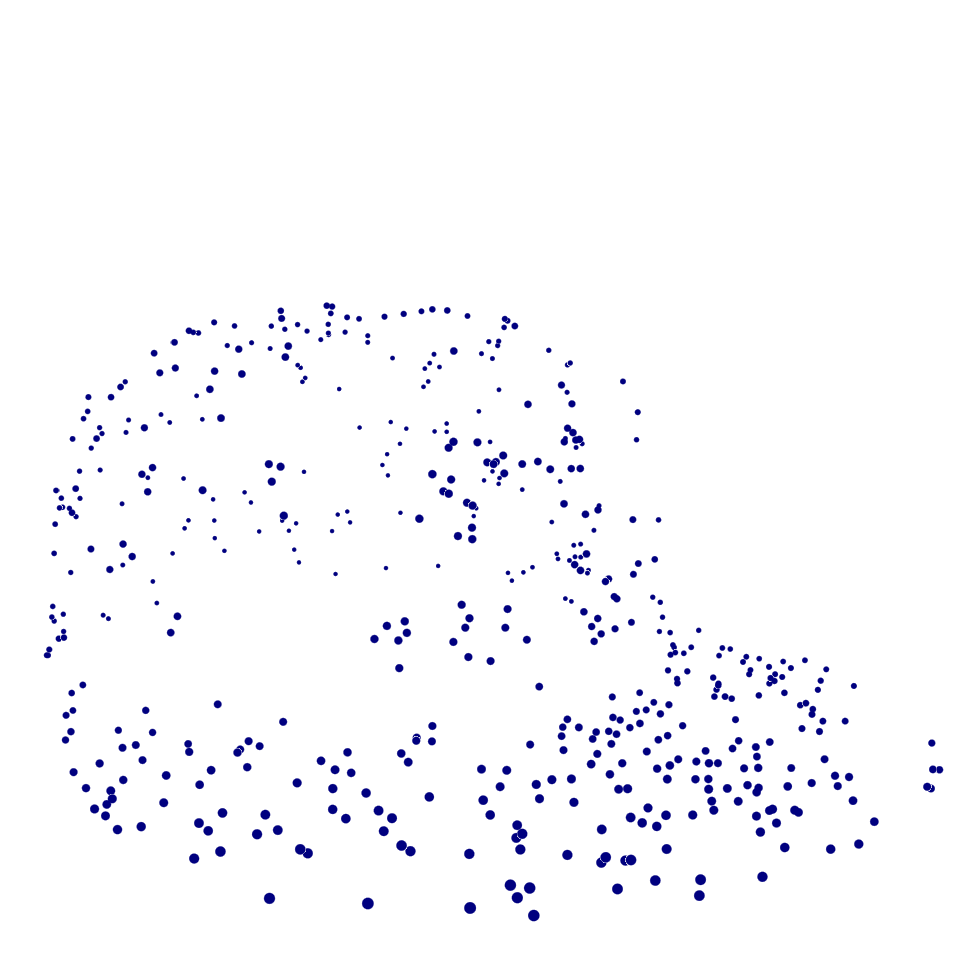}
			\end{minipage} &
			\begin{minipage}[b]{.11\textwidth}
				\centering
				\includegraphics[height=\linewidth]{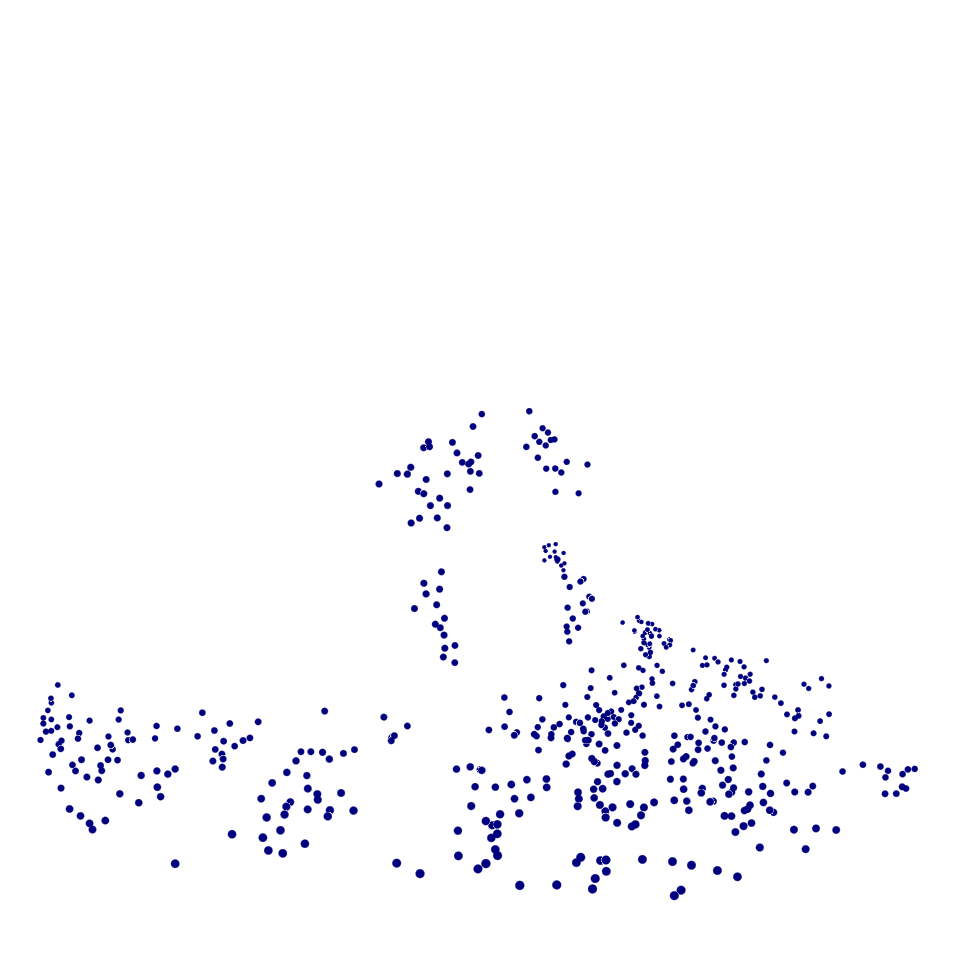}
			\end{minipage} &
			\begin{minipage}[b]{.11\textwidth}
				\centering
				\includegraphics[height=\linewidth]{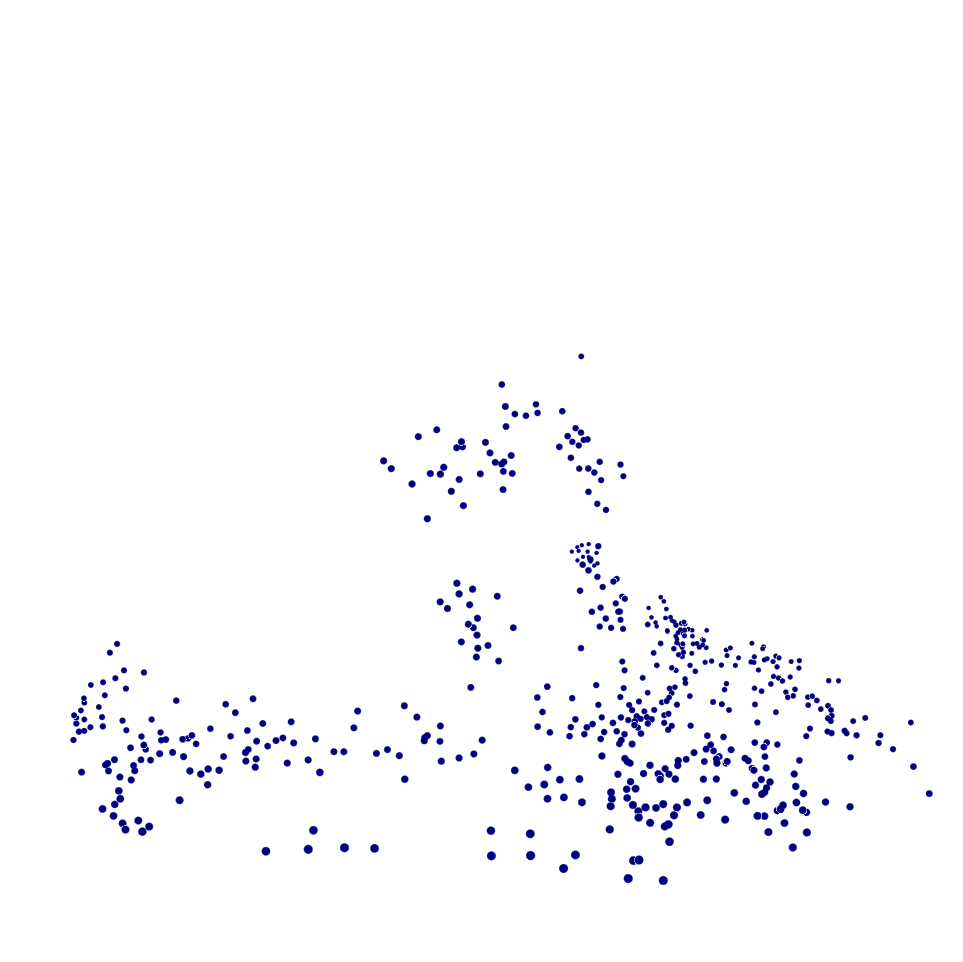}
			\end{minipage}\\
			\hline
			
			\begin{minipage}[b]{.11\textwidth}	
				\vspace{0.2mm}
				\centering
				\includegraphics[height=\linewidth]{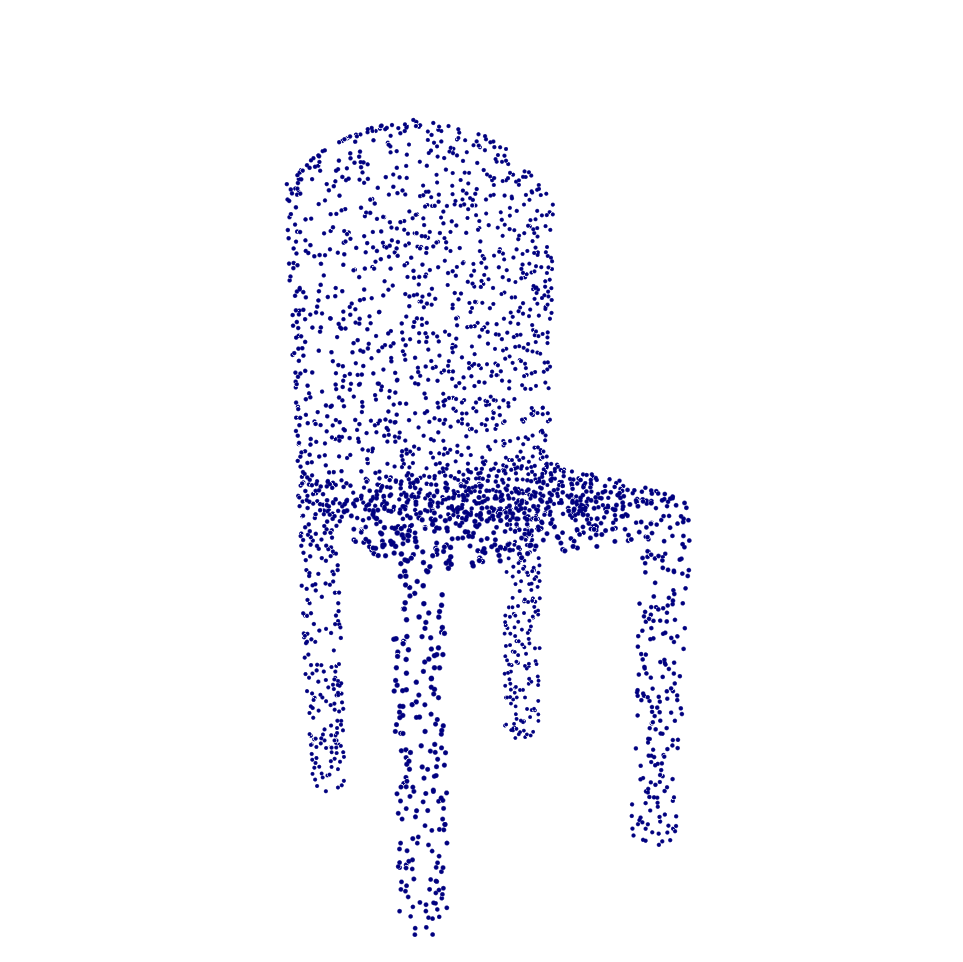}
			\end{minipage}&
			\begin{minipage}[b]{.11\textwidth}
				\centering
				\includegraphics[height=\linewidth]{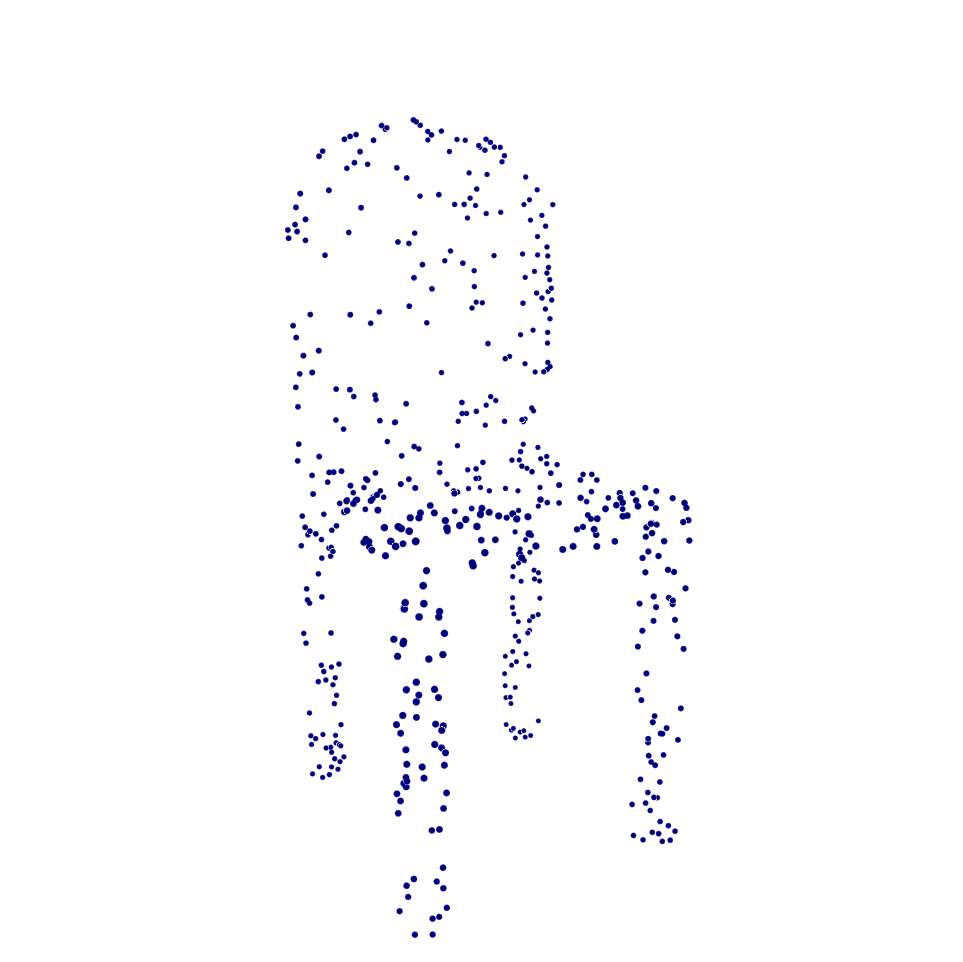}
			\end{minipage} &
			\begin{minipage}[b]{.11\textwidth}
				\centering
				\includegraphics[height=\linewidth]{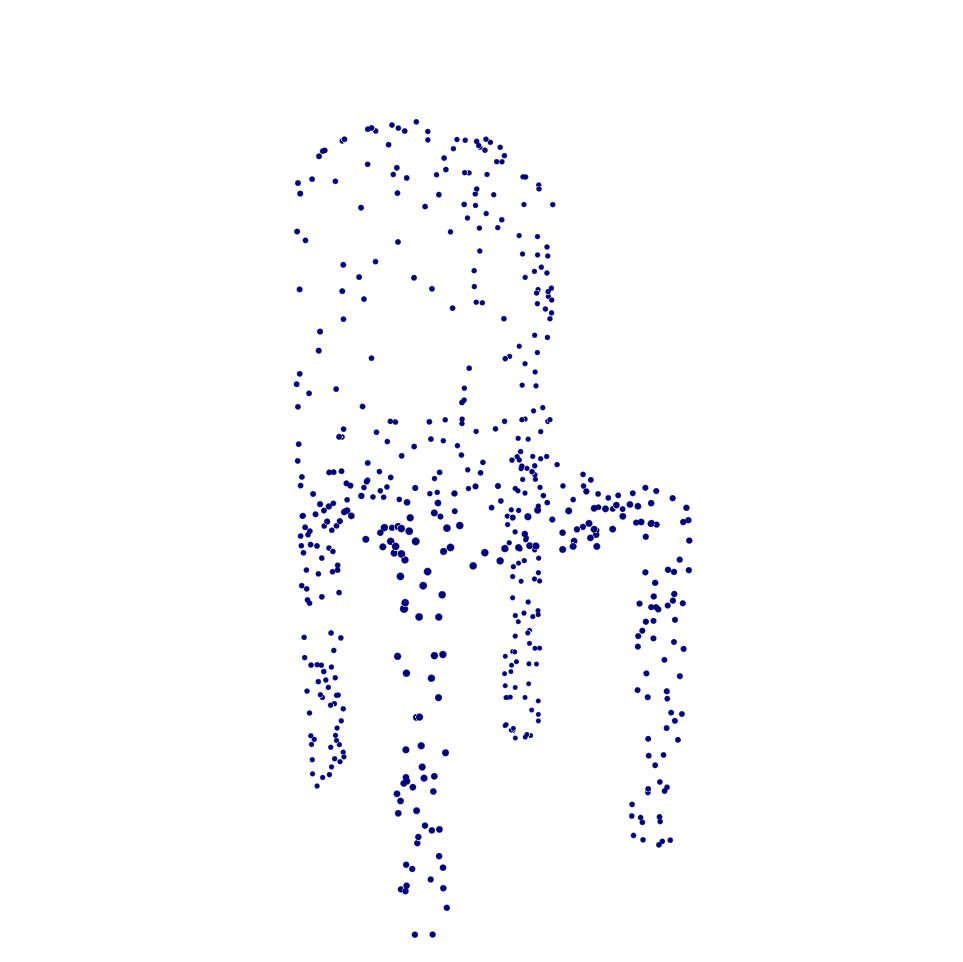}
			\end{minipage} &
			\begin{minipage}[b]{.11\textwidth}
				\centering
				\includegraphics[height=\linewidth]{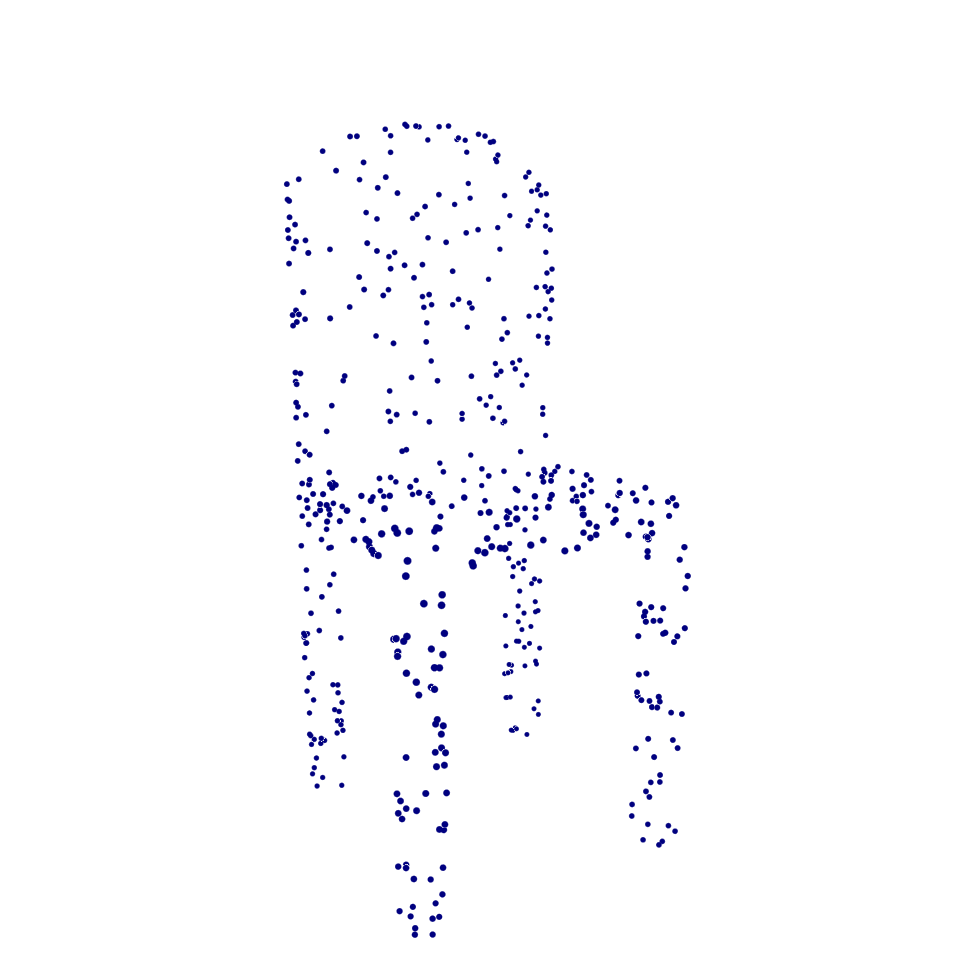}
			\end{minipage} &
			\begin{minipage}[b]{.11\textwidth}
				\centering
				\includegraphics[height=\linewidth]{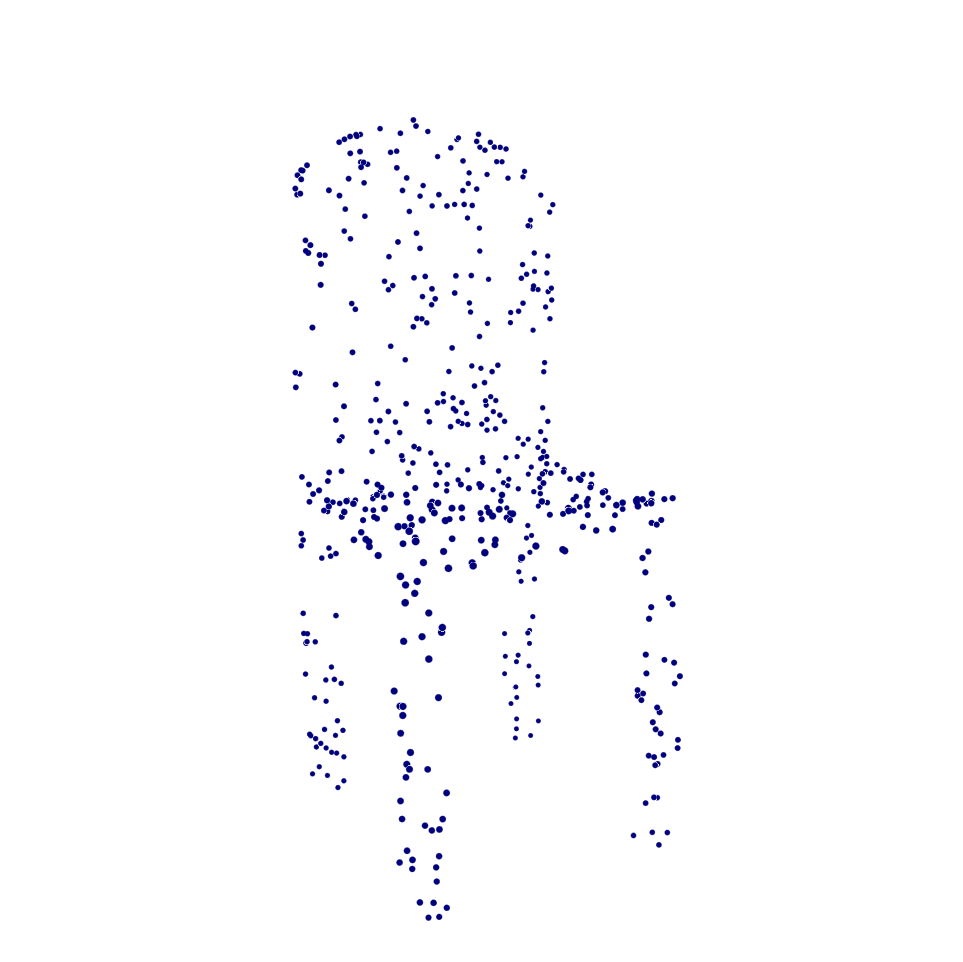}
			\end{minipage} &
			\begin{minipage}[b]{.11\textwidth}
				\centering
				\includegraphics[height=\linewidth]{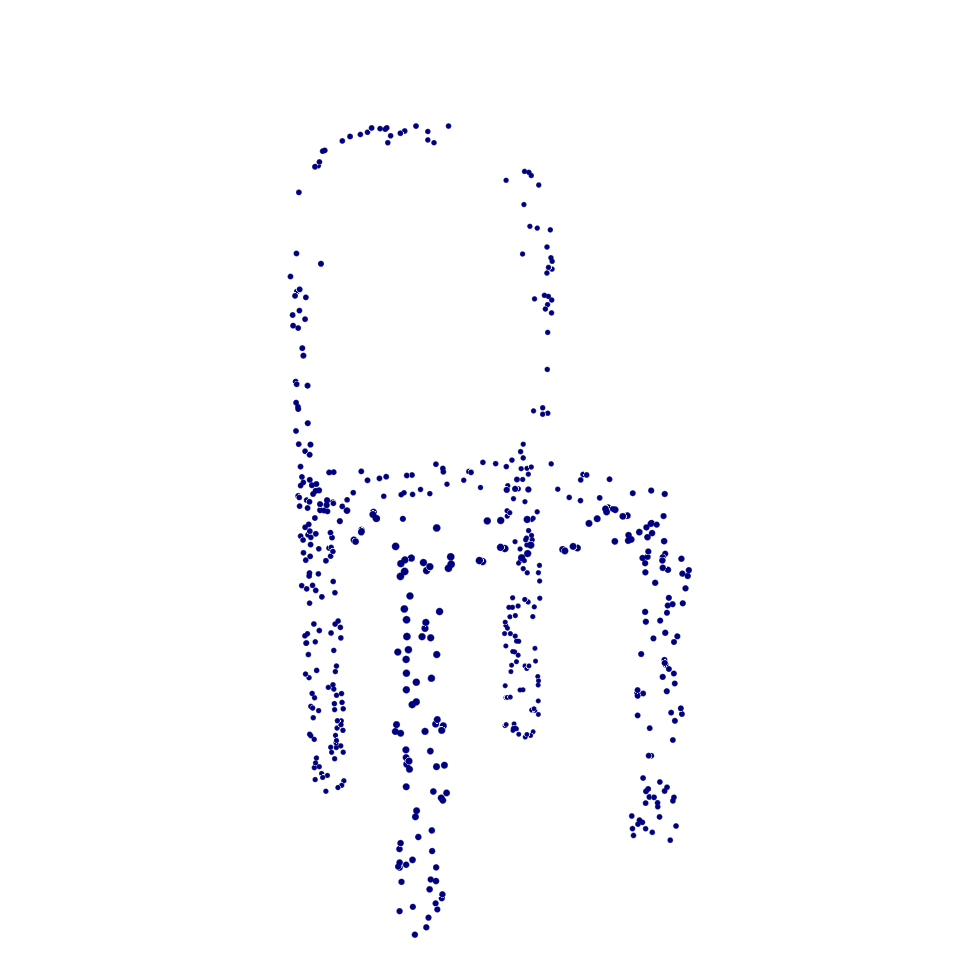}
			\end{minipage} &
			\begin{minipage}[b]{.11\textwidth}
				\centering
				\includegraphics[height=\linewidth]{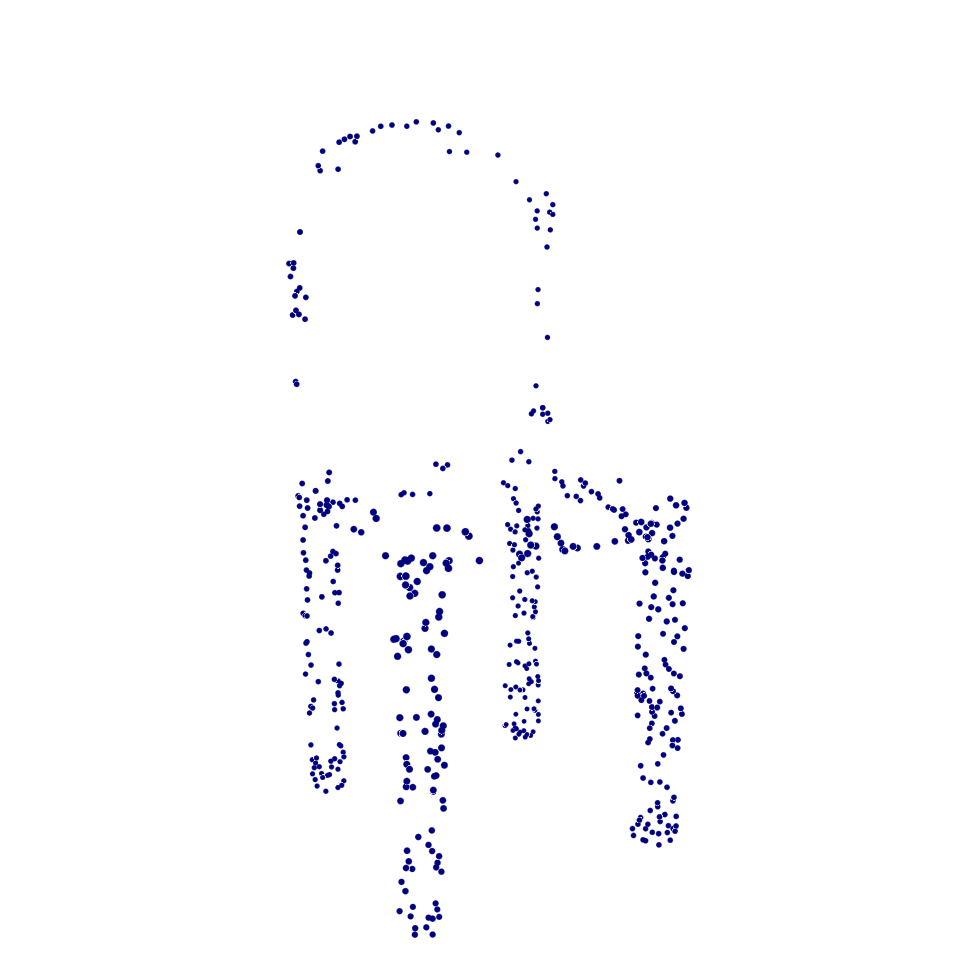}
			\end{minipage}\\
			\hline
			
			\begin{minipage}[b]{.11\textwidth}
				\vspace{0.1mm}	
				\centering
				\includegraphics[height=\linewidth]{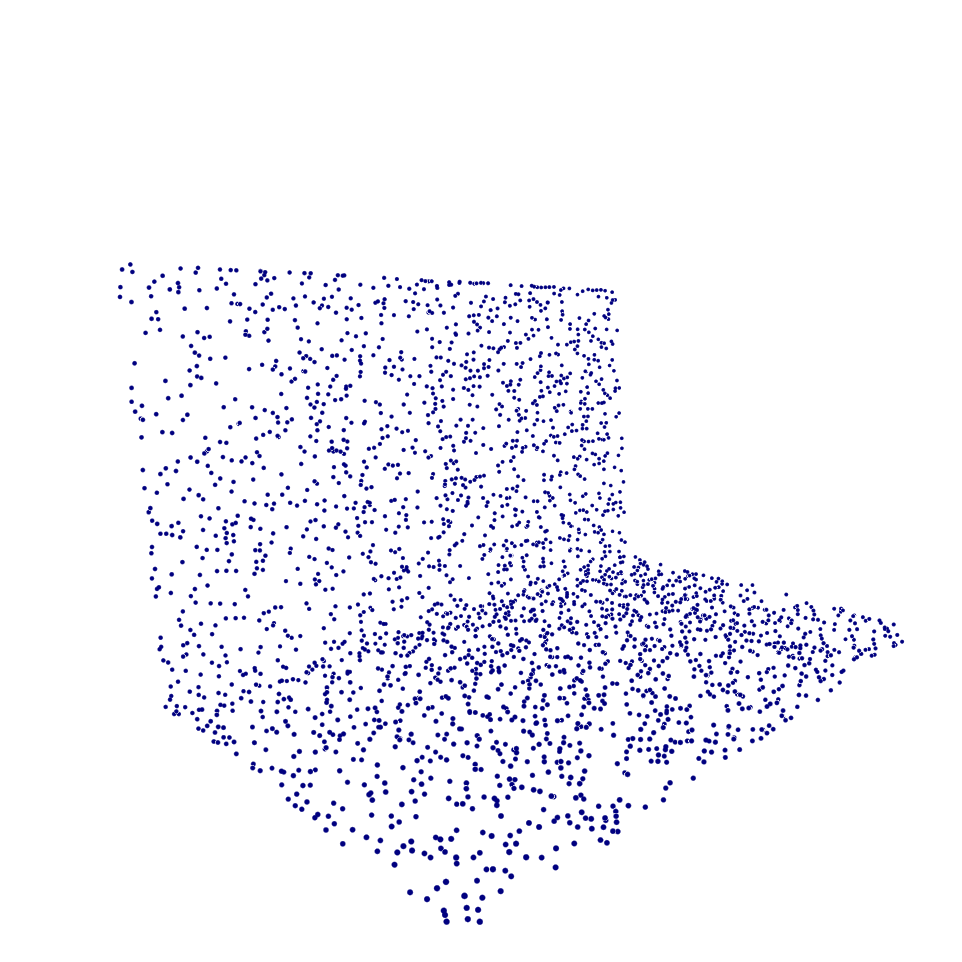}
			\end{minipage}&
			\begin{minipage}[b]{.11\textwidth}
				\centering
				\includegraphics[height=\linewidth]{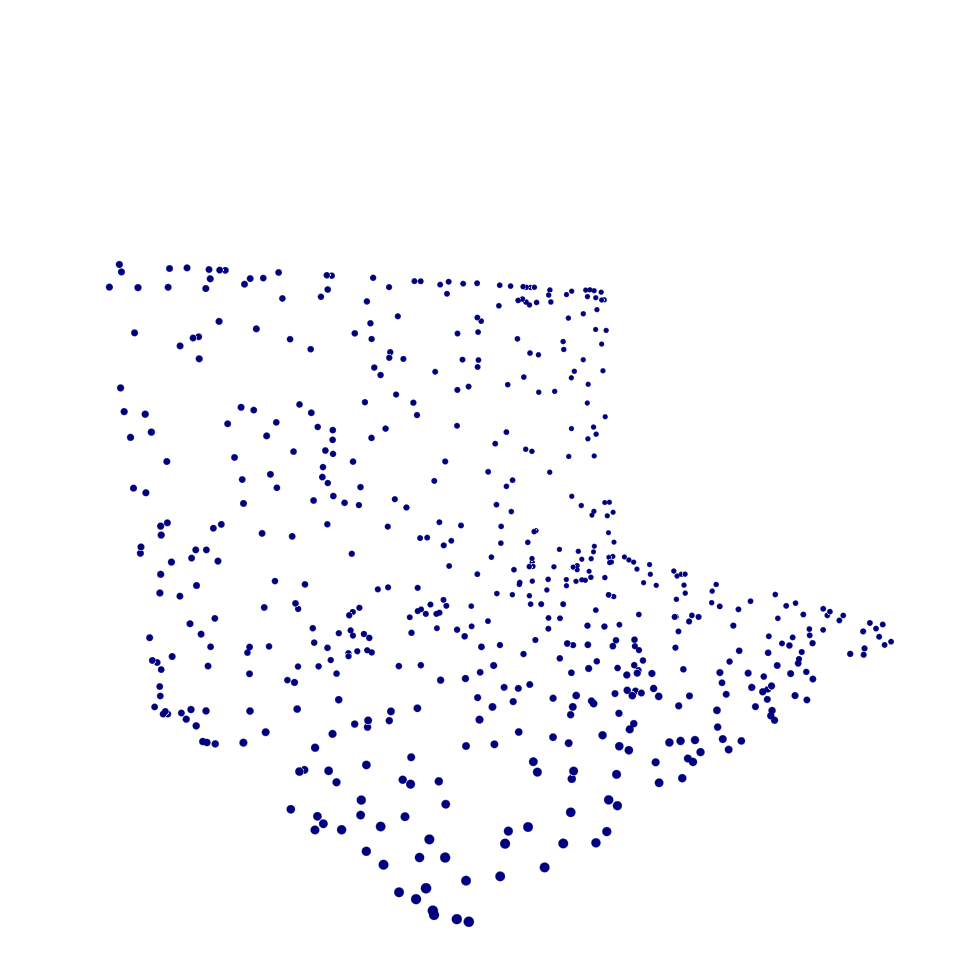}
			\end{minipage} &
			\begin{minipage}[b]{.11\textwidth}
				\centering
				\includegraphics[height=\linewidth]{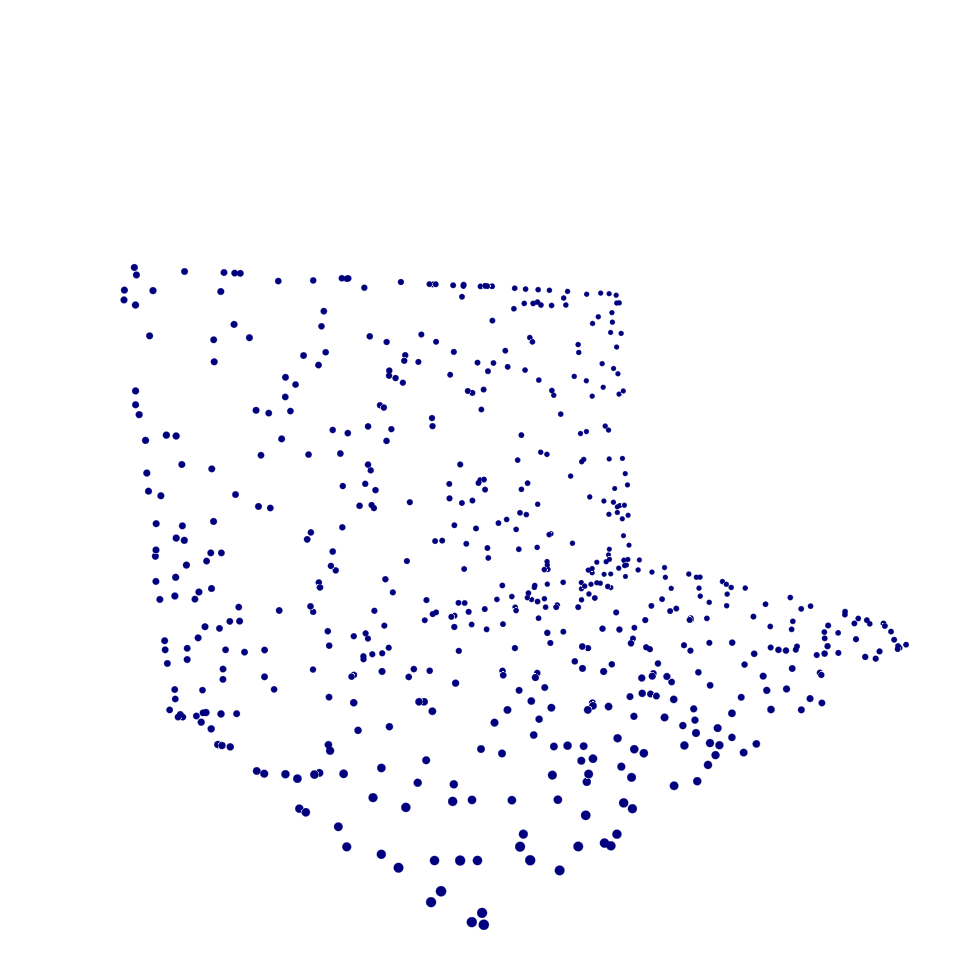}
			\end{minipage} &
			\begin{minipage}[b]{.11\textwidth}
				\centering
				\includegraphics[height=\linewidth]{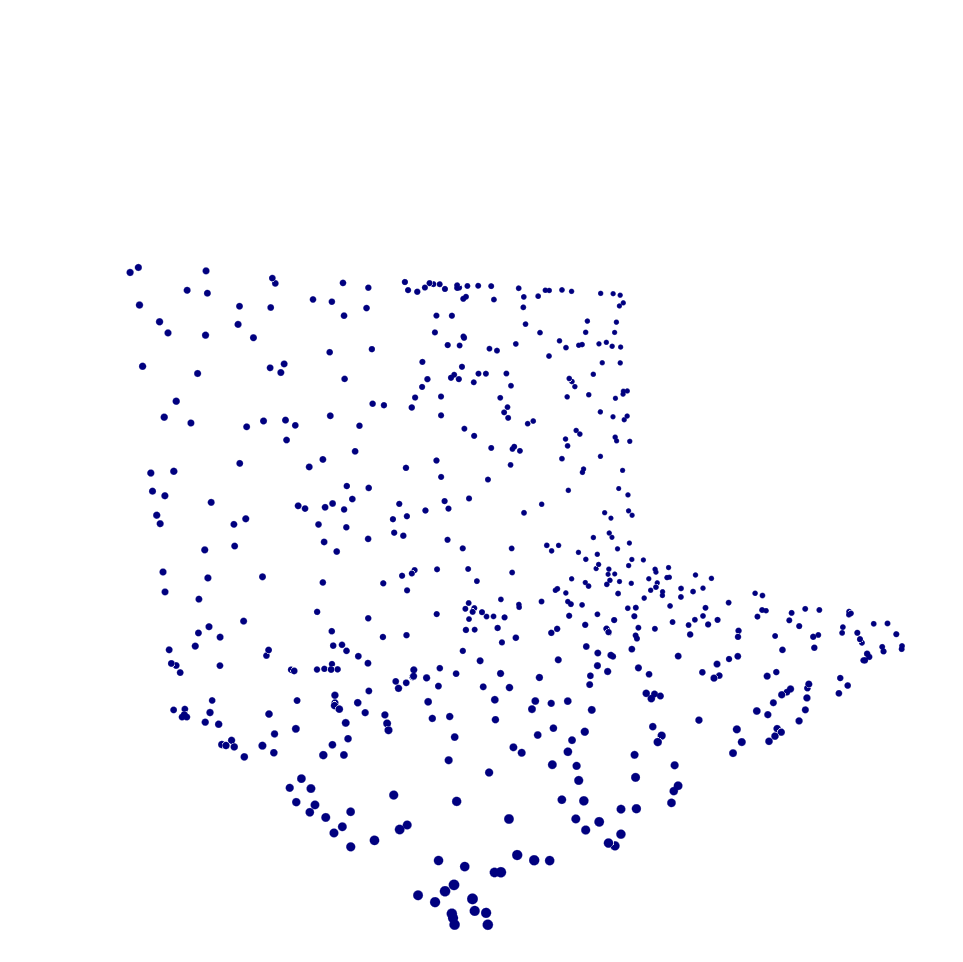}
			\end{minipage} &
			\begin{minipage}[b]{.11\textwidth}
				\centering
				\includegraphics[height=\linewidth]{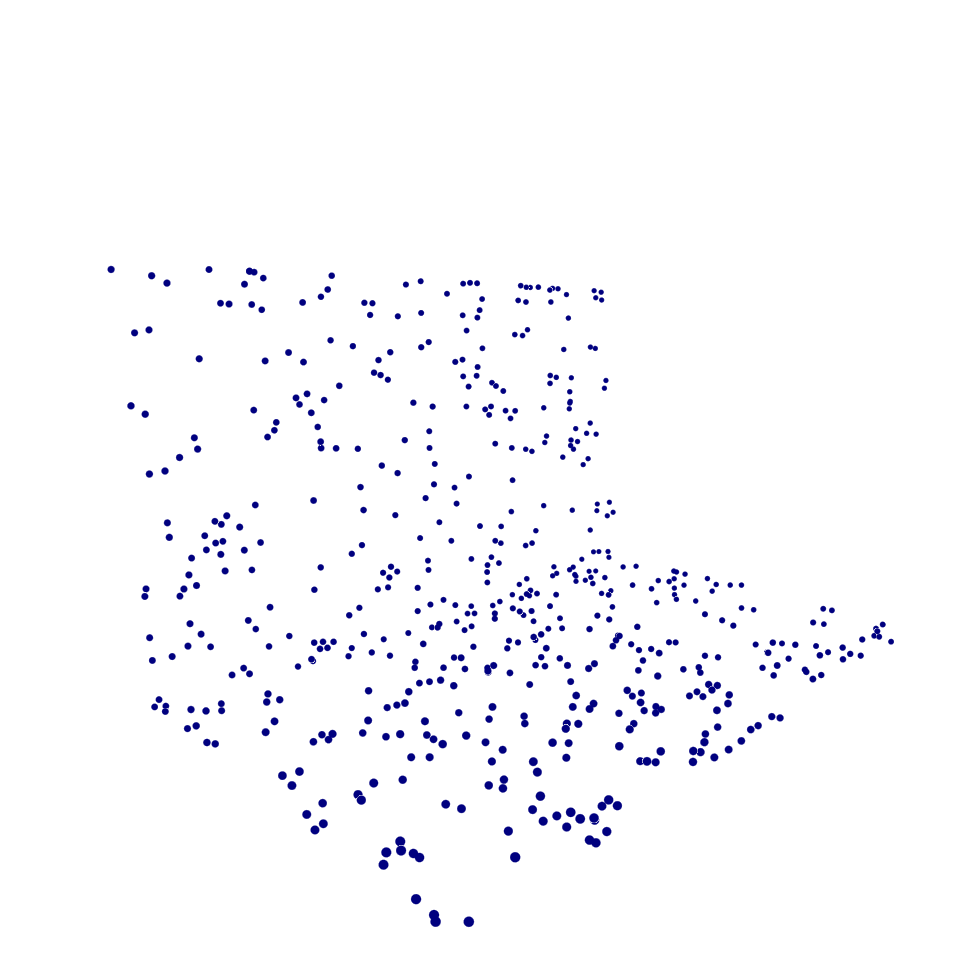}
			\end{minipage} &
			\begin{minipage}[b]{.11\textwidth}
				\centering
				\includegraphics[height=\linewidth]{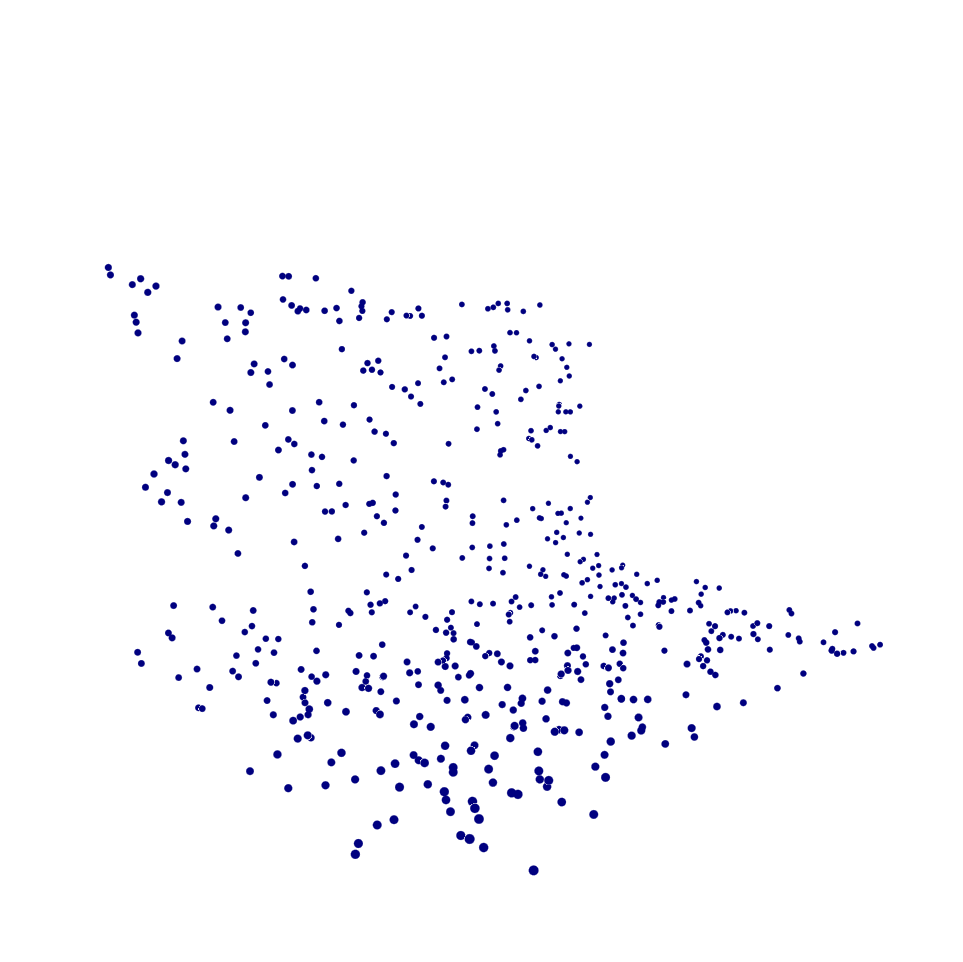}
			\end{minipage} &
			\begin{minipage}[b]{.11\textwidth}
				\centering
				\includegraphics[height=\linewidth]{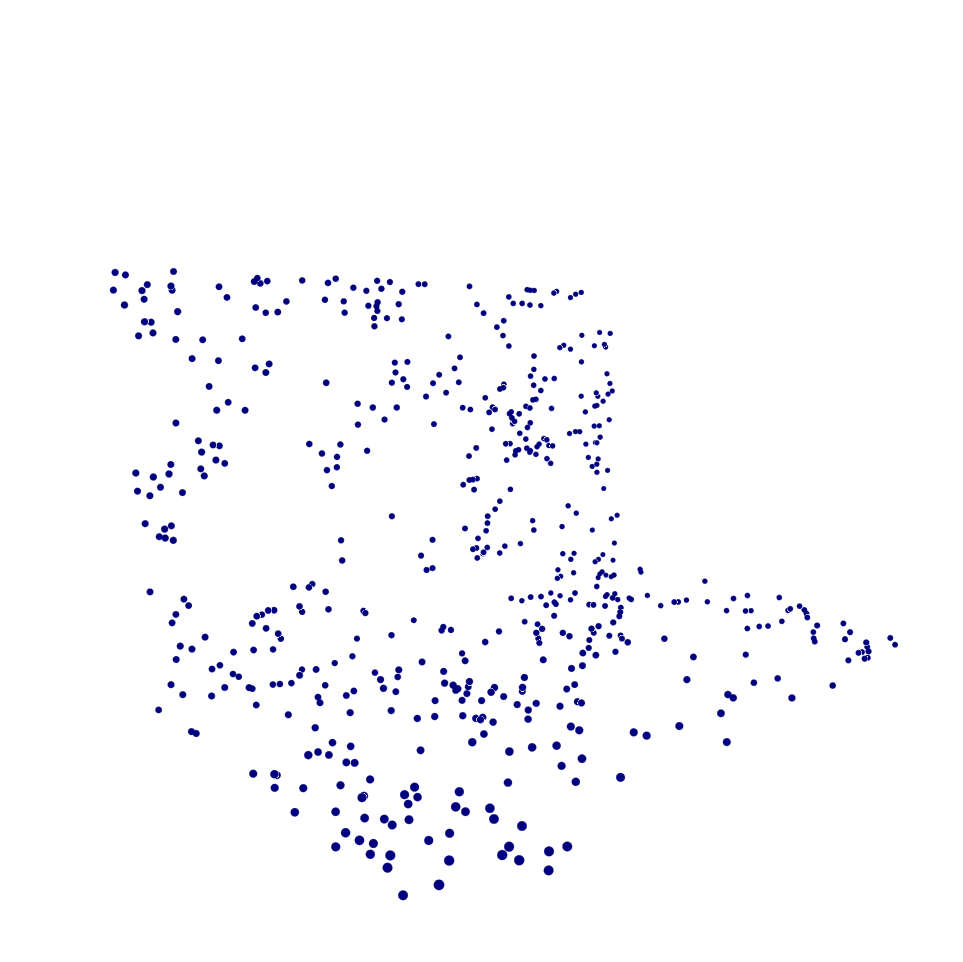}
			\end{minipage}\\
			\hline
			
			\begin{minipage}[b]{.11\textwidth}
				\vspace{0.1mm}
				\includegraphics[height=\linewidth]{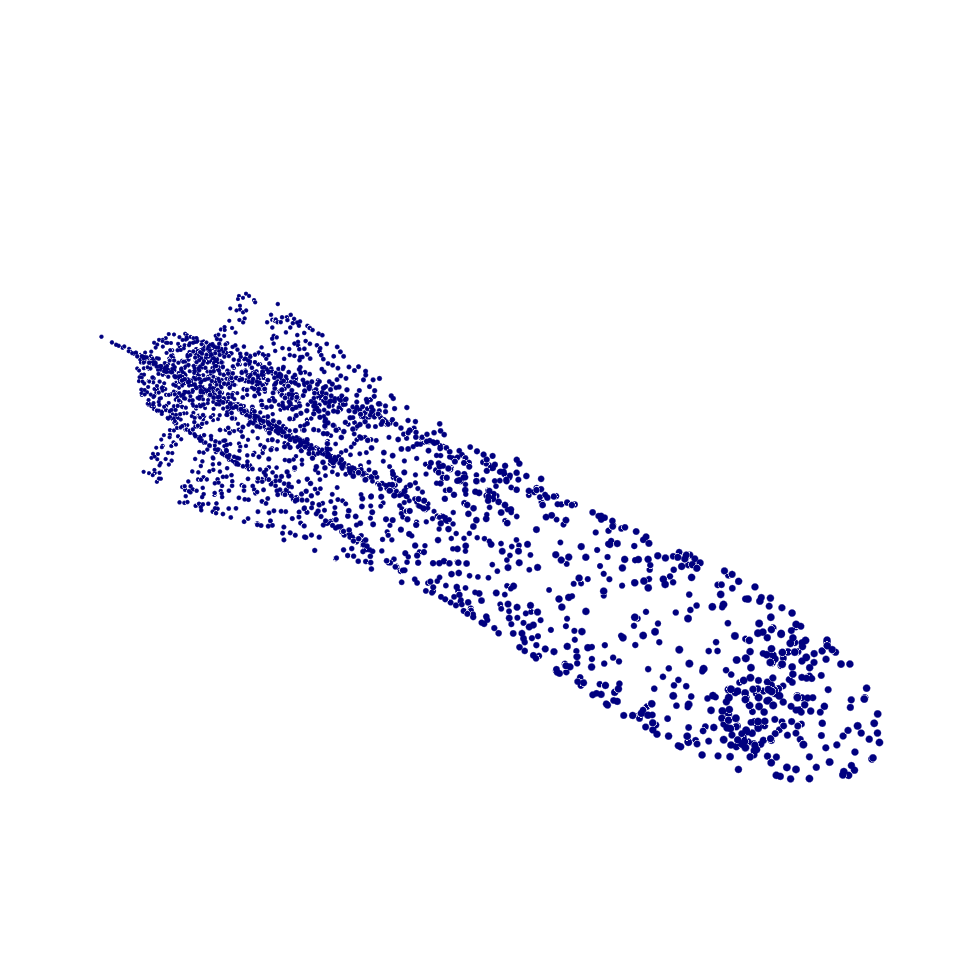}
			\end{minipage}&
			\begin{minipage}[b]{.11\textwidth}
				\centering
				\includegraphics[height=\linewidth]{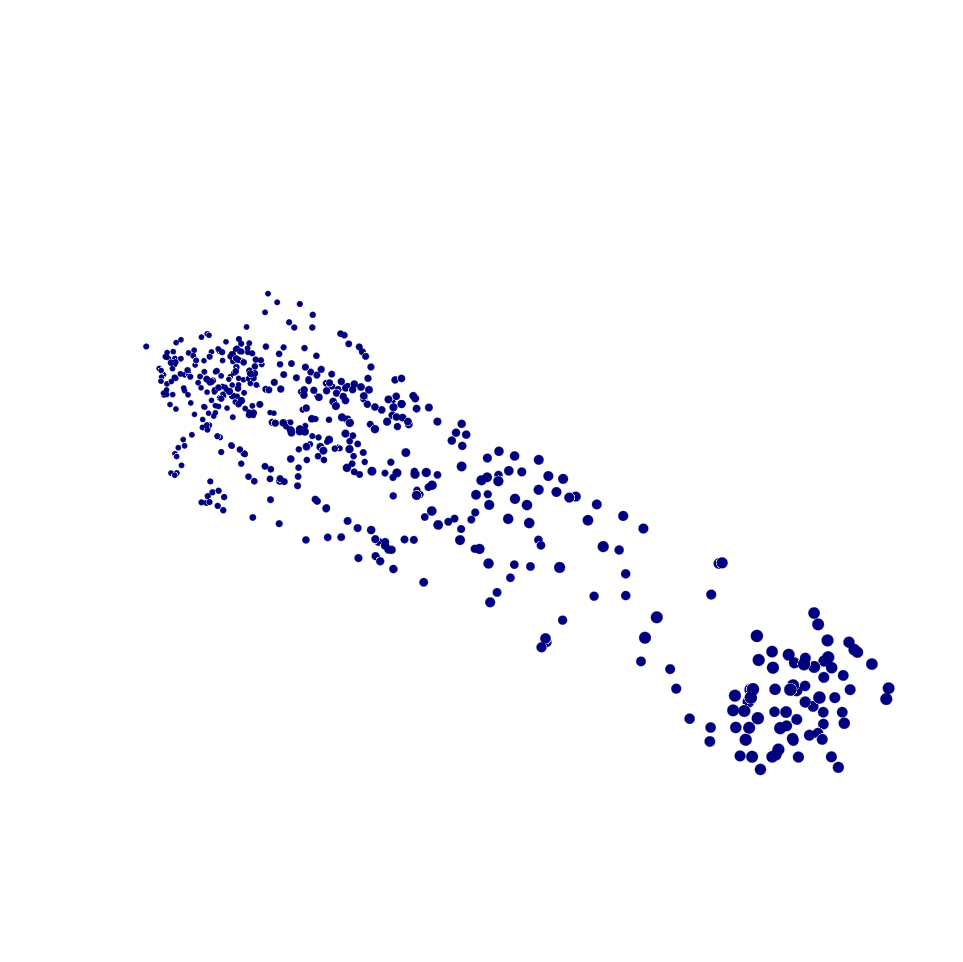}
			\end{minipage} &
			\begin{minipage}[b]{.11\textwidth}
				\centering
				\includegraphics[height=\linewidth]{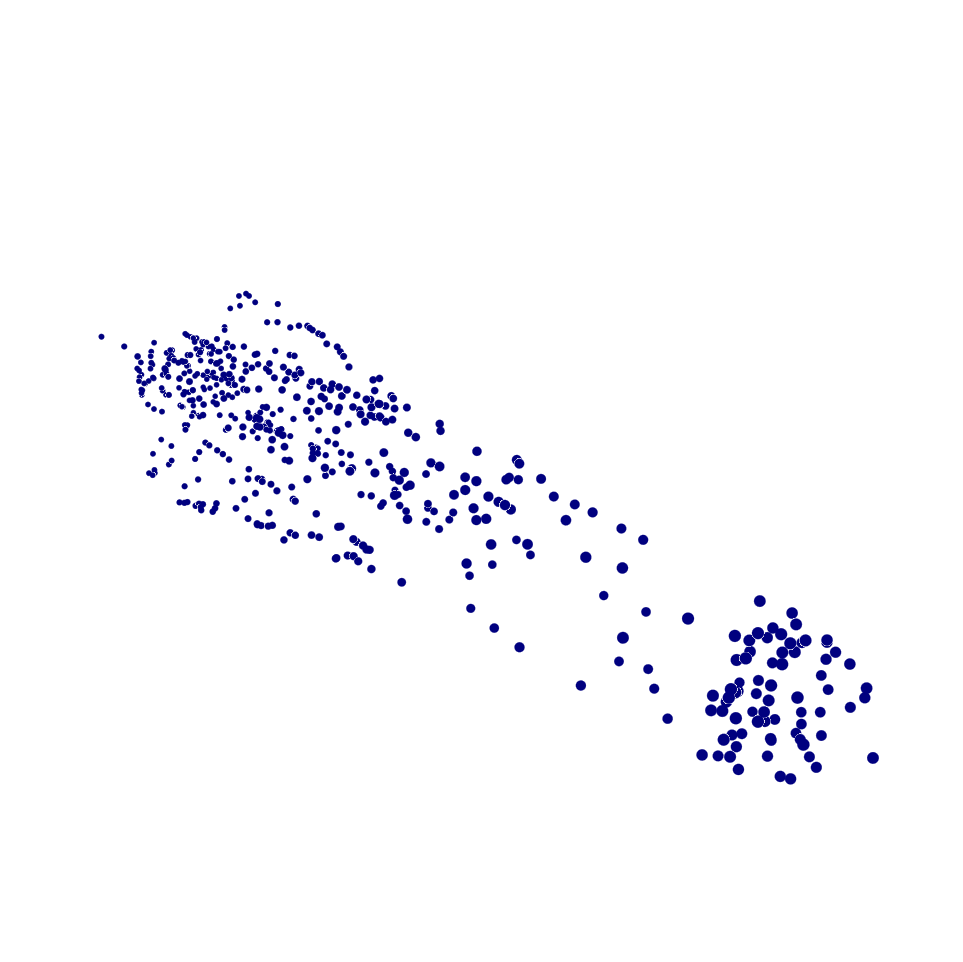}
			\end{minipage} &
			\begin{minipage}[b]{.11\textwidth}
				\centering
				\includegraphics[height=\linewidth]{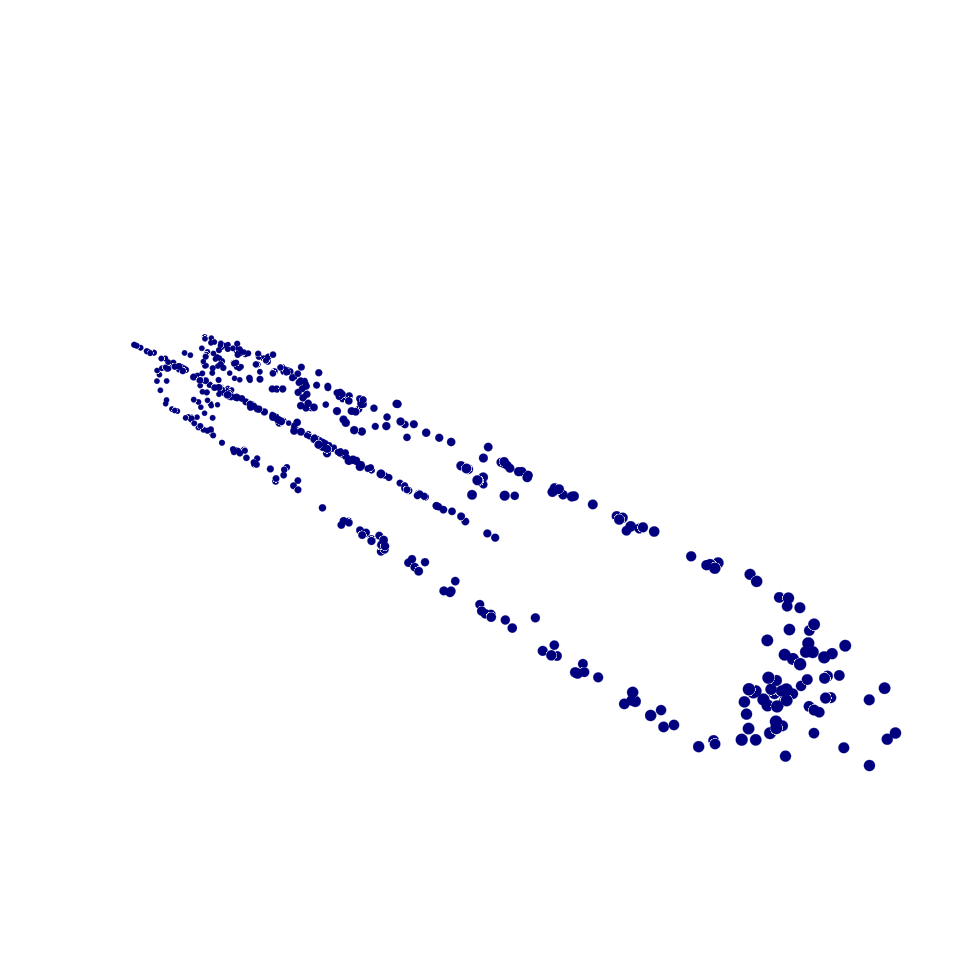}
			\end{minipage} &
			\begin{minipage}[b]{.11\textwidth}
				\centering
				\includegraphics[height=\linewidth]{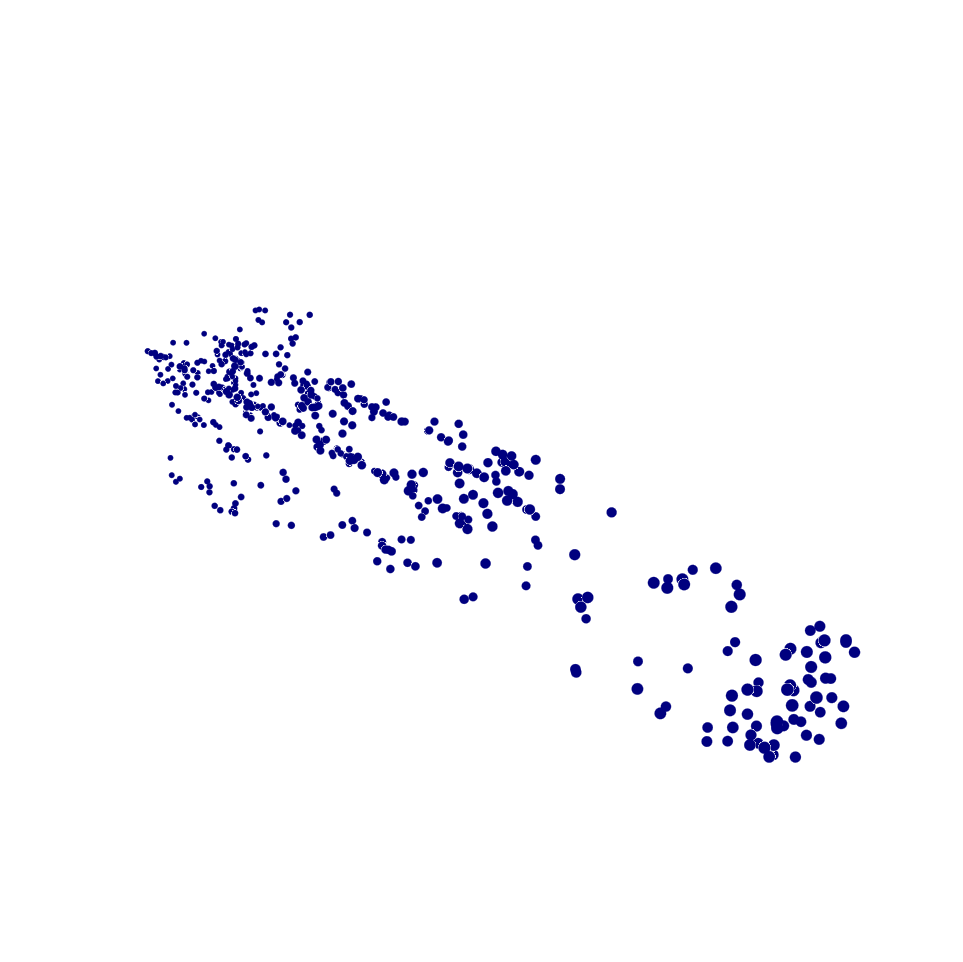}
			\end{minipage} &
			\begin{minipage}[b]{.11\textwidth}
				\centering
				\includegraphics[height=\linewidth]{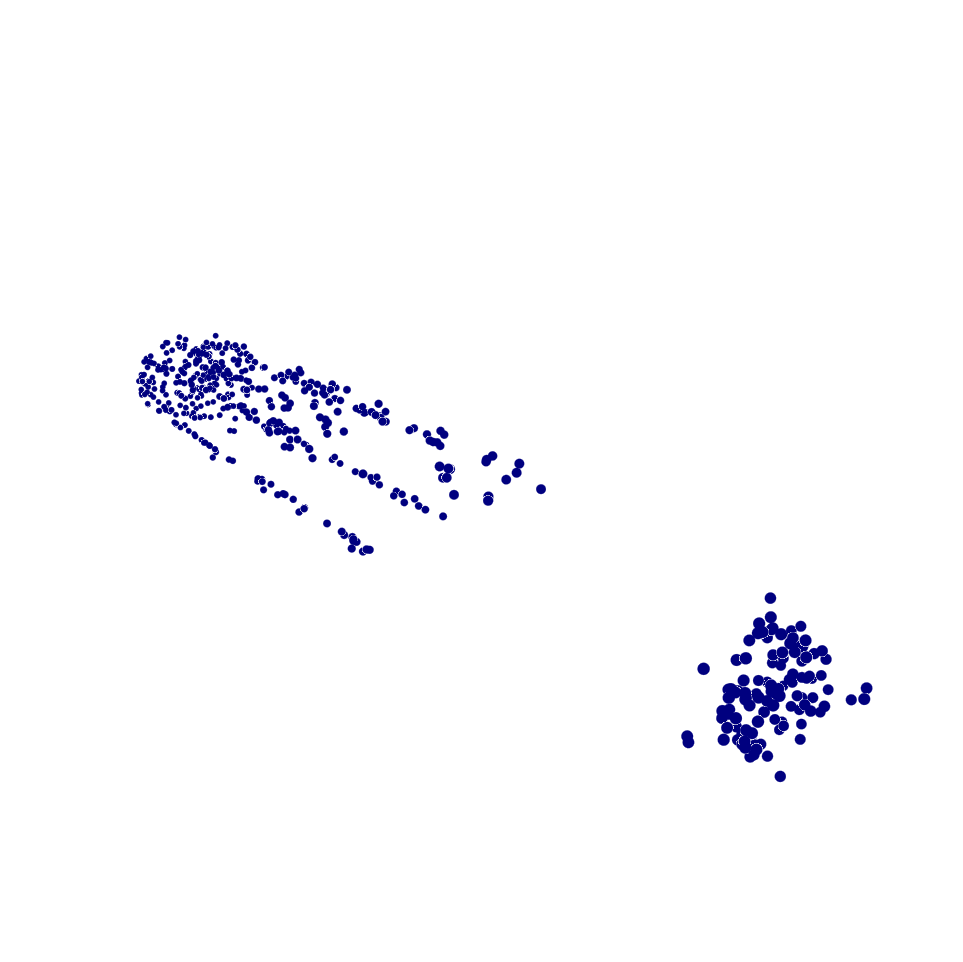}
			\end{minipage} &
			\begin{minipage}[b]{.11\textwidth}
				\centering
				\includegraphics[height=\linewidth]{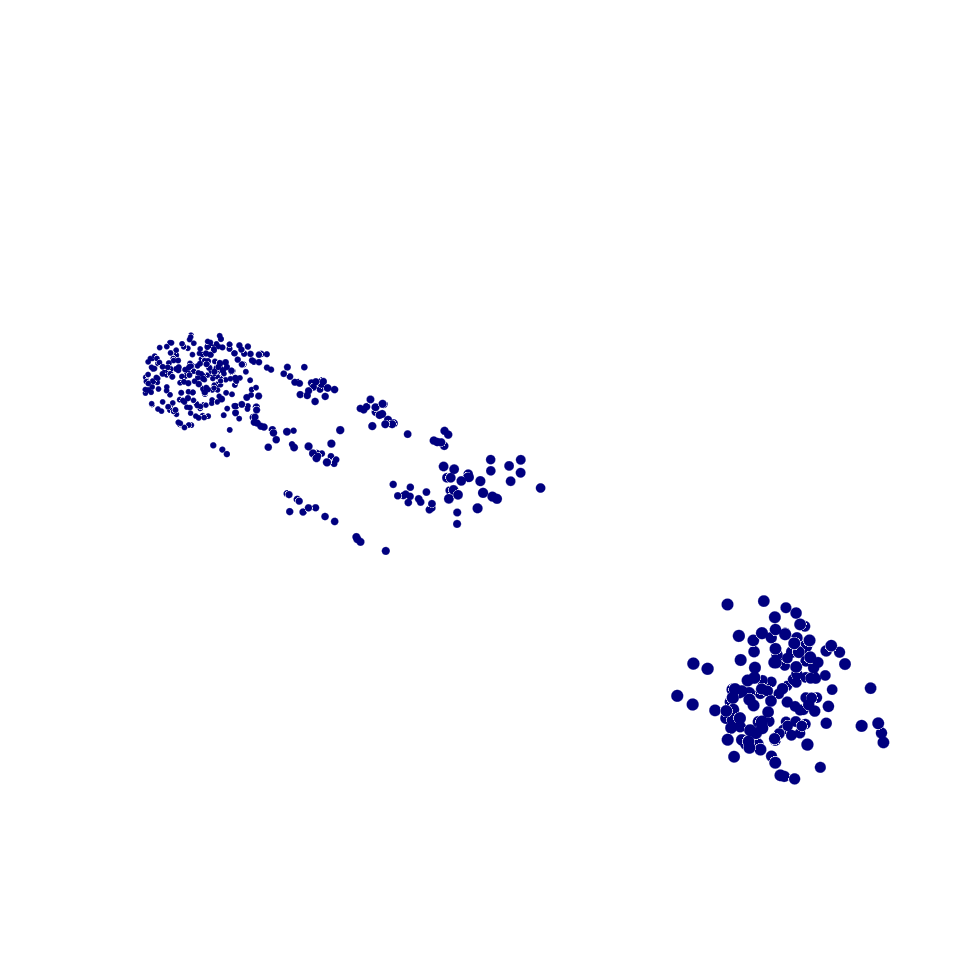}
			\end{minipage}\\
			\hline
			
			\begin{minipage}[b]{.11\textwidth}
				\vspace{0.1mm}
				\includegraphics[height=\linewidth]{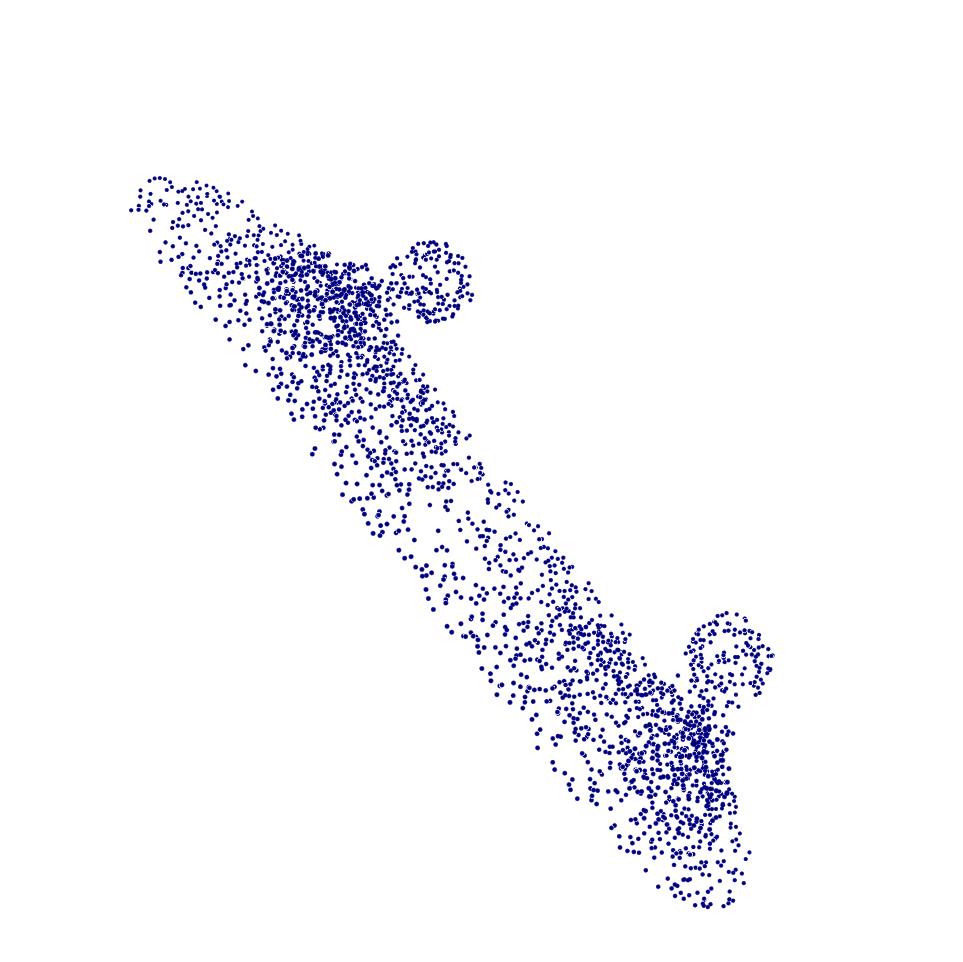}
			\end{minipage}&
			\begin{minipage}[b]{.11\textwidth}
				\centering
				\includegraphics[height=\linewidth]{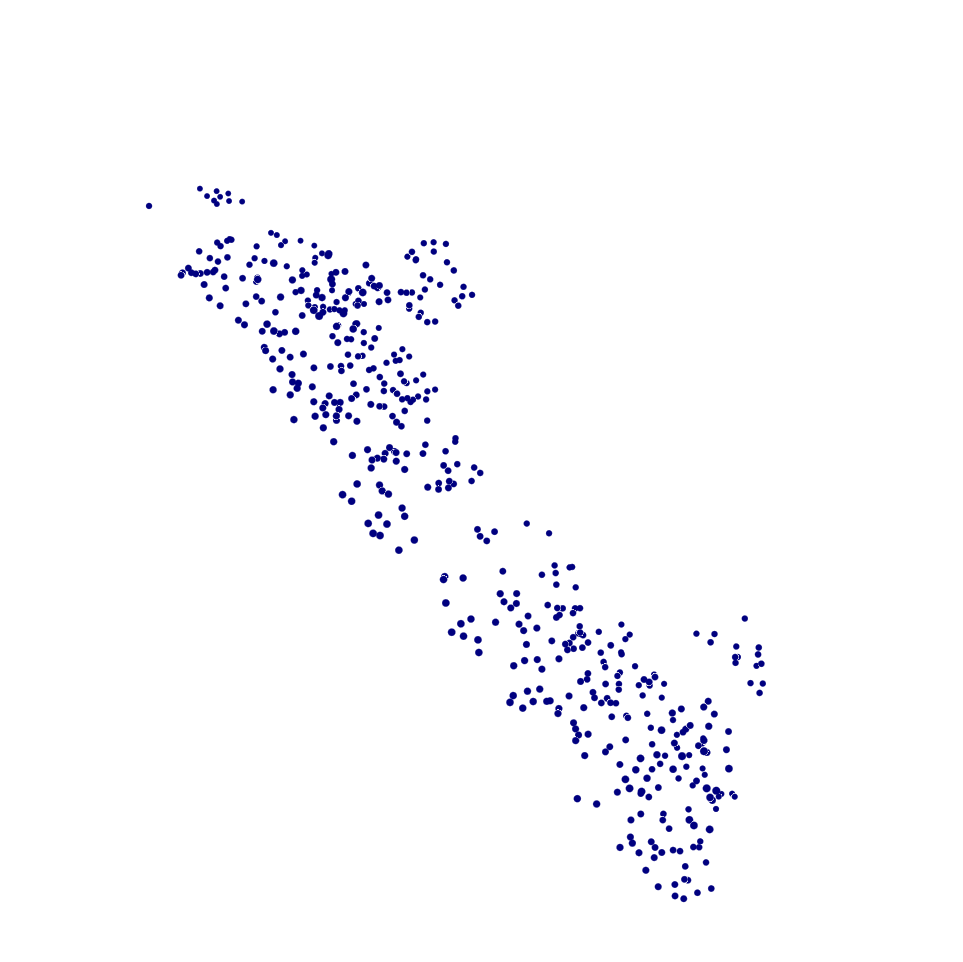}
			\end{minipage} &
			\begin{minipage}[b]{.11\textwidth}
				\centering
				\includegraphics[height=\linewidth]{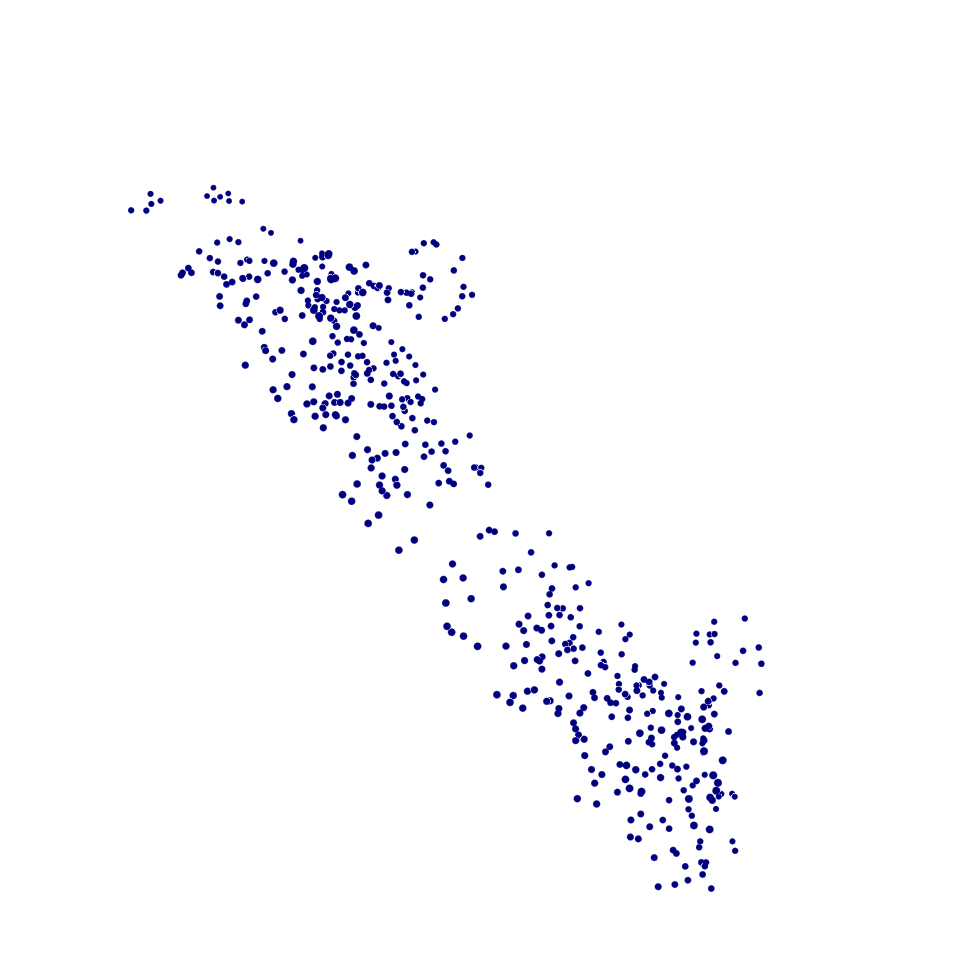}
			\end{minipage} &
			\begin{minipage}[b]{.11\textwidth}
				\centering
				\includegraphics[height=\linewidth]{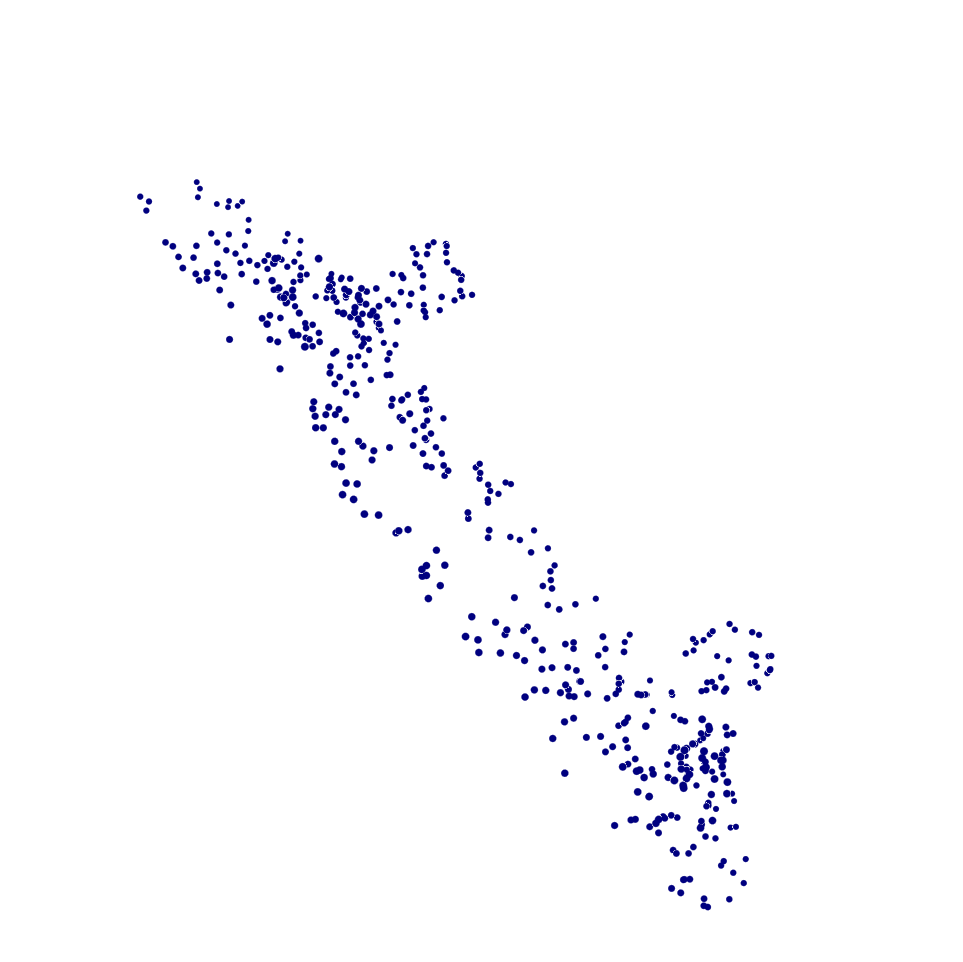}
			\end{minipage} &
			\begin{minipage}[b]{.11\textwidth}
				\centering
				\includegraphics[height=\linewidth]{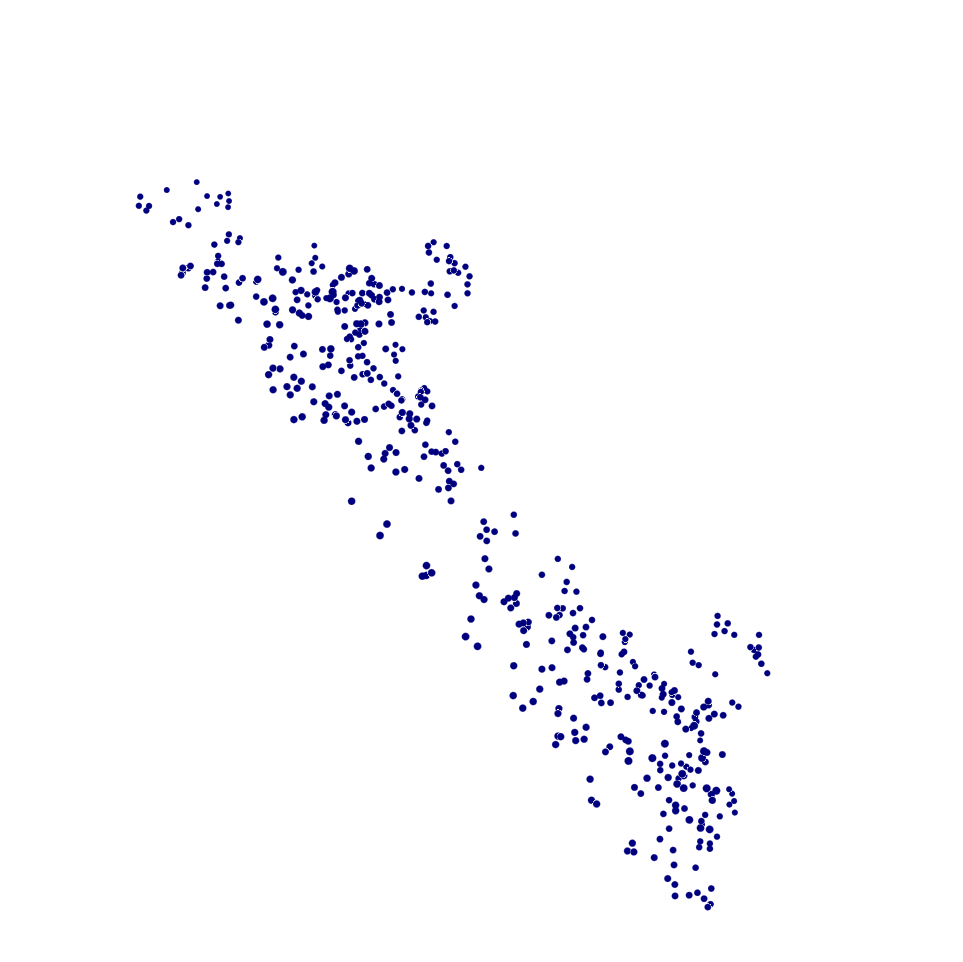}
			\end{minipage} &
			\begin{minipage}[b]{.11\textwidth}
				\centering
				\includegraphics[height=\linewidth]{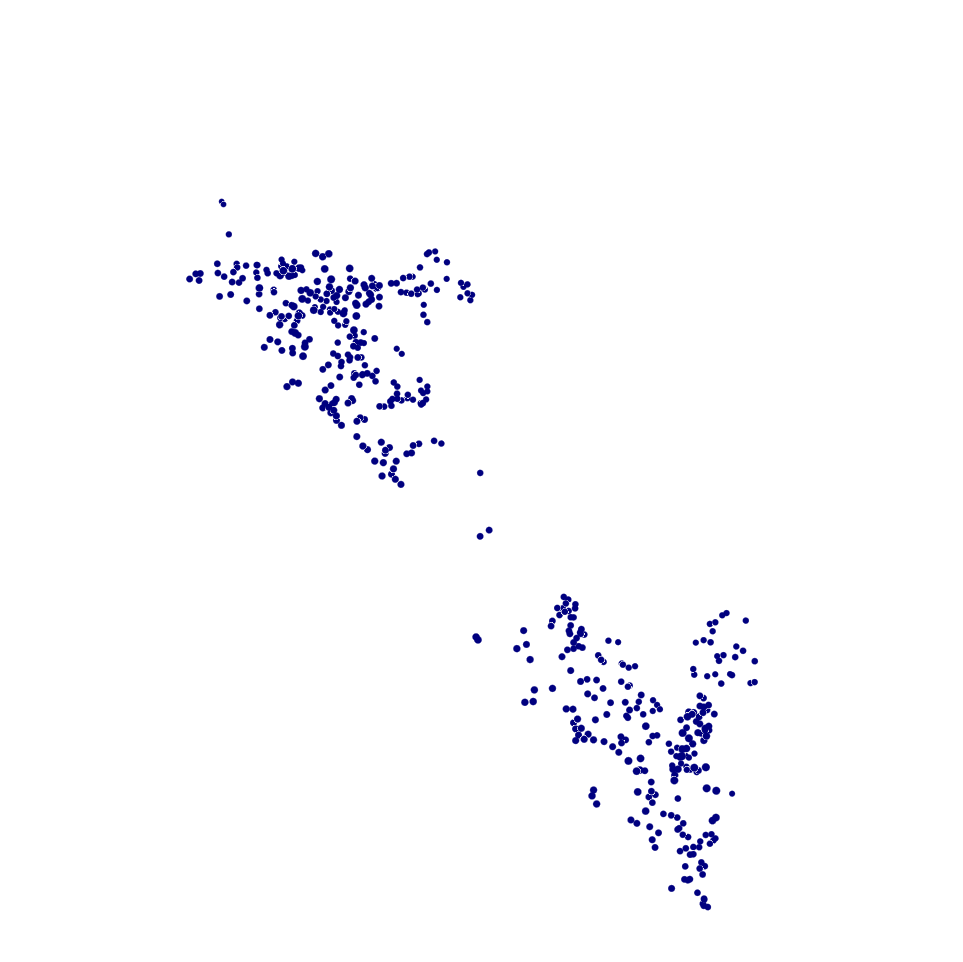}
			\end{minipage} &
			\begin{minipage}[b]{.11\textwidth}
				\centering
				\includegraphics[height=\linewidth]{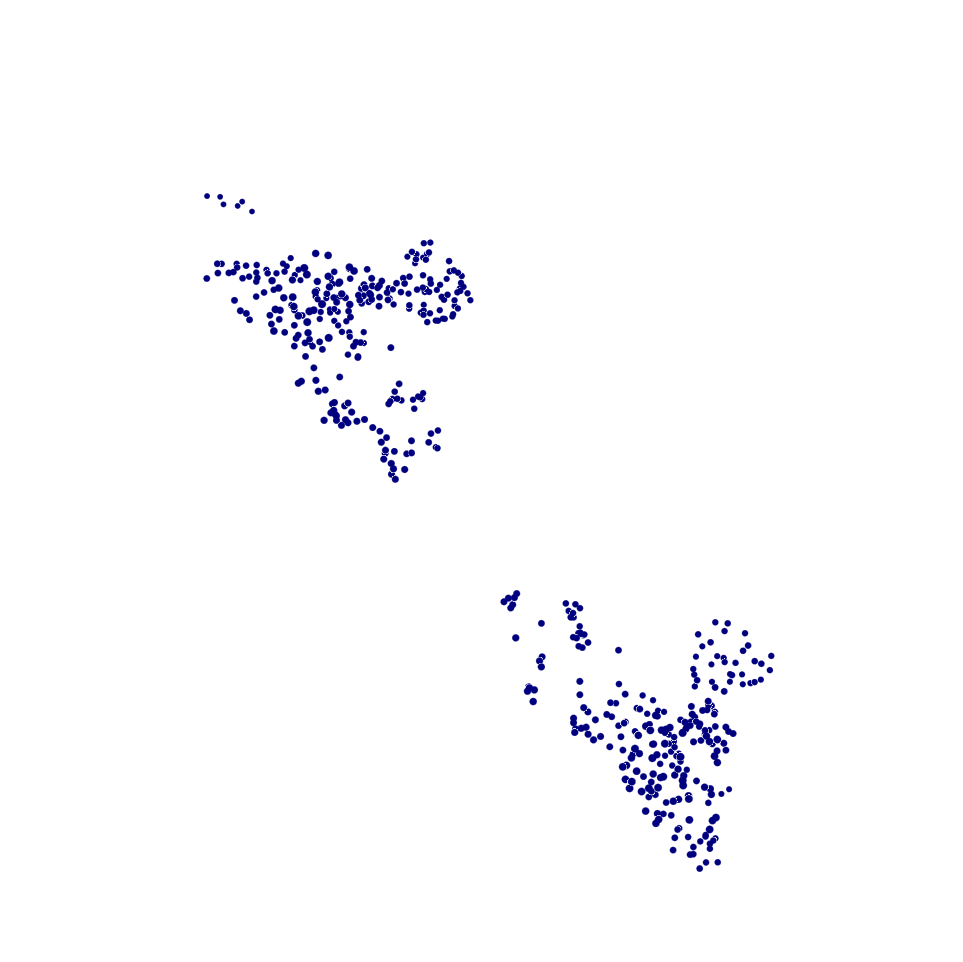}
			\end{minipage}\\
			\hline
			
		\end{tabular}
		\caption{Resampled point clouds}
	\end{subtable}
	\vspace{2mm}
	\\
	\begin{subtable}{0.9\textwidth}
		\begin{tabular}{|c|c|c|c|c|c|c|}
			\hline
			Original & \tabincell{c}{HKC} & \tabincell{c}{HKF} & \tabincell{c}{LHF} & \tabincell{c}{GFR} & \tabincell{c}{EA} & \tabincell{c}{PCA-AC}\\
			\hline		
			\begin{minipage}[b]{.11\textwidth}	
				\vspace{0.1mm}
				\centering
				\includegraphics[height=\linewidth]{Cap_oriN00.png}
			\end{minipage}&
			\begin{minipage}[b]{.11\textwidth}
				\centering
				\includegraphics[height=\linewidth]{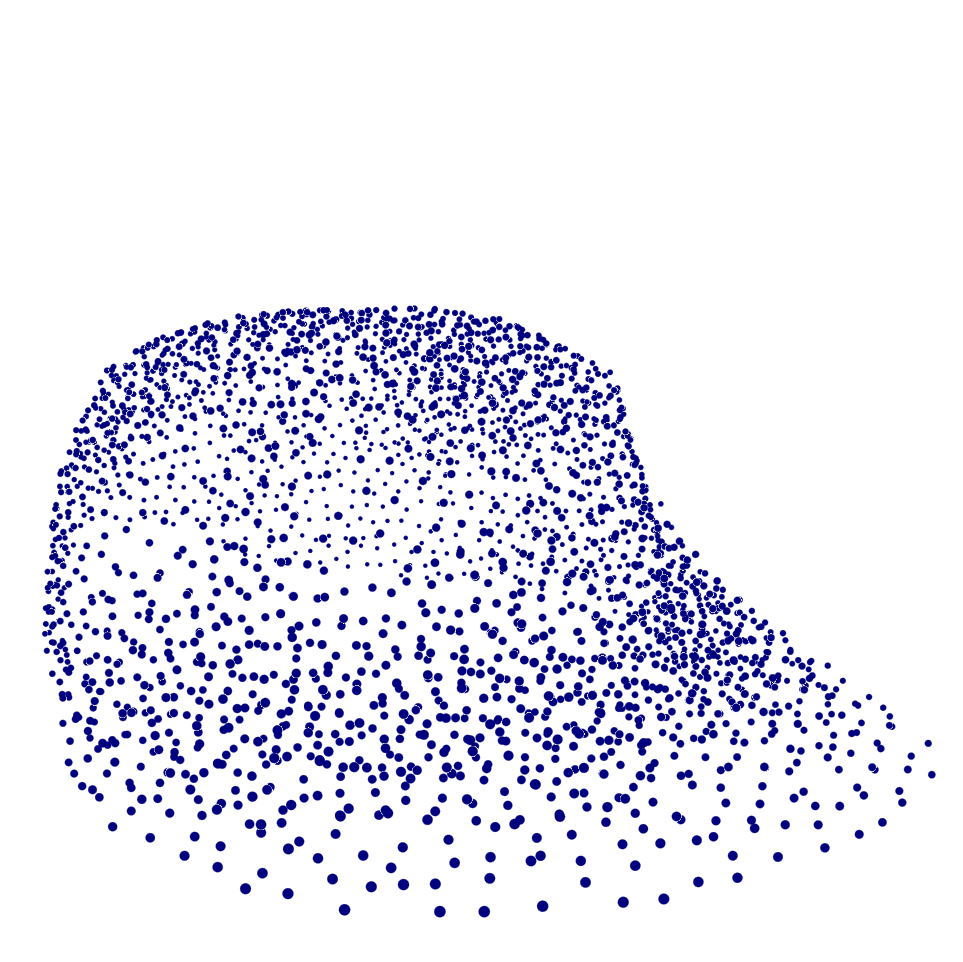}
			\end{minipage} &
			\begin{minipage}[b]{.11\textwidth}
				\centering
				\includegraphics[height=\linewidth]{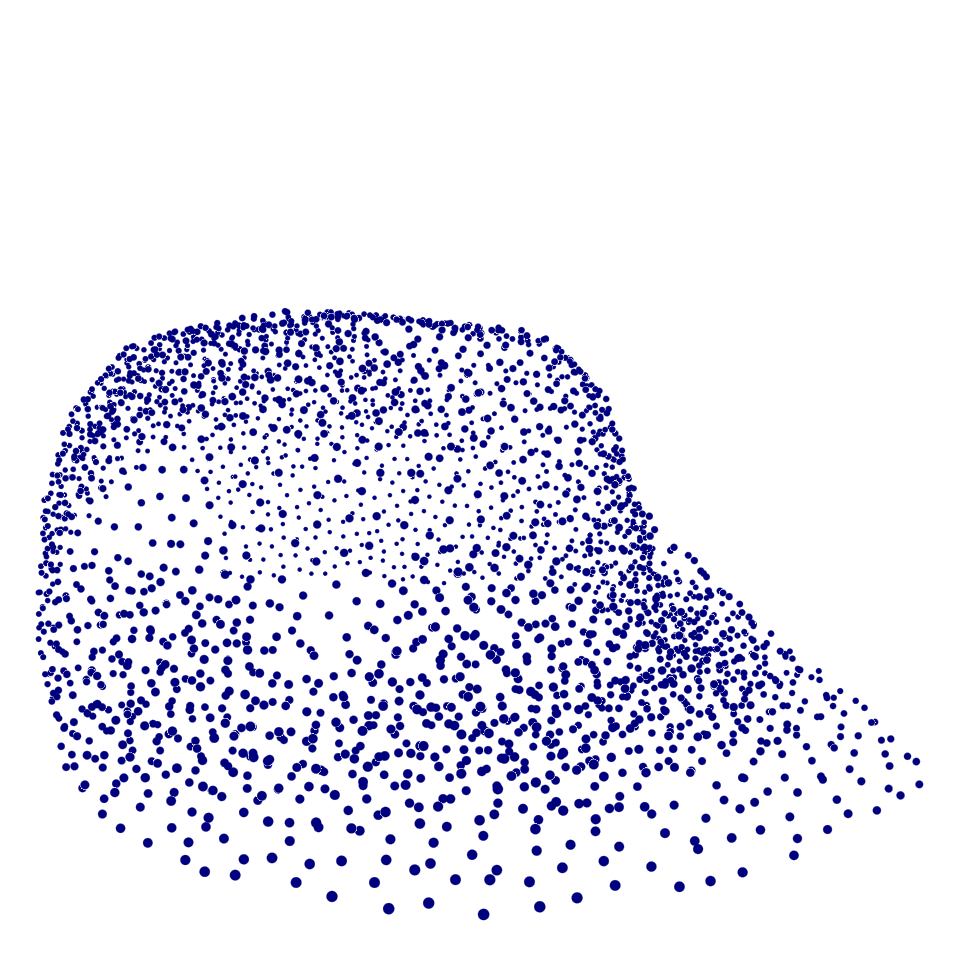}
			\end{minipage} &
			\begin{minipage}[b]{.11\textwidth}
				\centering
				\includegraphics[height=\linewidth]{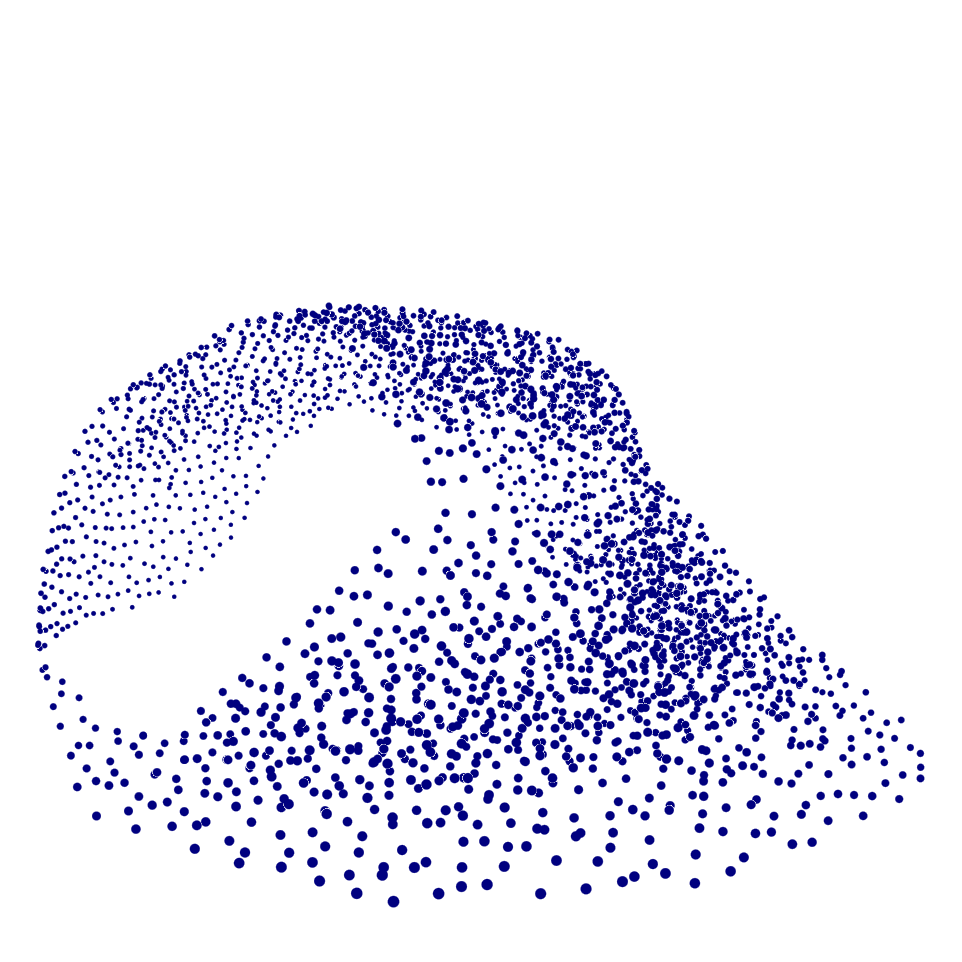}
			\end{minipage} &
			\begin{minipage}[b]{.11\textwidth}
				\centering
				\includegraphics[height=\linewidth]{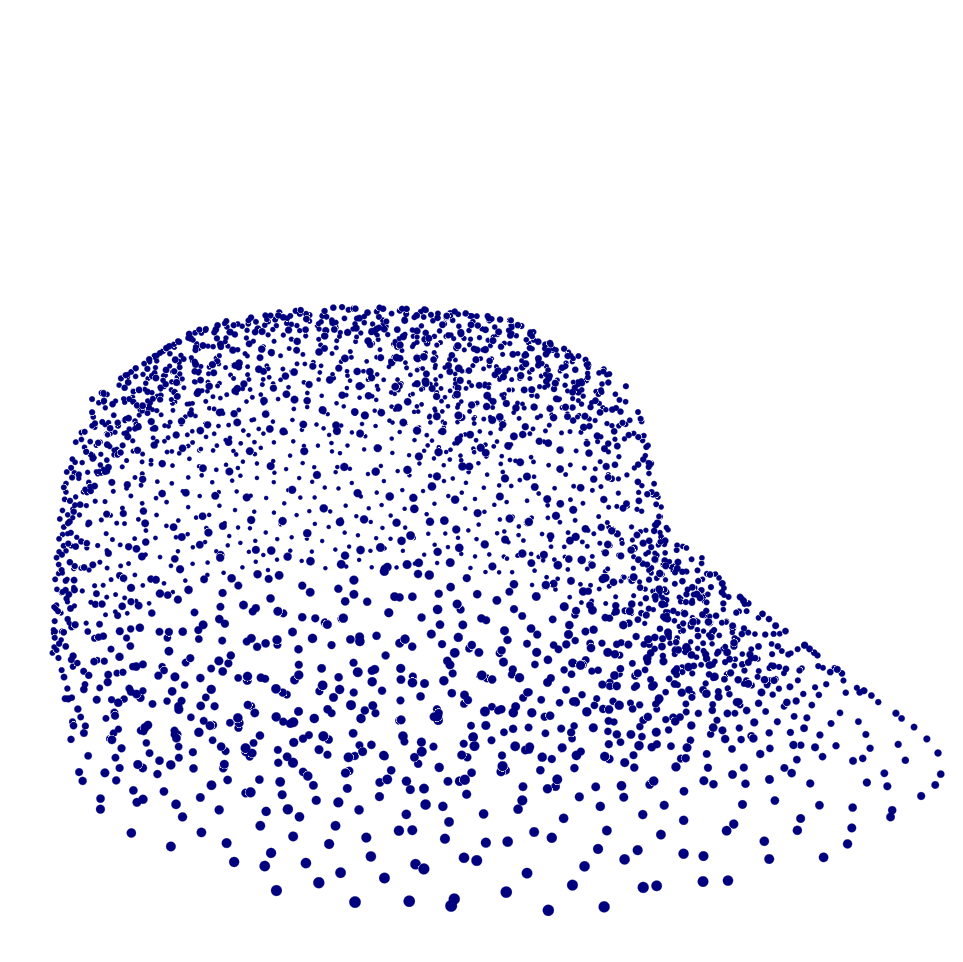}
			\end{minipage} &
			\begin{minipage}[b]{.11\textwidth}
				\centering
				\includegraphics[height=\linewidth]{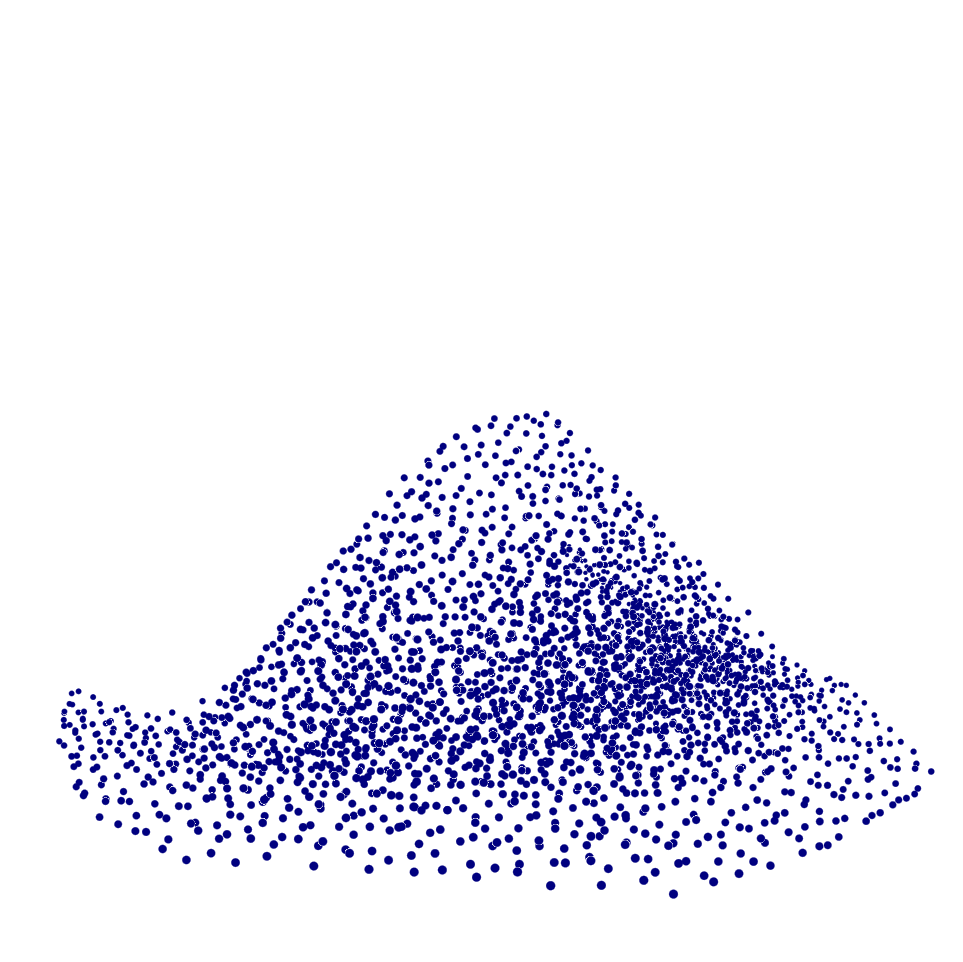}
			\end{minipage} &
			\begin{minipage}[b]{.11\textwidth}
				\centering
				\includegraphics[height=\linewidth]{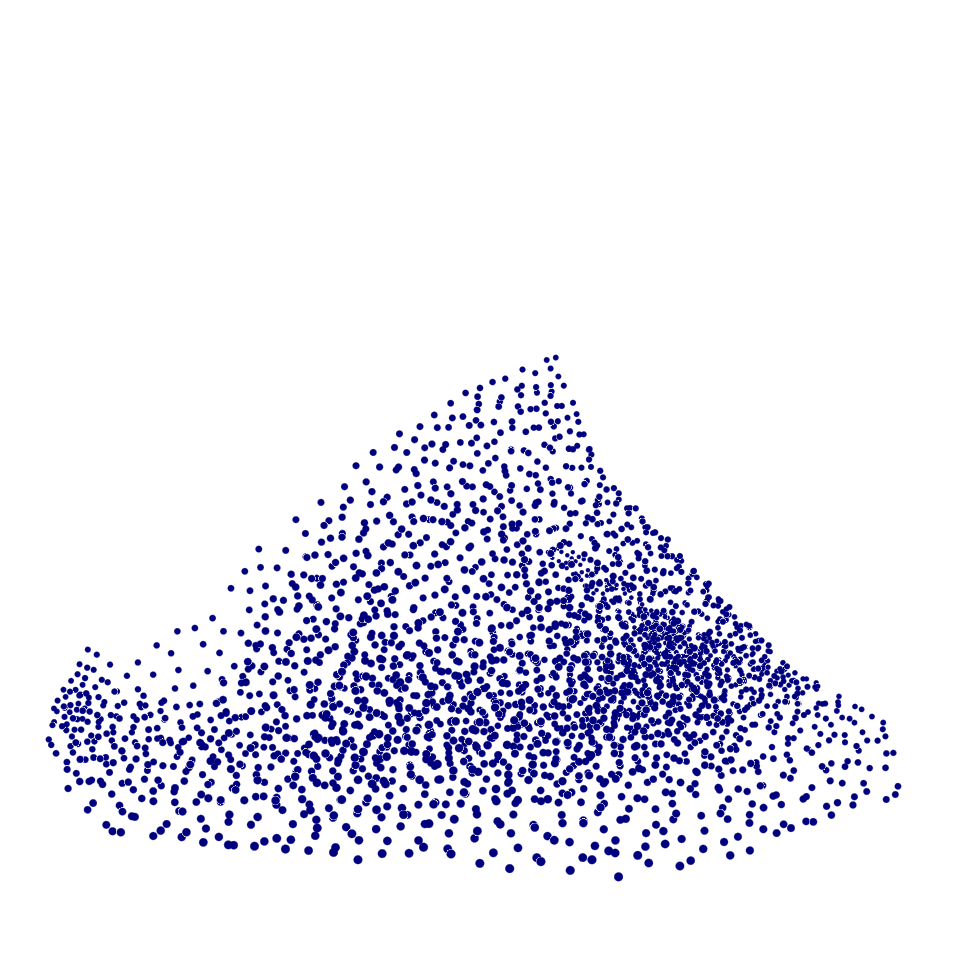}
			\end{minipage}\\
			\hline
			
			\begin{minipage}[b]{.11\textwidth}	
				\vspace{0.1mm}
				\centering
				\includegraphics[height=\linewidth]{Chair_OriN00.png}
			\end{minipage}&
			\begin{minipage}[b]{.11\textwidth}
				\centering
				\includegraphics[height=\linewidth]{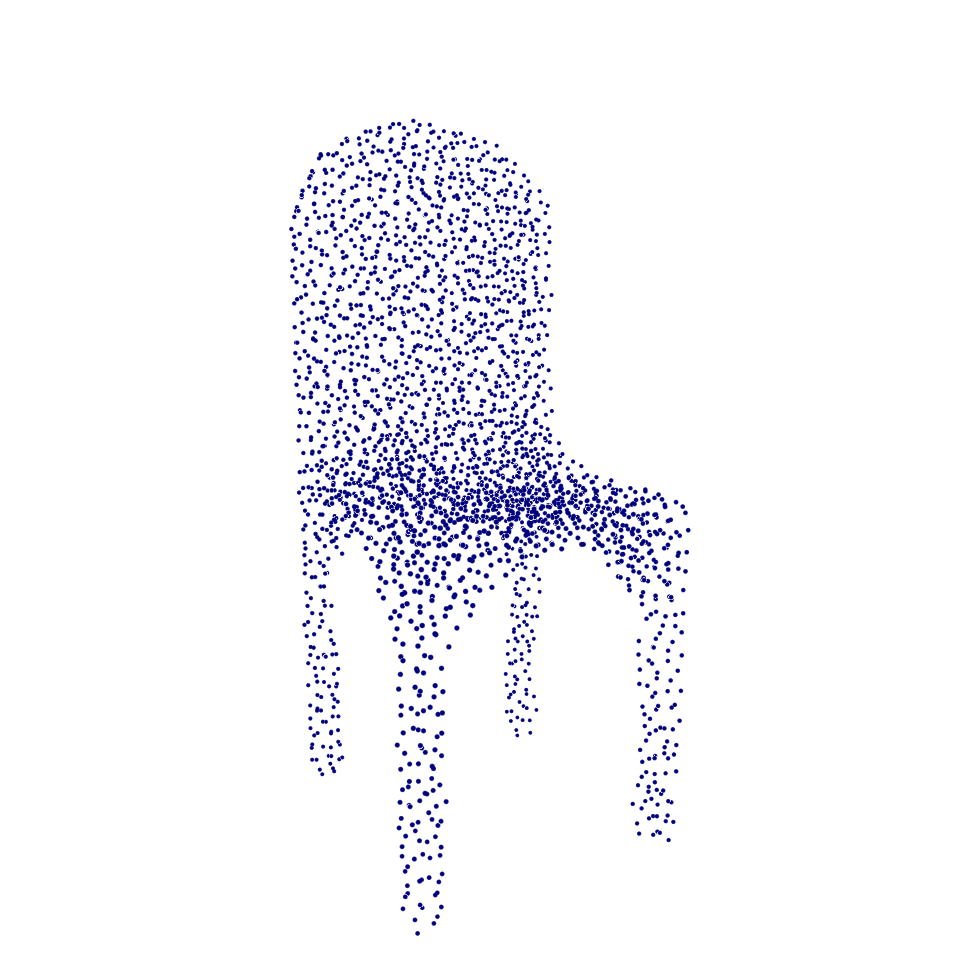}
			\end{minipage} &
			\begin{minipage}[b]{.11\textwidth}
				\centering
				\includegraphics[height=\linewidth]{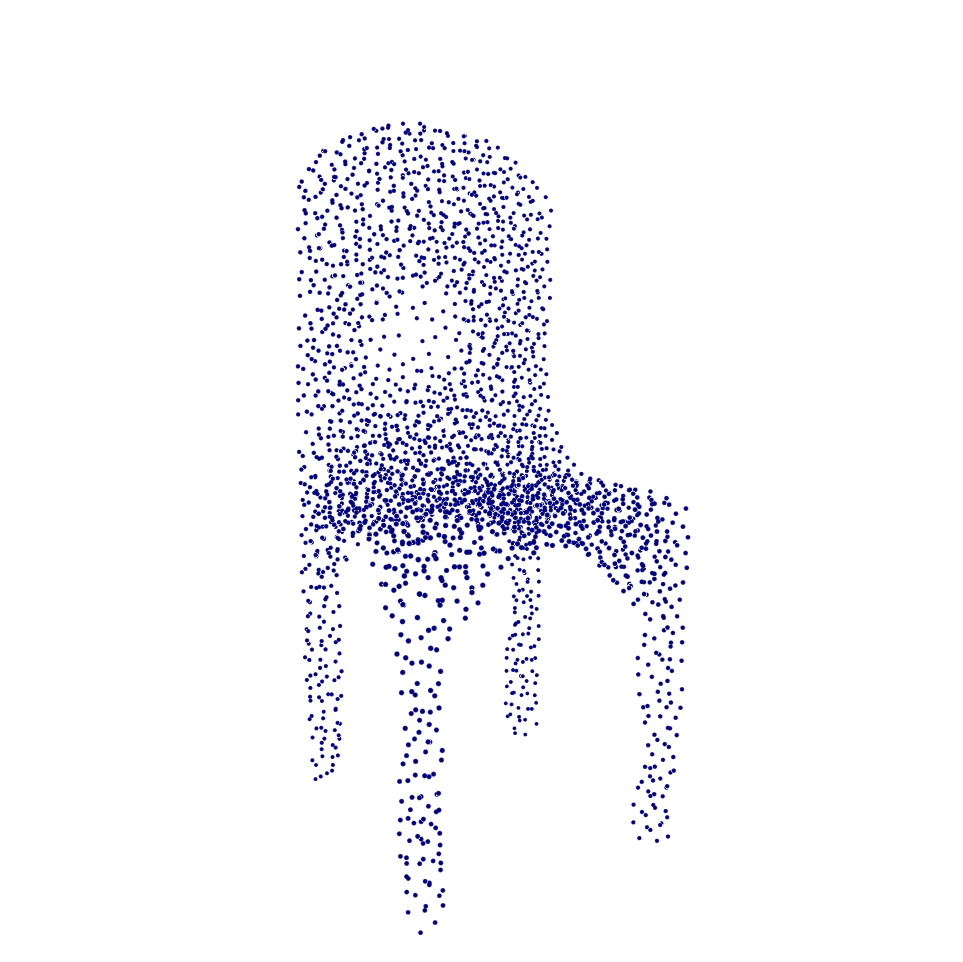}
			\end{minipage} &
			\begin{minipage}[b]{.11\textwidth}
				\centering
				\includegraphics[height=\linewidth]{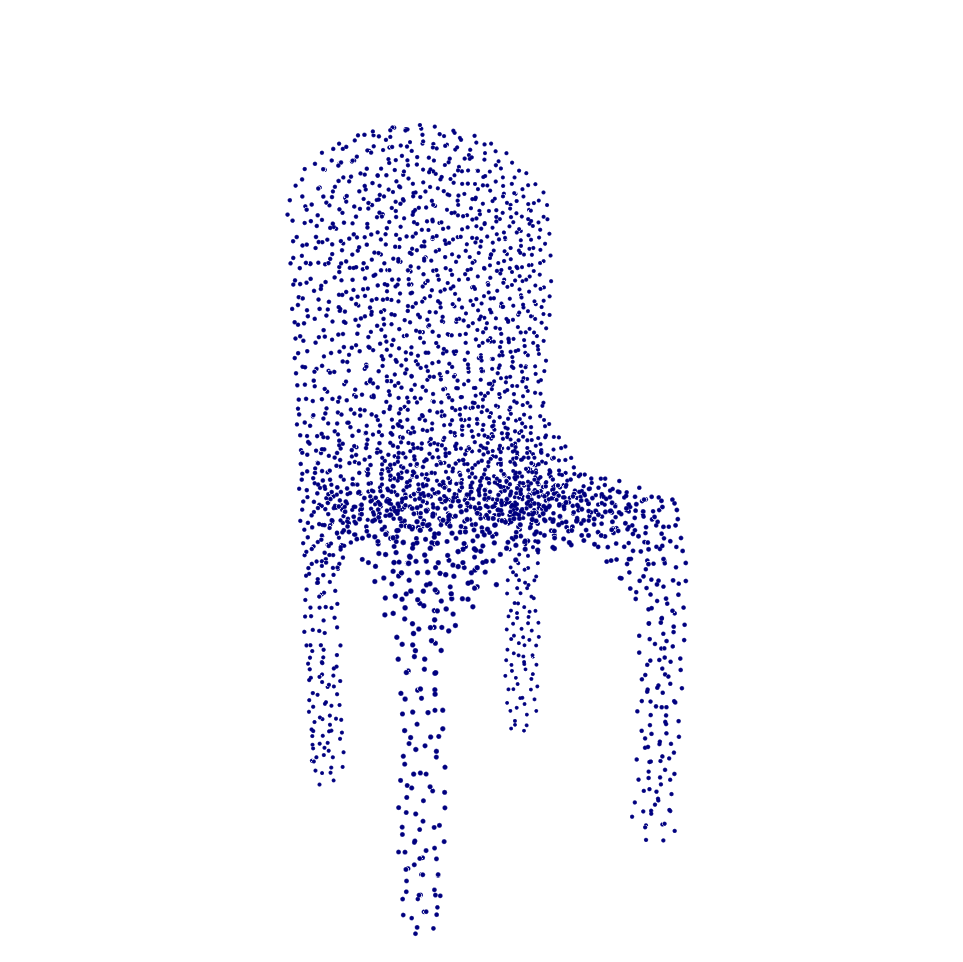}
			\end{minipage} &
			\begin{minipage}[b]{.11\textwidth}
				\centering
				\includegraphics[height=\linewidth]{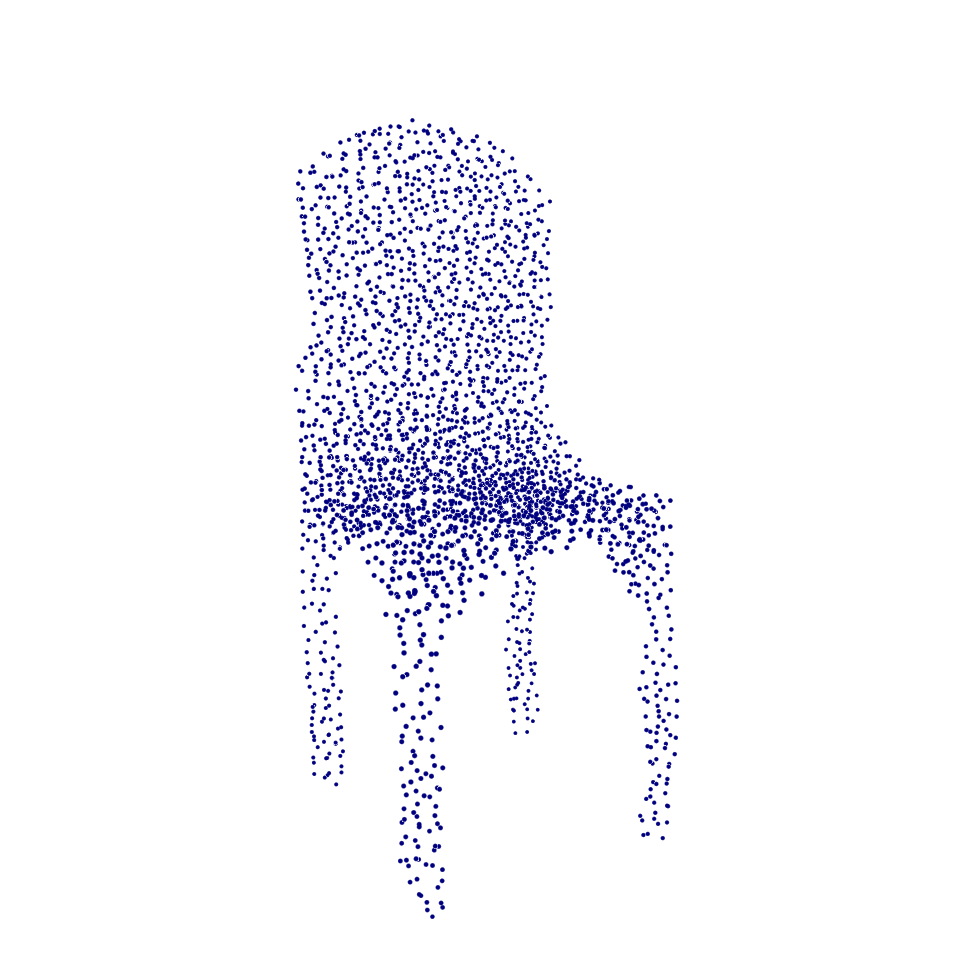}
			\end{minipage} &
			\begin{minipage}[b]{.11\textwidth}
				\centering
				\includegraphics[height=\linewidth]{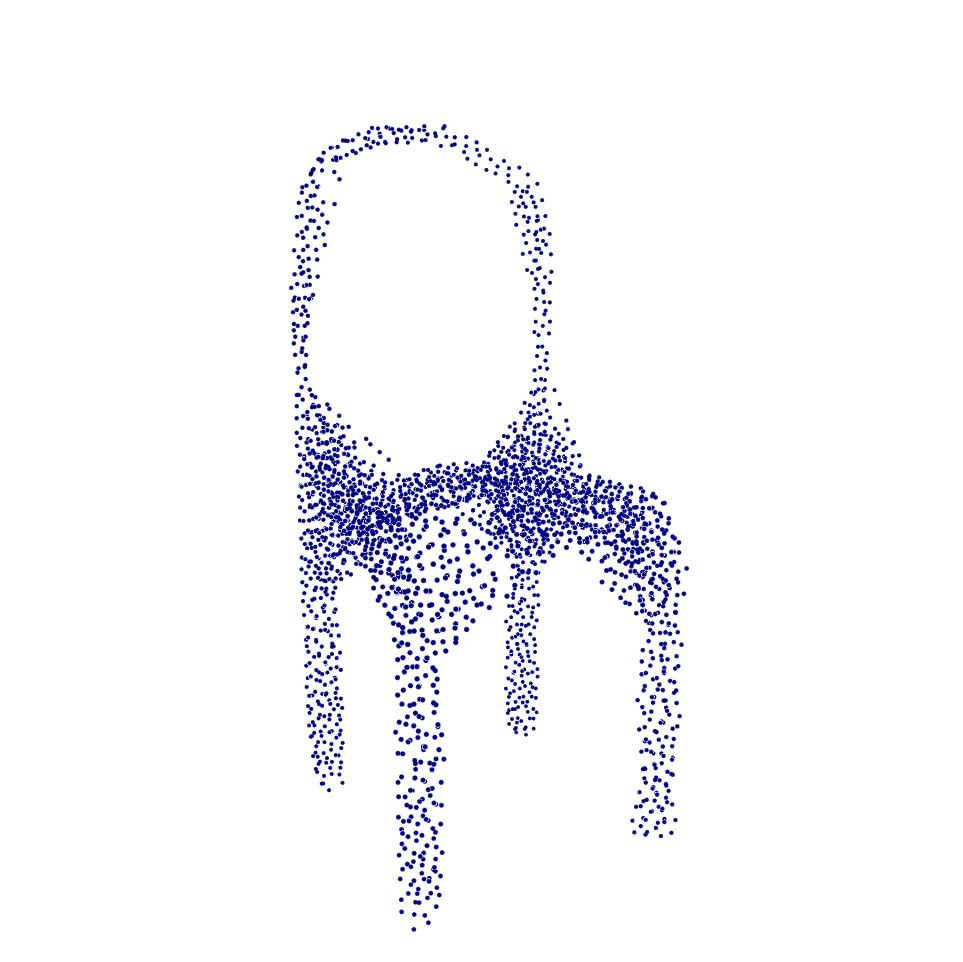}
			\end{minipage} &
			\begin{minipage}[b]{.11\textwidth}
				\centering
				\includegraphics[height=\linewidth]{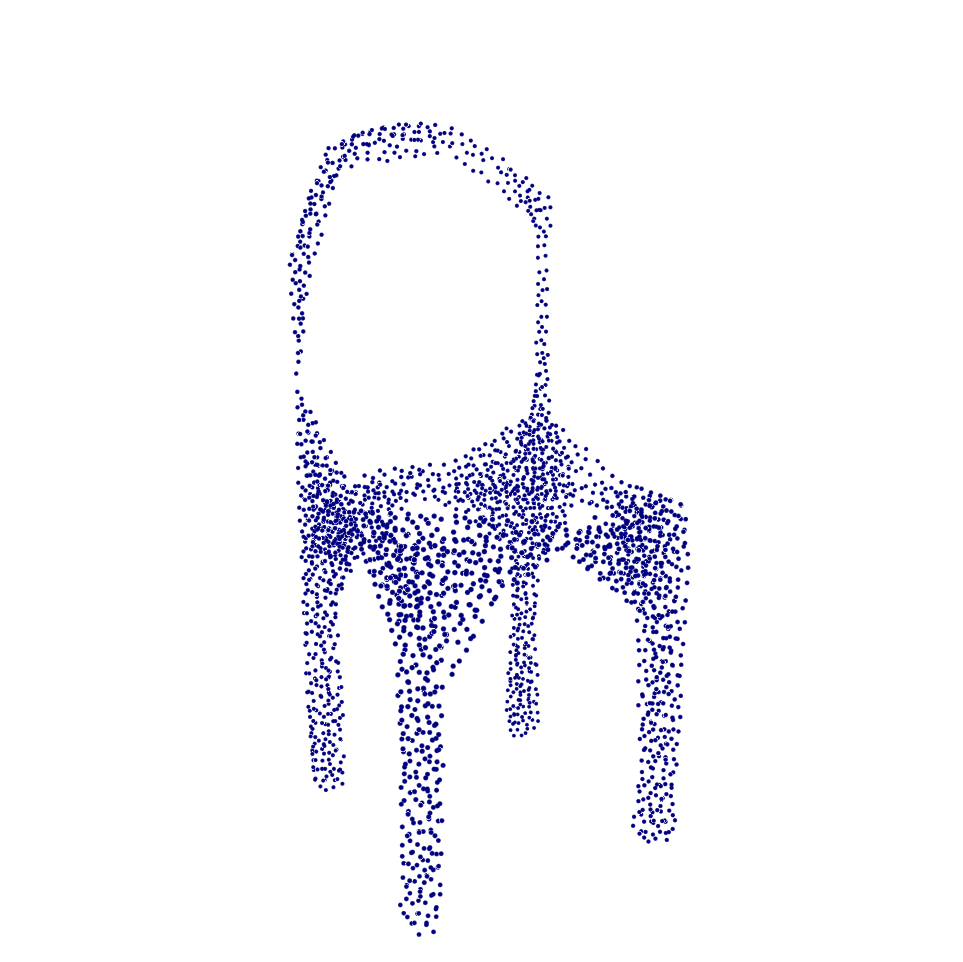}
			\end{minipage}\\
			\hline
			
			\begin{minipage}[b]{.11\textwidth}
				\vspace{0.2mm}	
				\centering
				\includegraphics[height=\linewidth]{Laptop_OriN00.png}
			\end{minipage}&
			\begin{minipage}[b]{.11\textwidth}
				\centering
				\includegraphics[height=\linewidth]{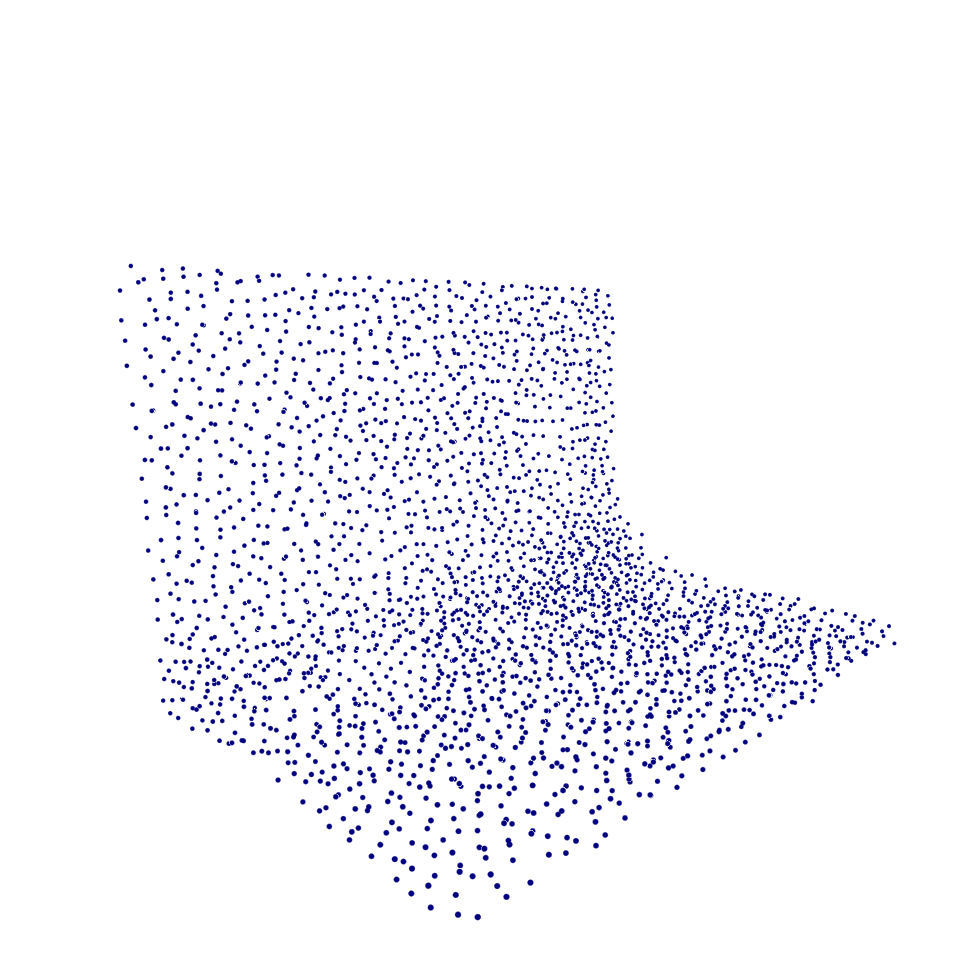}
			\end{minipage} &
			\begin{minipage}[b]{.11\textwidth}
				\centering
				\includegraphics[height=\linewidth]{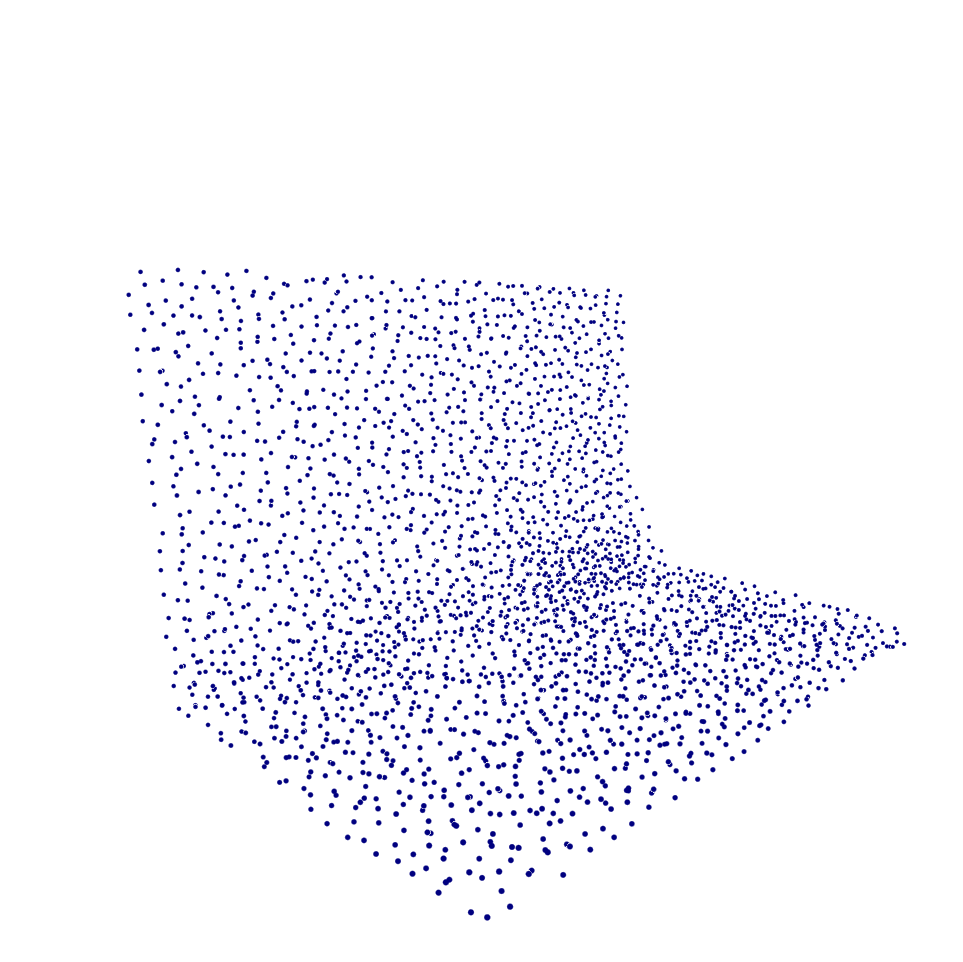}
			\end{minipage} &
			\begin{minipage}[b]{.11\textwidth}
				\centering
				\includegraphics[height=\linewidth]{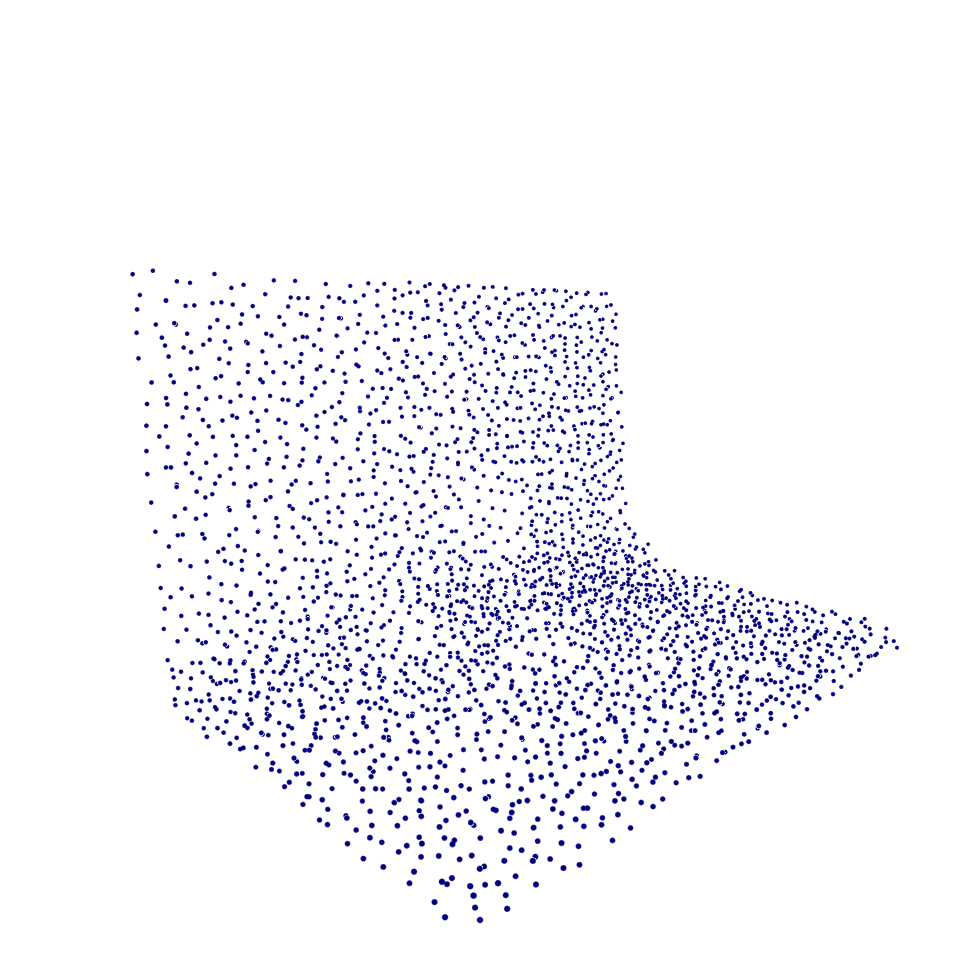}
			\end{minipage} &
			\begin{minipage}[b]{.11\textwidth}
				\centering
				\includegraphics[height=\linewidth]{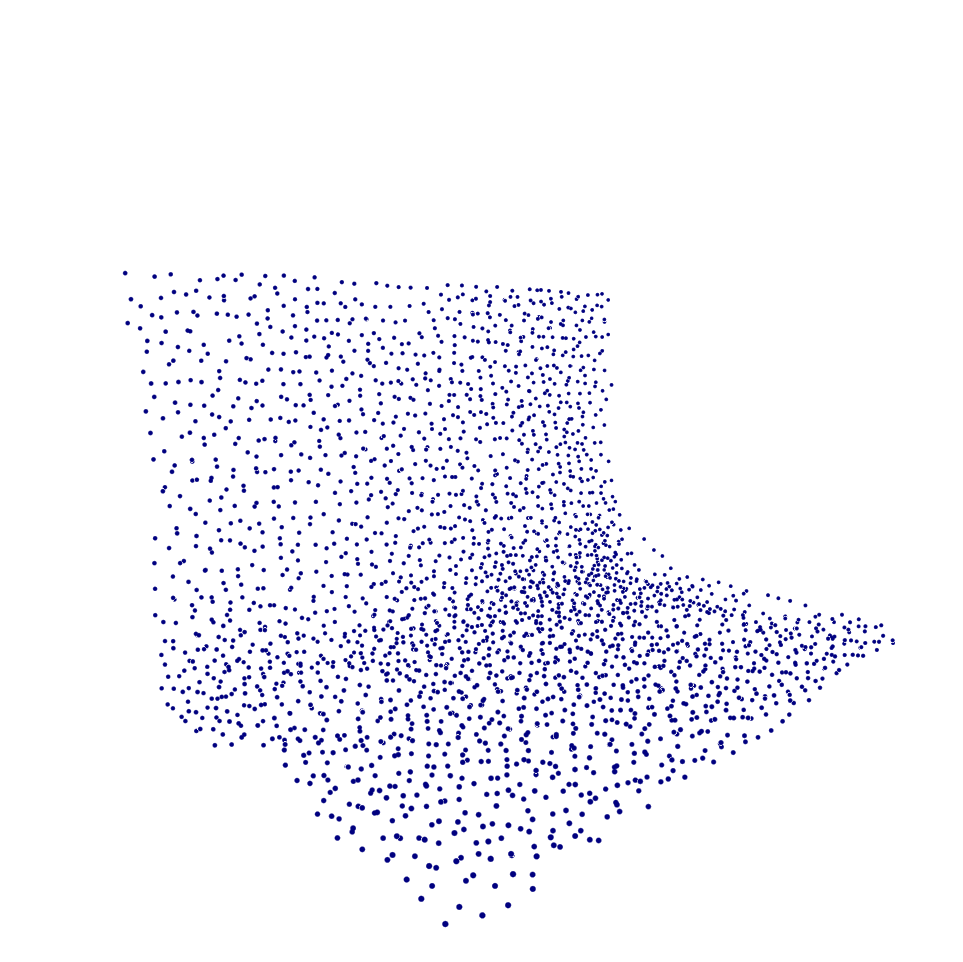}
			\end{minipage} &
			\begin{minipage}[b]{.11\textwidth}
				\centering
				\includegraphics[height=\linewidth]{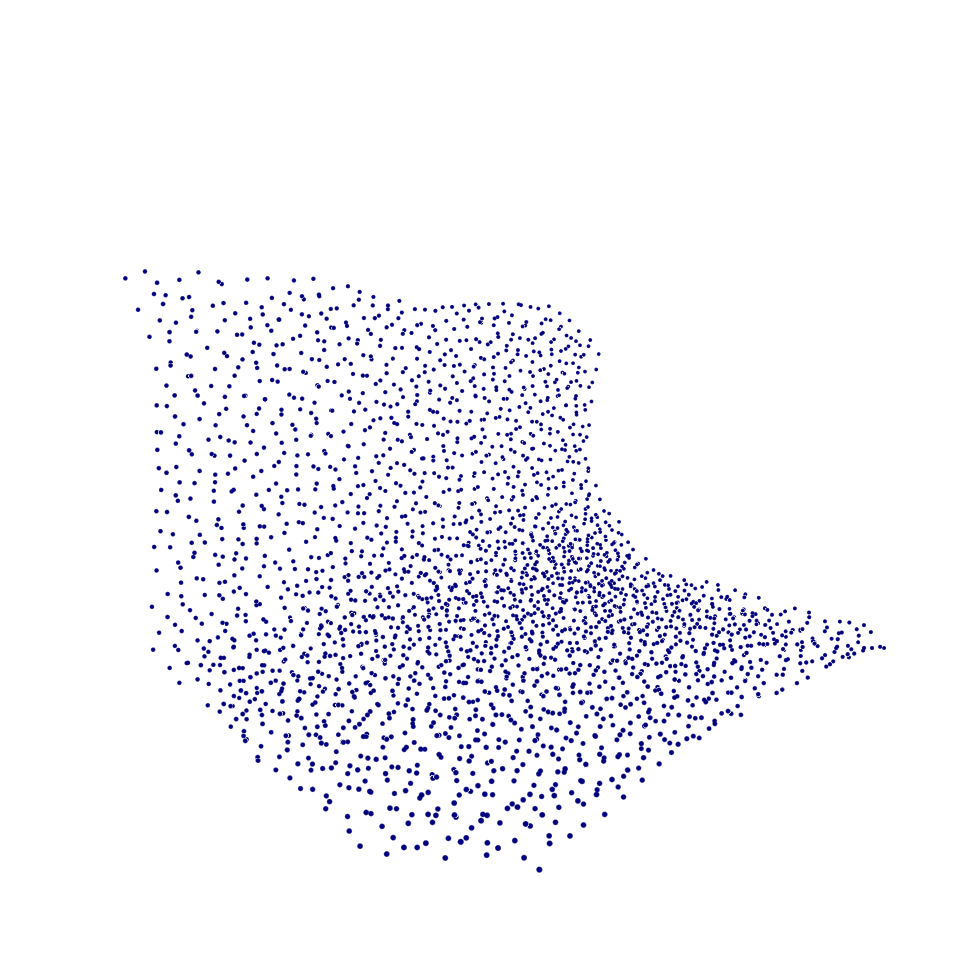}
			\end{minipage} &
			\begin{minipage}[b]{.11\textwidth}
				\centering
				\includegraphics[height=\linewidth]{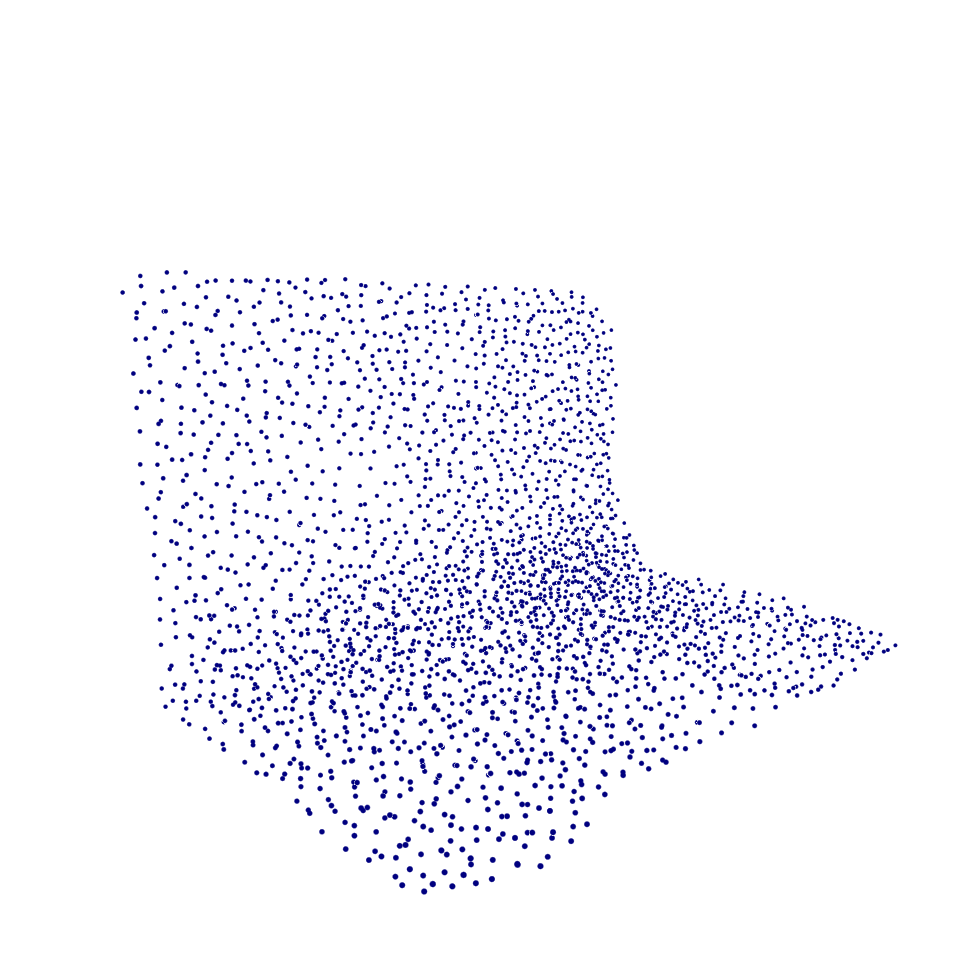}
			\end{minipage}\\
			\hline
			
			\begin{minipage}[b]{.11\textwidth}
				\vspace{0.1mm}
				\includegraphics[height=\linewidth]{Rocket_oriN00.png}
			\end{minipage}&
			\begin{minipage}[b]{.11\textwidth}
				\centering
				\includegraphics[height=\linewidth]{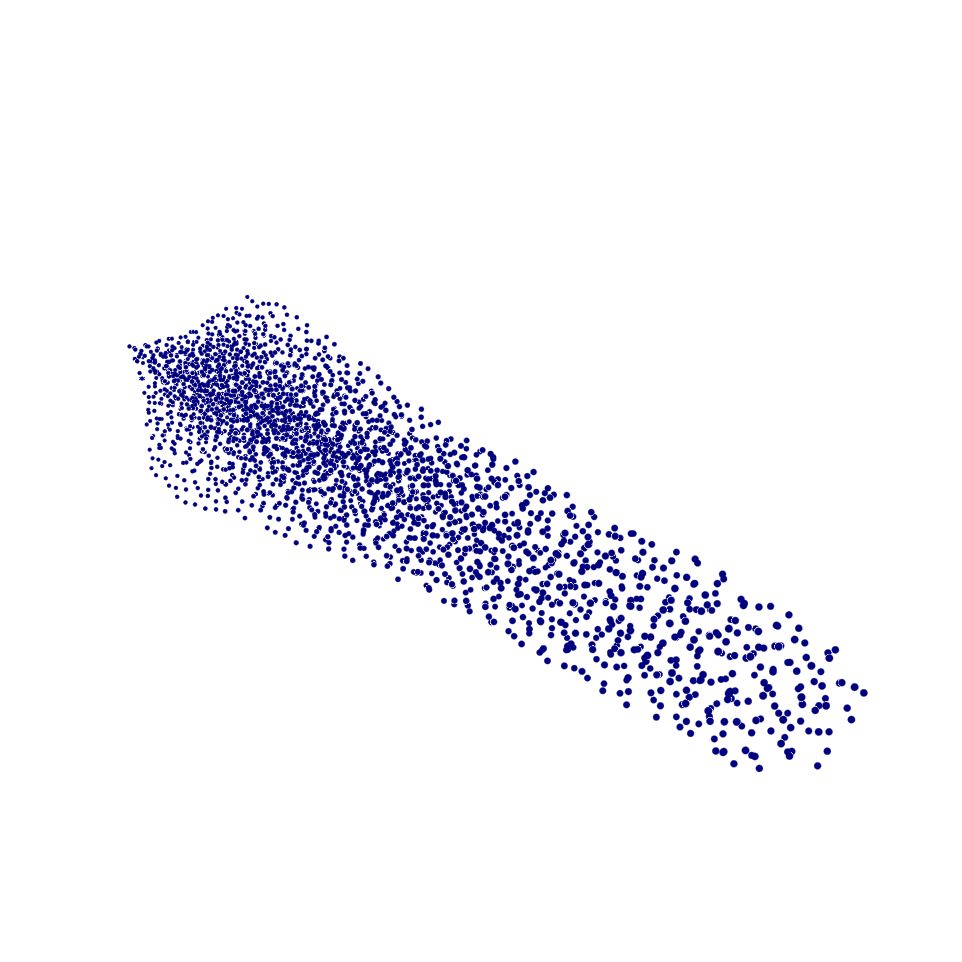}
			\end{minipage} &
			\begin{minipage}[b]{.11\textwidth}
				\centering
				\includegraphics[height=\linewidth]{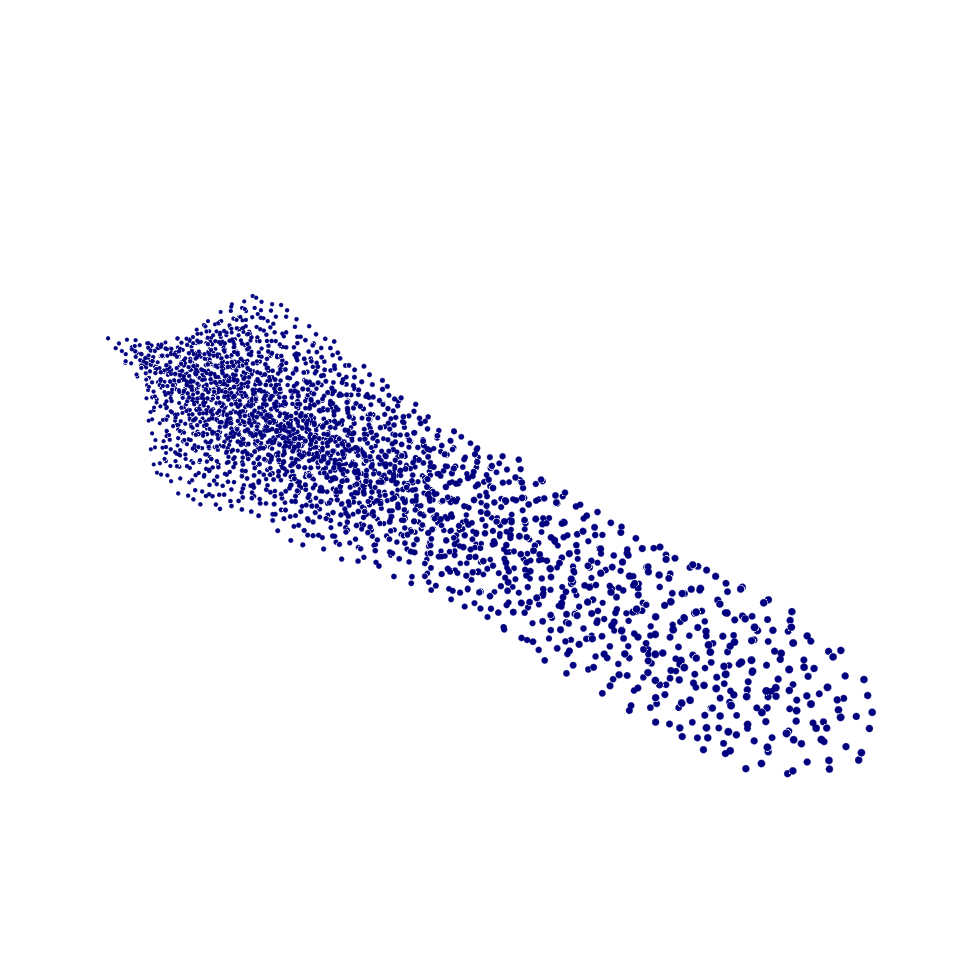}
			\end{minipage} &
			\begin{minipage}[b]{.11\textwidth}
				\centering
				\includegraphics[height=\linewidth]{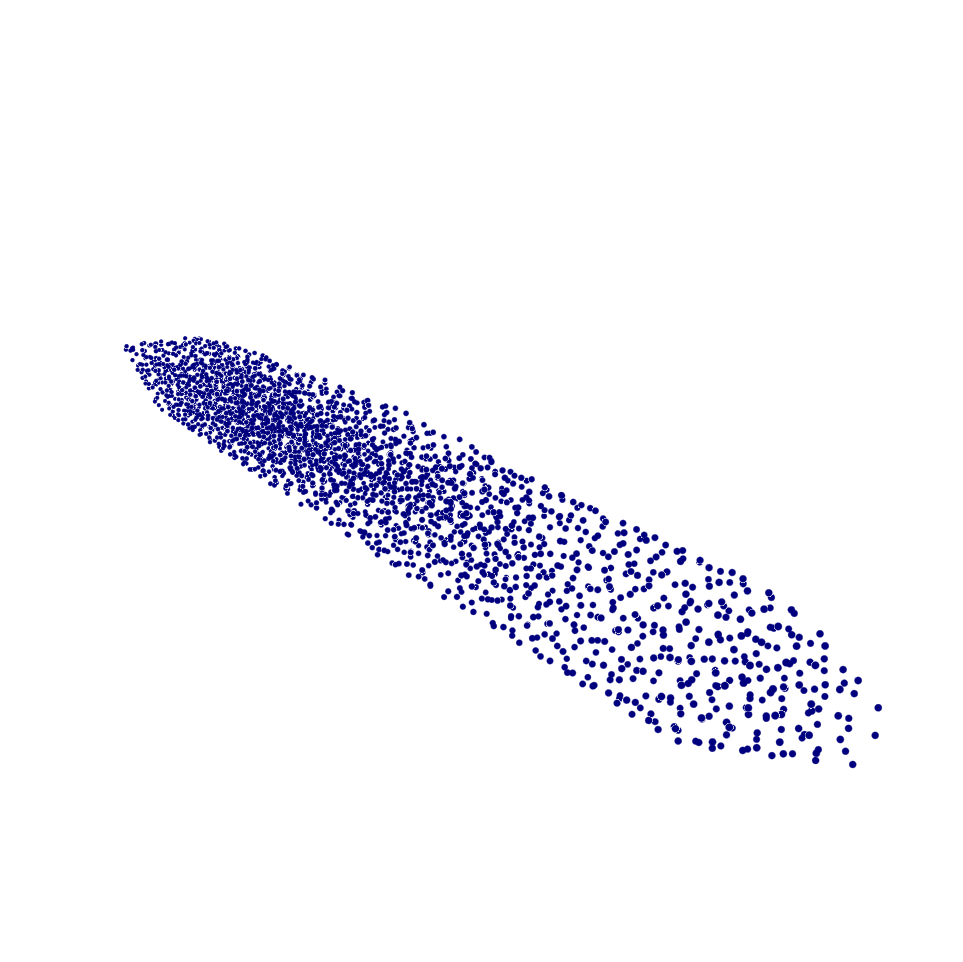}
			\end{minipage} &
			\begin{minipage}[b]{.11\textwidth}
				\centering
				\includegraphics[height=\linewidth]{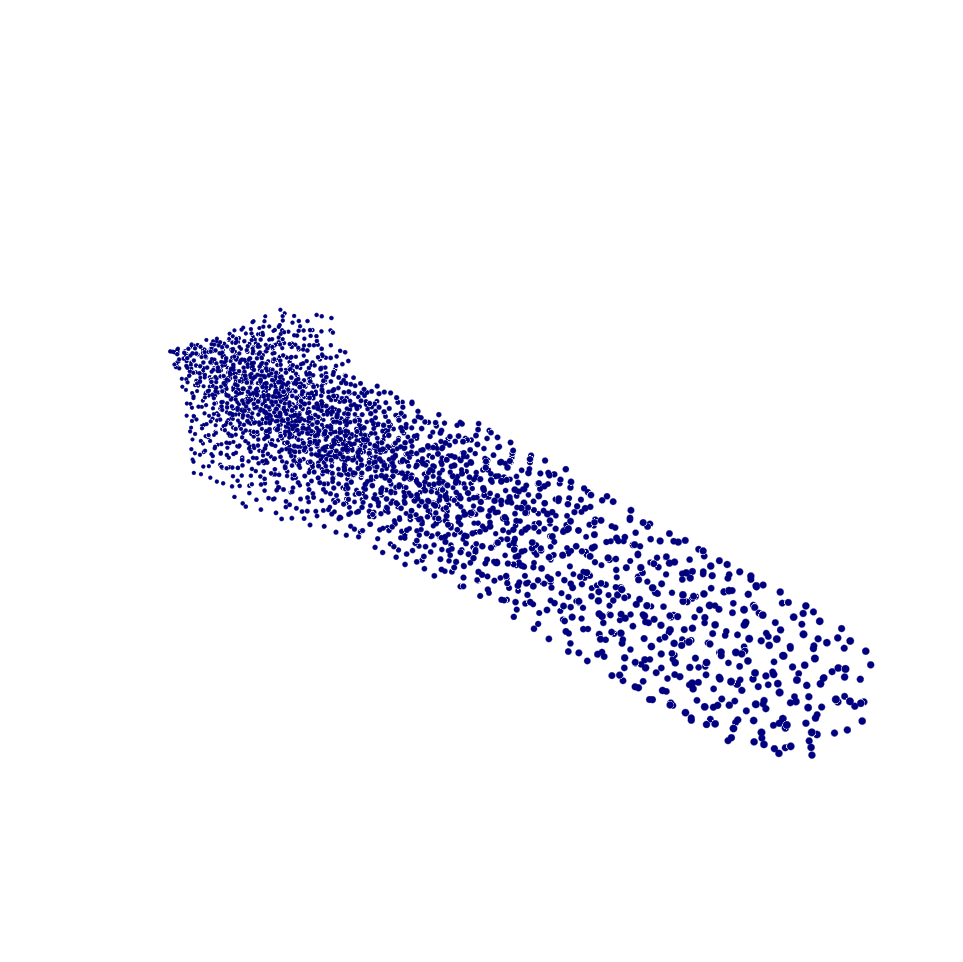}
			\end{minipage} &
			\begin{minipage}[b]{.11\textwidth}
				\centering
				\includegraphics[height=\linewidth]{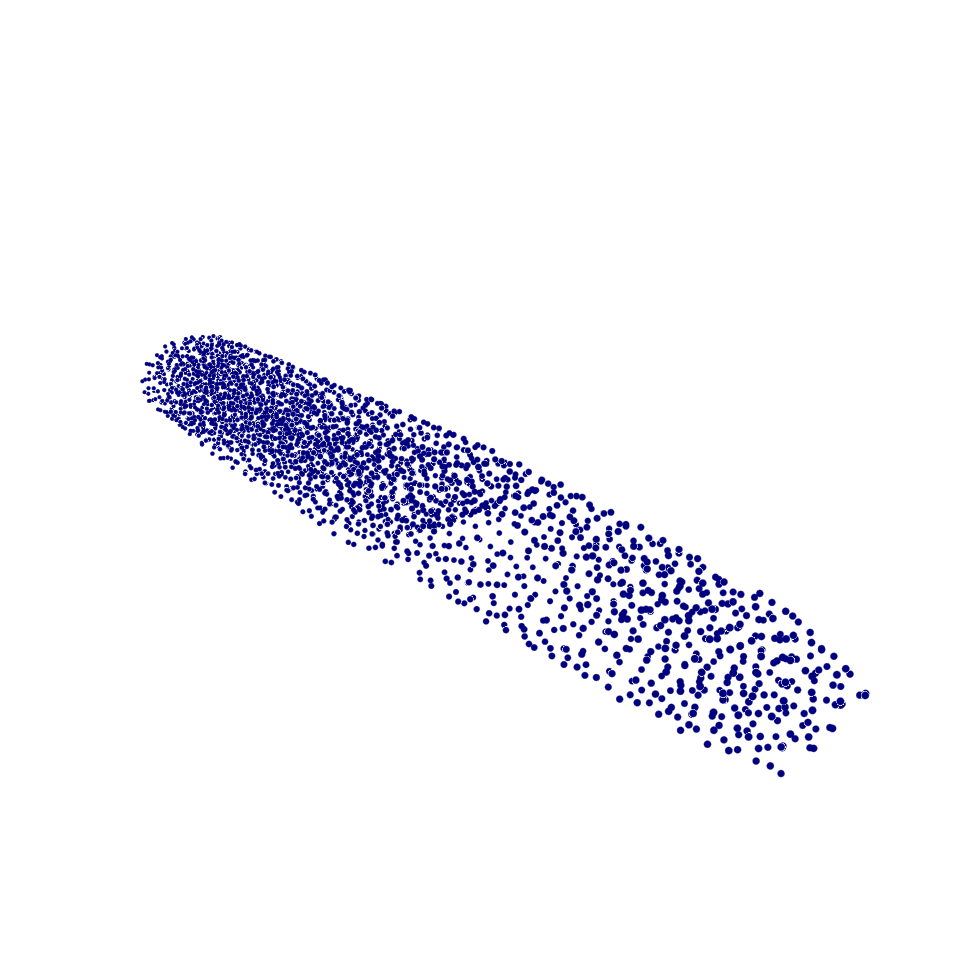}
			\end{minipage} &
			\begin{minipage}[b]{.11\textwidth}
				\centering
				\includegraphics[height=\linewidth]{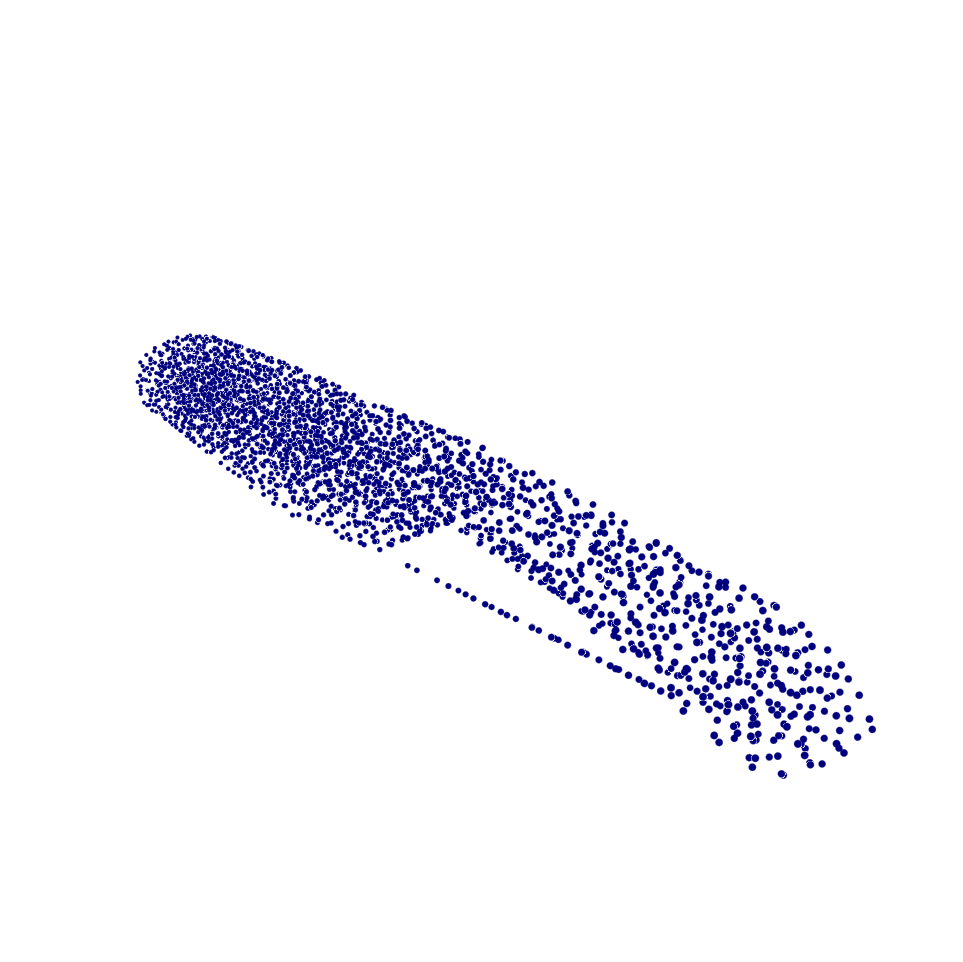}
			\end{minipage}\\
			\hline
			
			\begin{minipage}[b]{.11\textwidth}
				\vspace{0.1mm}
				\includegraphics[height=\linewidth]{Skateboard_OriN00.png}
			\end{minipage}&
			\begin{minipage}[b]{.11\textwidth}
				\centering
				\includegraphics[height=\linewidth]{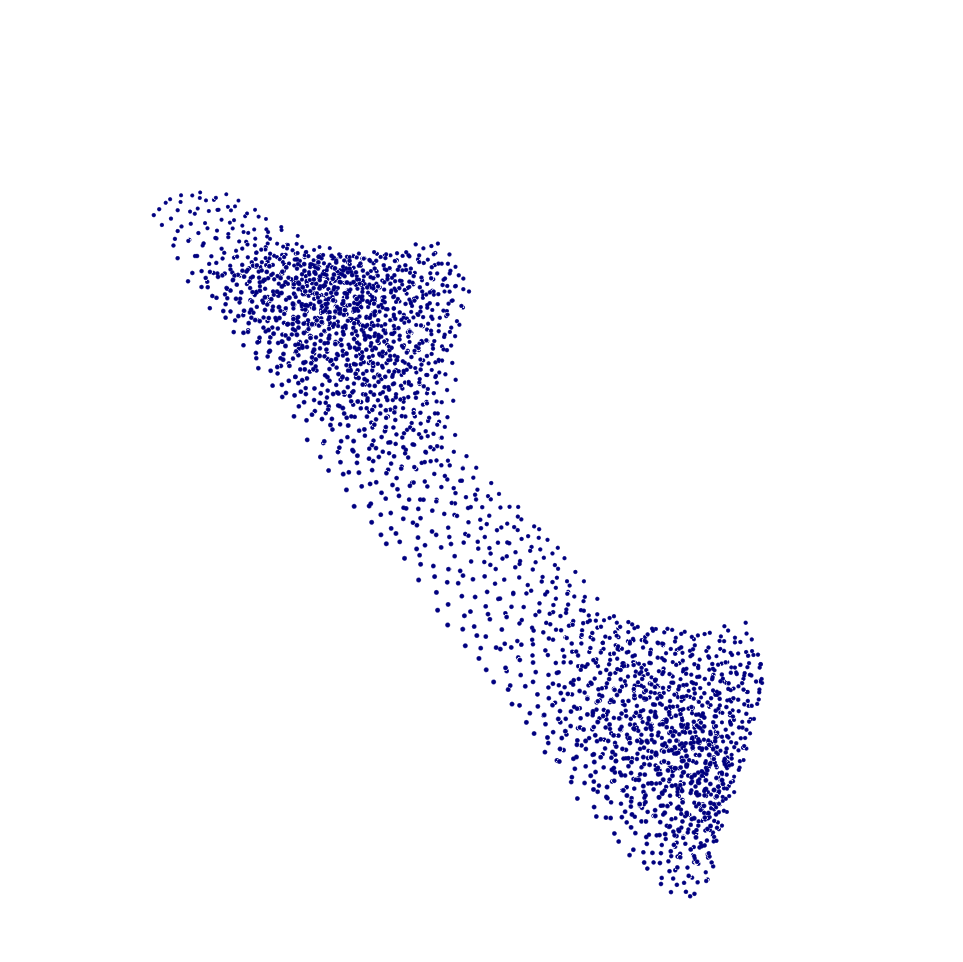}
			\end{minipage} &
			\begin{minipage}[b]{.11\textwidth}
				\centering
				\includegraphics[height=\linewidth]{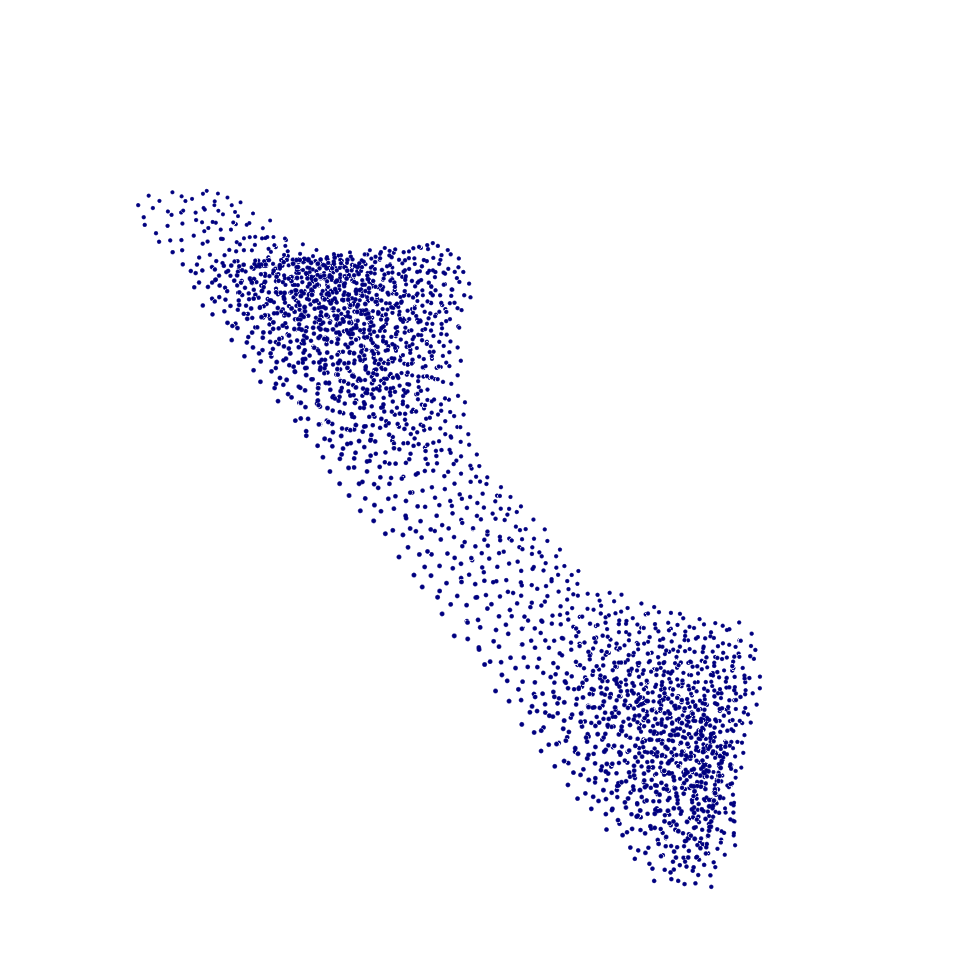}
			\end{minipage} &
			\begin{minipage}[b]{.11\textwidth}
				\centering
				\includegraphics[height=\linewidth]{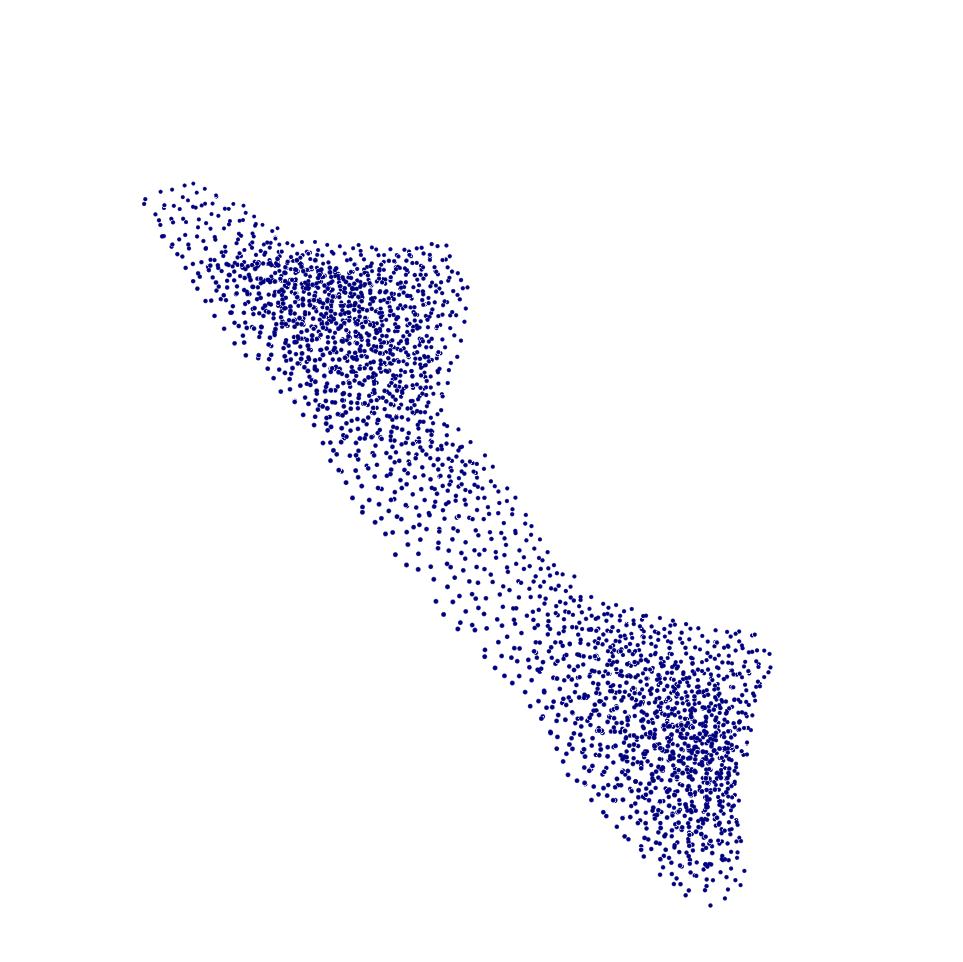}
			\end{minipage} &
			\begin{minipage}[b]{.11\textwidth}
				\centering
				\includegraphics[height=\linewidth]{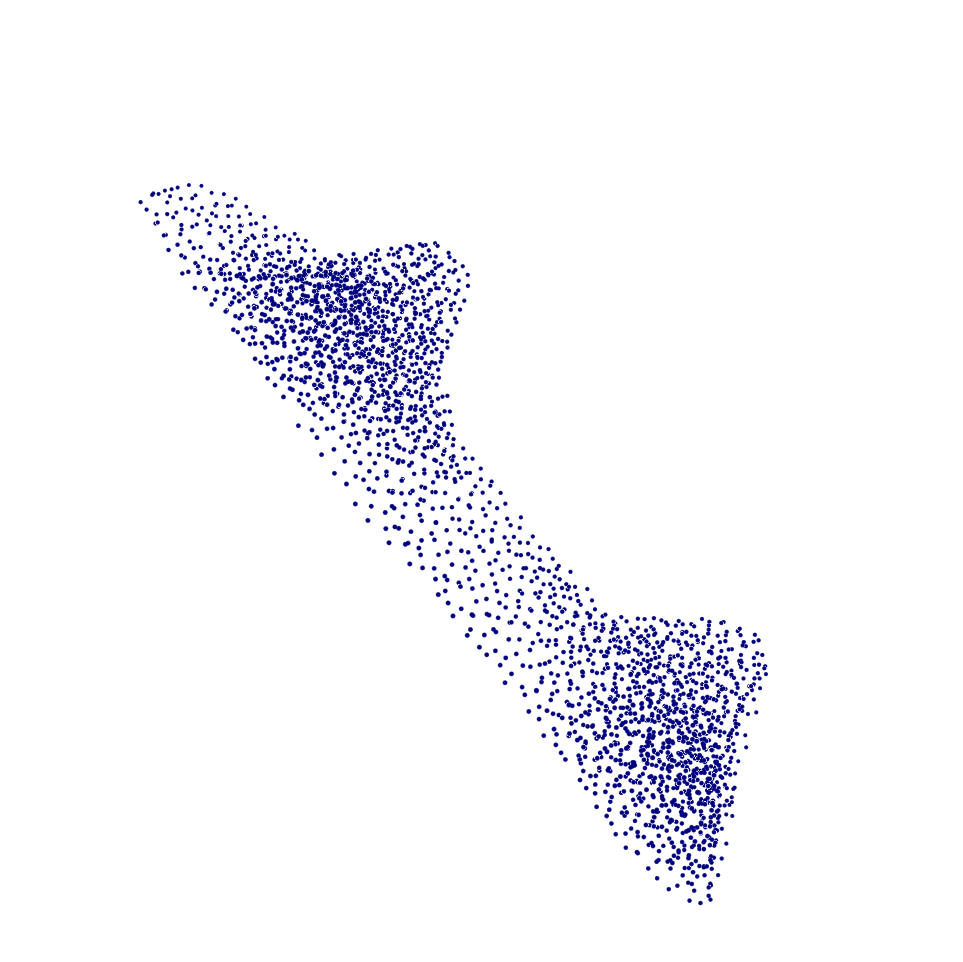}
			\end{minipage} &
			\begin{minipage}[b]{.11\textwidth}
				\centering
				\includegraphics[height=\linewidth]{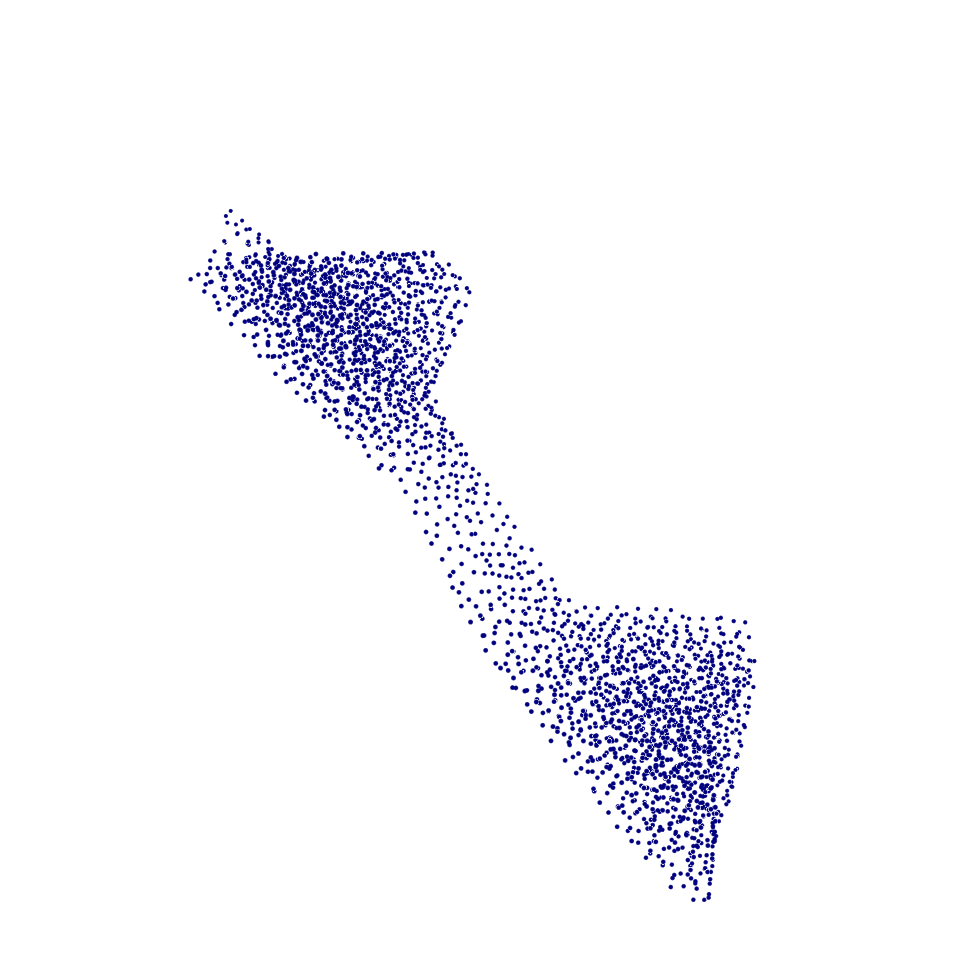}
			\end{minipage} &
			\begin{minipage}[b]{.11\textwidth}
				\centering
				\includegraphics[height=\linewidth]{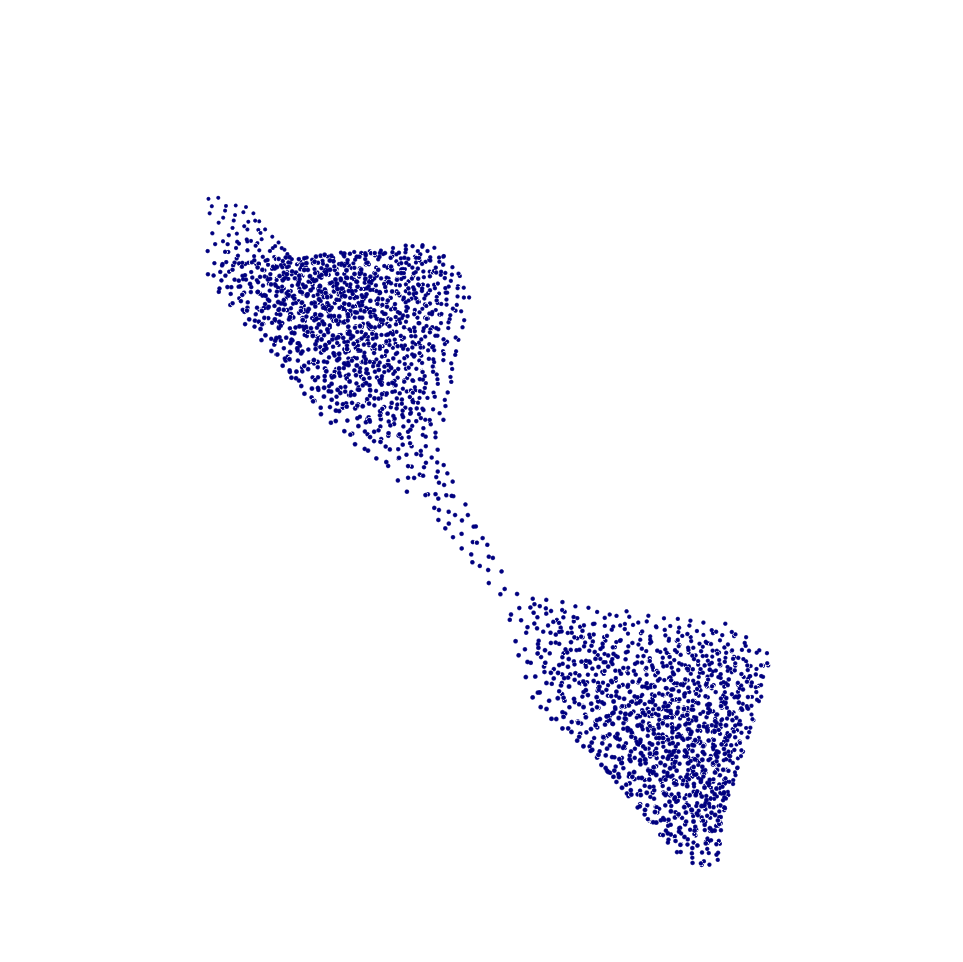}
			\end{minipage}\\
			\hline
			
		\end{tabular}
		\caption{Recovered point clouds.}
	\end{subtable}
	\caption{Example of original, resampled and recovered point clouds with resampling ratio $\alpha=0.2$.}
	\label{table:visExample}
	\vspace*{-4mm}
\end{figure*}

\begin{figure*}[bth]
	\begin{subfigure}{.3\textwidth}
		\centering
		\includegraphics[width=\linewidth]{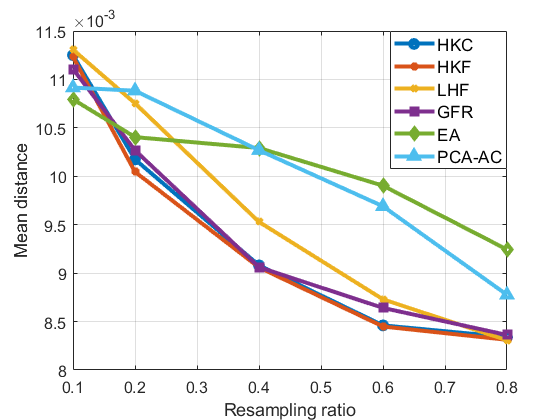}
		\caption{Mean distance against $\alpha$.}
	\end{subfigure}
	\hspace{5mm}
	\begin{subfigure}{.3\textwidth}
		\centering
		\includegraphics[width=\linewidth]{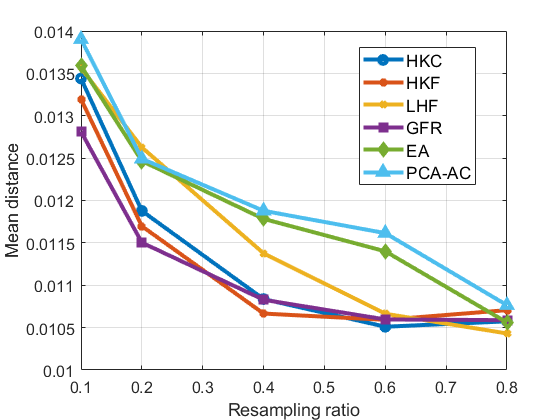}
		\caption{Average of distance and dual distance against $\alpha$.}
	\end{subfigure}
	\hspace{5mm}
	\begin{subfigure}{.3\textwidth}
		\centering
		\includegraphics[width=\linewidth]{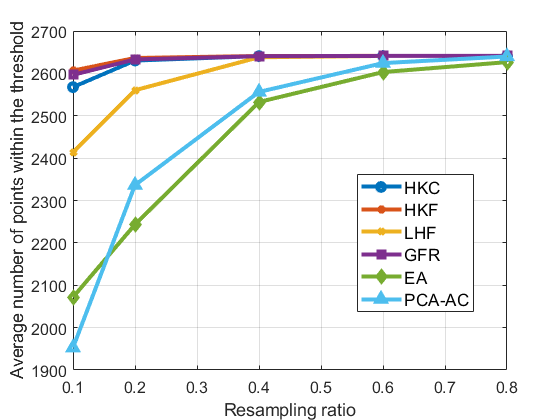}
		\caption{Average number of points in recovered point cloud $d_\theta$ against $\alpha$.}
	\end{subfigure}
	\caption{Plots of recovered accuracy against resampling ratio $\alpha$ of all methods}
	\label{fig:Curves}
\end{figure*}

\subsubsection{
\textbf{Dense Point Cloud Recovery}} 
A typical method for dense
point cloud recovery consists two steps: a) reconstructing 
the surface of object from the resampled point cloud; and b) sampling 
the reconstructed object surface to generate a
recovered point cloud. Since points of edge preserving resampled point 
clouds tend to concentrate near areas of high local variations, e.g.,
edges/corners, points of these resampled point clouds are not uniformly
distributed, as shown in {Fig.~\ref{fig:downsamplept}}. For this reason,
some generic surface reconstruction methods such as Poisson reconstruction \cite{e1} may perform poorly on such sparse point clouds. 
We must pay special attention to surface reconstruction methods
chosen for such type of resampled point cloud data. 

In order to reconstruct surfaces from edge preserving and sparsely
resampled point clouds, we propose to first construct
the alpha complex \cite{e2} from the resampled point cloud. 
To mitigate the potentially degrading impact of imperfect reconstruction, 
we decide to reconstruct six different surface models 
for each resampled point cloud
by applying different parameters. 
We then apply Poisson-disk resampling to sample the alpha complex 
to form a recovered point cloud. 
To further mitigate the effect due to the possible construction of 
extraneous surfaces absent from the original object,
we select a threshold distance $d_\theta$ 
three times the intrinsic resolution of the original point cloud. 
Using the threshold distance, we
only retain the best recovered point cloud 
which contains the largest number of points that are
within the threshold distance $d_\theta$ from the 
original point cloud.

\subsubsection{\textbf{Distance Between Point Clouds}}

To assess the quality of point cloud recovery, we 
need to define distances between the original and 
the recovered point clouds.
Let
$p_i$ denote a point in the original and $p_{c,j}$ denote a point
in the recovered point cloud. When
computing our distance between two point clouds, 
we neglect any distances between
point $p_i$ in the original point cloud and $p_{c,j}$ in the recovered point cloud such that the minimum distances $\min_j\|p_i-p_{c,j}\|$ and 
$\min_i\|p_i-p_{c,j}\|$ are greater than
$d_\theta$. 

We define a distance and a dual distance between the original and the recovered point cloud as
\begin{align}
D_0=\frac{1}{N_1} \sum_{i=1}^{N_1} \min_{j: \|p_i-p_{c,j}\|<d_\theta} \|p_i-p_{c,j}\|,\label{distance}\\
{\check D}_{0}=\frac{1}{N_2} \sum_{j=1}^{N_2}\min_{i: \|p_i-p_{c,j}\|<d_\theta}\|p_i-p_{c,j}\|,\label{dual_distance}
\end{align}
where $N_1$ is the number of points in the original point cloud
that satisfy $ \|p_i-p_{c,j}\|<d_\theta$ for some $p_{c,j}$
and $N_2$ is the number of points in the recovered point cloud
that satisfy $\|p_i-p_{c,j}\|<d_\theta$ for some $p_i$.
In other words, $D_0$ is the average distance for points 
that are 
in the original point cloud within $d_\theta$ from 
the closest points in the best recovered point cloud.
The dual distance ${\check D}_{0}$ is the average 
distance for points in the best recovered point cloud that are
within $d_\theta$ from their closest points in the original 
point cloud.

\begin{figure}
	\begin{subfigure}{.23\textwidth}
		\centering
		\includegraphics[width=\linewidth]{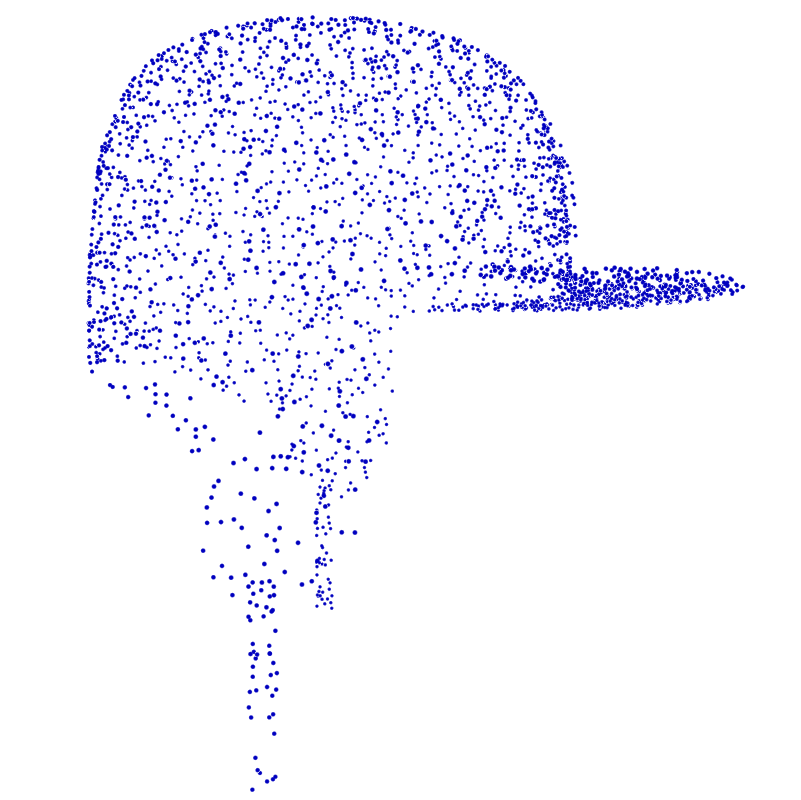}
		\caption{Original Point Cloud.}
	\end{subfigure}
	\begin{subfigure}{.23\textwidth}
		\centering
		\includegraphics[width=\linewidth]{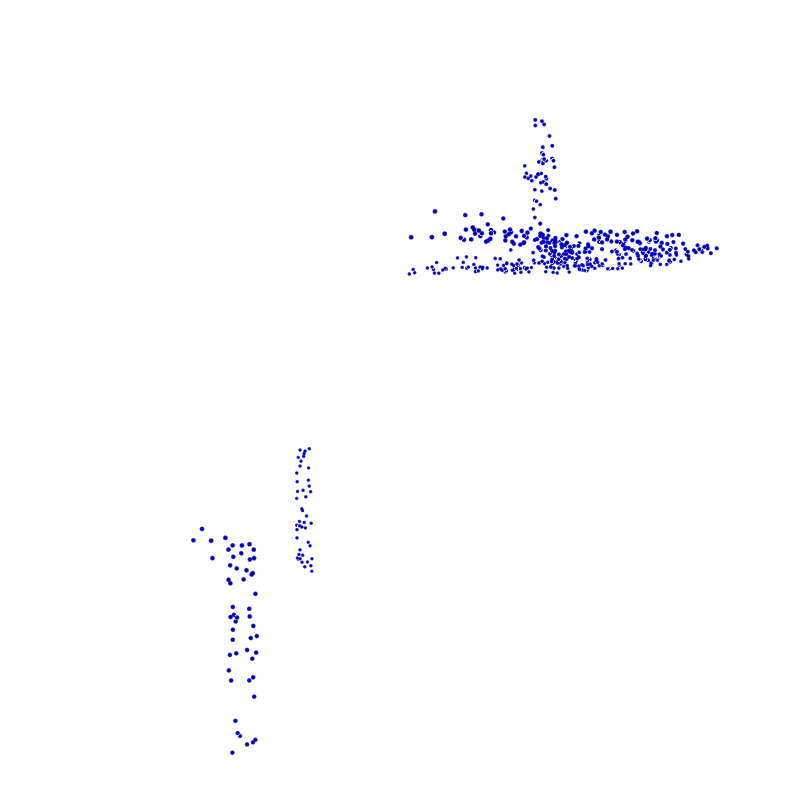}
		\caption{Resampled Point Cloud.}
	\end{subfigure}
	\caption{{Example of results from the EA method for $\alpha=0.2$.}}
	\label{fig:Eig1}
\end{figure}

\subsubsection{\textbf{Visual and Numerical Results}} 
We use six different categories of point clouds from ShapeNet 
\cite{c12} in our experiments. Similar to experiments discussed earlier,
we test our HKF method together
with the GSP-based GFR method in \cite{c3} plus
the EA and PCA-AC methods from \cite{c5}. 
For fairness, we use the
same resampling ratio for all the methods.

Our experiments follow the following steps.
First, we apply resampling and edge detection methods to 
calculate the resampled point clouds with resampling ratio $\alpha$.
Next, we apply our proposed recovery method to 
generate recovered point clouds. Based on
the resampled point clouds, we compute the numerical performance
metrics for different algorithms in comparison.
We measure the performance under three metrics: 1) distance defined in Eq. (\ref{distance}); 2) {average of distance and} dual distance as defined in Eq. (\ref{distance}) and Eq. (\ref{dual_distance}), respectively; 
and 3) average number $N_1$ of points within the threshold $d_\theta$ between the original and recovered point cloud. 
Smaller distance and larger number of points within the threshold indicate better performance.

To start, {Fig.~\ref{table:visExample}} provides a set of
the resampled point clouds and their corresponding recovered ones 
from different methods. Visual inspection shows that our proposed
HKC, HKF, and LHF algorithms generally deliver consistently strong
results in resampling and recovery of point clouds, regardless of the
dataset under study.

\begin{table*}[t]
	\centering
	\begin{subtable}{0.9\linewidth}
		\captionsetup{labelformat=parens}
		\centering
		\begin{tabular}{|c||c|c|c|c|c|c|}
			\hline
			Categories & \tabincell{c}{HKC} & \tabincell{c}{HKF} & \tabincell{c}{LHF} & \tabincell{c}{GFR} & \tabincell{c}{EA} & \tabincell{c}{PCA-AC}\\
			\hline
			Cap & 0.0111 & 0.0101 & 0.0117 & 0.0102 & \textbf{0.0087} & 0.0115 \\
			\hline
			Chair & \textbf{0.0111} & 0.0113 & 0.0116 & 0.0118 & 0.0125 & 0.0126 \\
			\hline
			Laptop & 0.0106 & 0.0106 & \textbf{0.0103} & 0.0105 & 0.0110 & 0.0110 \\
			\hline
			Mug & \textbf{0.0134} & \textbf{0.0134} & 0.0150 & 0.0141 & 0.0150 & 0.0147 \\
			\hline
			Rocket & \textbf{0.0069} & \textbf{0.0069} & 0.0078 & 0.0070 & 0.0070 & 0.0073 \\
			\hline
			Skateboard & \textbf{0.0079} & 0.0080 & 0.0080 & \textbf{0.0079} & 0.0082 & 0.0083 \\
			\hline
		\end{tabular}
		\caption{Mean distance between the best recovered point cloud and the corresponding original point cloud.}
	\end{subtable}
	\\
	\par\bigskip
	\begin{subtable}{0.9\linewidth}
		\captionsetup{labelformat=parens}
		\centering
		\begin{tabular}{|c||c|c|c|c|c|c|}
			\hline
			Categories & \tabincell{c}{HKC} & \tabincell{c}{HKF} & \tabincell{c}{LHF} & \tabincell{c}{GFR} & \tabincell{c}{EA} & \tabincell{c}{PCA-AC}\\
			\hline
			Cap & 2628.3 & 2632.4 & 2315.7 & \textbf{2635} & 1427.5 & 2160.4 \\
			\hline
			Chair & 2656.1 & 2657.9 & \textbf{2658} & 2653.9 & 2458.2 & 2469.9 \\
			\hline
			Laptop & 2754.4 & 2784.4 & \textbf{2785.6} & 2774.3 & 2626.4 & 2584.6 \\
			\hline
			Mug & 2818.6 & \textbf{2819.9} & 2716.6 & 2810.7 & 2633.7 & 2418.2 \\
			\hline
			Rocket & \textbf{2364.4} & 2361 & 2317.4 & 2360.6 & 1904.4 & 2004.5 \\
			\hline
			Skateboard & 2559.8 & 2562 & \textbf{2568.2} & 2563.1 & 2409.9 & 2381.2 \\
			\hline
		\end{tabular}
		\caption{Average number of points within $d_\theta$ between the best recovered point cloud and the corresponding original point cloud.}
	\end{subtable}
	\caption{Mean distance and the average number of points within $d_\theta$ between the best recovered point cloud and the
	 original point cloud for resampling ratio $\alpha=0.2$.}
	\label{table:2}
\end{table*}

To quantitatively illustrate the performance comparison, 
{Table.~\ref{table:2}} presents the numerical results for $\alpha=0.2$.
From the test results, we observe that our HKF algorithm exhibits
the best performance in terms of both
mean distance and number of matched points versus the
traditional GFR graph method and two edge detection methods.
Compared with the edge detection methods, our proposed algorithms
consistently retain larger numbers of points within the threshold 
$d_\theta$ in most point cloud categories. 

The comparison also
demonstrates a potential issue of the specialized
edge detection based methods. 
In particular,  resampled point cloud of edge detection methods 
may over-emphasize only part of the original point cloud, 
as the example of ``cap'' in {Fig.~\ref{fig:Eig1}} shows. As a result, 
substantial number of points may not be retained by 
the generic edge detection methods
during resampling. 

We also examine the effect of sampling ratio $\alpha$
and graphically illustrate the
variation of mean distance, average of distance and dual distance, and average number within threshold against 
different sampling ratios in {Fig.~\ref{fig:Curves}}.
{Here we use the same set of point clouds in the numerical results for $\alpha=0.2$.}
It is clear and intuitive that higher resampling ratio leads to better
performance of all methods under study. 
It is important to note that our proposed methods 
exhibit performances superior to the traditional methods in terms
of mean distance for various resampling ratios.
This result indicates that our proposed hypergraph-based methods 
tend to preserve the geometric information more efficiently
in resampling. 
The hypergraph-based methods also exhibits better results
with respect to the number of corresponding nodes 
between reconstructed and original point clouds.

In summary, our test results demonstrate the efficiency of 
the proposed resampling while preserving the underlying structural
features and geometric information among point cloud data. 
They further demonstrate that hypergraph presents a promising
alternative beyond regular graph 
for modeling point clouds in some point cloud related applications. 

\section{Conclusion}

This work investigates new ways for efficient and feature preserving
resampling of 
3D point cloud based on hypergraph signal processing
(HGSP).  We establish HGSP as an efficient tool to model 
multilateral point relationship and to extract features
in point cloud applications. We propose three new  methods based on 
HGSP kernel convolution and spectrum filtering. Although typical
HGSP tools tend to require high computational complexity,
our proposed algorithms bypass certain steps for hypergraph
construction and only require modest complexity
to implement. Our 
experimental results demonstrated that the proposed hypergraph
resampling algorithms can outperform traditional 
graph-based methods in terms of feature preservation
and robustness to measurement noises. 

Future works should consider the integration of HGSP methods 
with feature-based edge detection approaches 
to further enhance the edge preserving property of
resampling algorithms. Another interesting direction
of exploration is the design of different HGSP-based filters 
to capture multi-resolution local features from point cloud
data. 

\bibliographystyle{IEEEbib}

\begin{thebibliography}{99}
	\bibitem{d1} Z. C. Marton, R. B. Rusu and M. Beetz, ``On fast surface reconstruction methods for large and noisy point clouds,''
	\textit{2009 IEEE International Conference on Robotics and Automation}, Kobe, Japan, May 2009, pp. 3218-3223.
	
	\bibitem{d2} R. Schnabel, S. Möser and R. Klein, ``A Parallelly Decodeable Compression Scheme for Efficient Point-Cloud Rendering,'' in
	\textit{SPBG}, Prague, Czech Republic, Sep. 2007, pp. 119-128.
	
	\bibitem{d3} S. Gumhold, X. Wang and R. S. MacLeod, ``Feature Extraction From Point Clouds,'' in
	\textit{IMR}, Newport Beach, USA, Oct. 2001, pp. 293-305.
	
	\bibitem{g1} R. Q. Charles, H. Su, M. Kaichun and L. J. Guibas, ``PointNet: Deep Learning on Point Sets for 3D Classification and Segmentation,'' 
	\textit{2017 IEEE Conference on Computer Vision and Pattern Recognition (CVPR)}, Honolulu, HI, 2017, pp. 77-85.
	
	\bibitem{g2} Z. Yang, Y. Sun, S. Liu and J. Jia, ``3DSSD: Point-Based 3D Single Stage Object Detector,''
	\textit{2020 IEEE/CVF Conference on Computer Vision and Pattern Recognition (CVPR)}, Seattle, WA, USA, 2020, pp. 11037-11045.
	
	\bibitem{g3} T. Hackel, N. Savinov, L. Ladicky, J. D. Wegner, K. Schindler and M. Pollefeys, ``Semantic3D.net: A new Large-scale Point Cloud Classification Benchmark,''
	\textit{arXiv preprint arXiv:1704.03847}, Apr. 2017.
	
	\bibitem{c3} S. Chen, D. Tian, C. Feng, A. Vetro and J. Kovačević, ``Fast Resampling of Three-Dimensional Point Clouds via Graphs,'' in
	\textit{IEEE Transactions on Signal Processing},
	vol. 66, no. 3, pp. 666-681, Feb. 2018.
	
	\bibitem{c1} Z. Chen, T. Zhang, J. Cao, Y. J. Zhang and C. Wang, ``Point cloud resampling using centroidal Voronoi tessellation methods,''
	\textit{Computer-Aided Design},
	vol. 102, pp. 12-21, Apr. 2018.
	
	\bibitem{c2} S. Orts-Escolano, V. Morell, J. García-Rodríguez and M. Cazorla, ``Point cloud data filtering and downsampling using growing neural gas,''
	\textit{The 2013 International Joint Conference on Neural Networks (IJCNN)},
	Dallas, TX, USA, Aug. 2013, pp. 1-8.
	
	\bibitem{f1} A. Anis, P. A. Chou and A. Ortega, ``Compression of dynamic 3D point clouds using subdivisional meshes and graph wavelet transforms,''
	\textit{2016 IEEE International Conference on Acoustics, Speech and Signal Processing (ICASSP)}, Shanghai, China, Mar. 2016, pp. 6360-6364.
	
	\bibitem{d4} S. Chen, D. Tian, C. Feng, A. Vetro and J. Kovačević, ``Contour-enhanced resampling of 3D point clouds via graphs,''
	\textit{2017 IEEE International Conference on Acoustics, Speech and Signal Processing (ICASSP)}, New Orleans, USA, Mar. 2017, pp. 2941-2945.
	
	\bibitem{f1b} B. Kathariya, A. Karthik, Z. Li and R. Joshi, "Embedded binary tree for dynamic point cloud geometry compression with graph signal resampling and prediction," 2017 IEEE Visual Communications and Image Processing (VCIP), St. Petersburg, FL, Dec. 2017, pp. 1-4.
	
	\bibitem{c4} C. Weber, S. Hahmann and H. Hagen, ``Sharp feature detection in point clouds,''
	\textit{2010 Shape Modeling International Conference}, Aix-en-Provence, France, Jun. 2010, pp. 175-186.
	
	\bibitem{c5} D. Bazazian, J. R. Casas and J. Ruiz-Hidalgo, ``Fast and Robust Edge Extraction in Unorganized Point Clouds,''
	\textit{2015 International Conference on Digital Image Computing: Techniques and Applications (DICTA)}, Adelaide, Australia, Nov. 2015, pp. 1-8.
	
	\bibitem{f2} H. Huang, S. Wu, M. Gong, D. Cohen-Or, U. Ascher, and H. Zhang, ``Edge-aware point set resampling," in \textit{ACM transactions on graphics (TOG)}, vol. 32, no. 1, pp. 1-12, Feb. 2013.
	
	\bibitem{d5} S. Zhang, S. Cui and Z. Ding, ``Hypergraph-Based Image Processing,''
	\textit{2020 IEEE International Conference on Image Processing (ICIP)}, Abu Dhabi, United Arab Emirates, Oct. 2020, pp. 216-220.
	
	\bibitem{f3} A. Bretto, (2013). Hypergraph theory. An introduction. Mathematical Engineering. Cham: Springer.
	
	\bibitem{d11} A. Banerjee, A. Char and B. Mondal, ``Spectra of general hypergraphs,''
	\textit{Linear Algebra and its Applications}, vol. 518, pp. 14-30, Apr. 2017.
		
		
	\bibitem{d9} A. Ortega, P. Frossard, J. Kovačević, J. M. F. Moura and P. Vandergheynst, ``Graph Signal Processing: Overview, Challenges, and Applications,'' in
	\textit{Proceedings of the IEEE}, vol. 106, no. 5, pp. 808-828, May 2018.
	
	\bibitem{d8} S. Barbarossa and M. Tsitsvero, ``An introduction to hypergraph signal processing,''
	\textit{2016 IEEE International Conference on Acoustics, Speech and Signal Processing (ICASSP)}, Shanghai, China, Mar. 2016, pp. 6425-6429.
	
		\bibitem{d6} S. Zhang, Z. Ding and S. Cui, ``Introducing Hypergraph Signal Processing: Theoretical Foundation and Practical Applications,'' in
	\textit{IEEE Internet of Things Journal}, vol. 7, no. 1, pp. 639-660, Jan. 2020.
	
	\bibitem{d7} S. Zhang, S. Cui and Z. Ding, ``Hypergraph Spectral Clustering for Point Cloud Segmentation,'' in
	\textit{IEEE Signal Processing Letters}, vol. 27, pp. 1655-1659, Sep. 2020.
	
	\bibitem{c11} S. Zhang, S. Cui and Z. Ding, ``Hypergraph Spectral Analysis and Processing in 3D Point Cloud," in \textit{IEEE Transactions on Image Processing}, vol. 30, pp. 1193-1206, Dec. 2021.
		
	\bibitem{f5} A. Afshar, J. C. Ho, B. Dilkina, I. Perros, E. B. Khalil, L. Xiong, and V.
	Sunderam, ``Cp-ortho: an orthogonal tensor factorization framework for
	spatio-temporal data," in \textit{Proceedings of the 25th ACM SIGSPATIAL International Conference on Advances in Geographic Information Systems},
	Redondo Beach, CA, USA, Jan. 2017, p. 1-4.
	
	\bibitem{f6} J. Pan, and M. K. Ng, ``Symmetric orthogonal approximation to symmetric tensors with applications to image reconstruction," \textit{Numerical Linear Algebra with Applications}, vol. 25, no. 5, e2180, Apr. 2018.
	
	
	\bibitem{f7} T. G. Kolda, ``Orthogonal tensor decompositions," \textit{SIAM Journal on Matrix Analysis and Applications}, vol. 23, no. 1, pp. 243-255, Jul. 2006.	
	
	\bibitem{f8} A. K. Cherri, and M. A. Karim, ``Optical symbolic substitution: edge detection using Prewitt, Sobel, and Roberts operators," \textit{Applied Optics}, vol. 28, no. 21, pp. 4644-4648, Nov. 1989.	
		
	\bibitem{c6} R. B. Rusu, Z. C. Marton, N. Blodow, M. Dolha and M. Beetz, ``Towards 3D point cloud based object maps for household environments,''
	\textit{Robotics and Autonomous Systems}, vol. 56, no. 11, pp. 927-941, Nov. 2018.
	
	\bibitem{e1} M. Kazhdan, M. Bolitho and H. Hoppe, ``Poisson surface reconstruction,'' in
	\textit{Proceedings of the fourth Eurographics symposium on Geometry processing}, vol. 7, Jun. 2006.
	
	\bibitem{e2} H. Edelsbrunner and E. P. Mücke, ``Three-dimensional alpha shapes,''
	\textit{ACM Transactions on Graphics (TOG)}, vol. 13, no. 1, pp. 43-72, Jan. 1994.

	\bibitem{c12} A. X. Chang, T. Funkhouser, L. Guibas, P. Hanrahan, Q. Huang, Z. Li, S. Savarese, M. Savva, S. Song, H. Su and J. Xiao, ``Shapenet: An information-rich 3d model repository,''
	\textit{arXiv preprint arXiv:1512.03012}, Dec. 2015
	
\end{thebibliography}

\end{document}